%% file: main.tex
\documentclass[pra, aps, a4paper, amsfonts, amssymb, amsmath, reprint, nofootinbib, twoside, floatfix]{revtex4-1}

\input{preamble}
\usepackage[hidelinks,unicode]{hyperref}
\usepackage{cleveref}

\begin{document}

\title{Gradient Guided Furthest Point Sampling for Robust Training Set Selection}

\author{Morris Trestman$^{1,2}$, Stefan Gugler$^{1,2}$, Felix A. Faber$^{8}$, O. A. von Lilienfeld$^{1,2,3,4,5,6,7}$}
\email[Correspondence email addresses: ]{mosstrestman@gmail.com, stefan.gugler@tu-berlin.de, faber.felix1@gmail.com, anatole.vonlilienfeld@gmail.com}
\affiliation{$^1$Machine Learning Group, Technische Universität Berlin, 10587 Berlin, Germany}
\affiliation{$^2$Berlin Institute for the Foundations of Learning and Data, 10587 Berlin, Germany}
\affiliation{$^3$Chemical Physics Theory Group, Department of Chemistry, University of Toronto, St. George Campus, Toronto, ON, Canada}
\affiliation{$^4$Department of Materials Science and Engineering, University of Toronto, St. George Campus, Toronto, ON, Canada}
\affiliation{$^5$Vector Institute for Artificial Intelligence, Toronto, ON, Canada}
\affiliation{$^6$Department of Physics, University of Toronto, St. George Campus, Toronto, ON, Canada}
\affiliation{$^7$Acceleration Consortium, University of Toronto, Toronto, ON, Canada}
\affiliation{$^8$Data Science and Modelling, Pharmaceutical Sciences R$\&$D, AstraZeneca, Gothenburg, Sweden}

\begin{abstract}
Training set sampling methods are used to improve model performance and lower data costs in machine learning problems relevant to chemistry. We introduce Gradient Guided Furthest Point Sampling (GGFPS), a simple extension of Furthest Point Sampling (FPS) that leverages molecular force norms to guide efficient sampling of configurational spaces of molecules.
Numerical evidence is presented for a toy system (the Styblinski-Tang function) as well as for molecular dynamics trajectories from the MD17 dataset.
Our numerical results indicate superior data efficiency and model robustness when using GGFPS compared to FPS and uniform random sampling (URS), as well as established supervised FPS-style selectors, PCov-FPS and PCov-CUR.
Distribution analysis of the MD17 data suggests that FPS systematically under-samples equilibrium geometries, resulting in large test errors for relaxed structures.
GGFPS cures this artifact and (i) enables up to twofold reductions in training cost without sacrificing predictive accuracy compared to FPS in the 2-dimensional Styblinski-Tang system, (ii) systematically lowers prediction errors for equilibrium as well as strained structures in MD17, and (iii) systematically decreases prediction error variances across all of the MD17 configuration spaces.
These results suggest that gradient-aware sampling methods hold great promise as effective training set selection tools, and that naive use of FPS may result in imbalanced training and inconsistent prediction outcomes.
\end{abstract}

\maketitle
\input{sections/introduction.tex}

\input{sections/methods.tex}

\input{sections/results.tex}

\input{sections/conclusions.tex}

\input{sections/acknowledgements.tex}
\input{sections/data_availability.tex}

\bibliographystyle{apsrev4-1}
\bibliography{references}
\newpage
\input{sections/algo_appendix.tex}
\input{sections/appendix1.tex}

\end{document}

%% file: preamble.tex
\usepackage[utf8]{inputenc}
\usepackage[T1]{fontenc}
\usepackage[english]{babel}
\usepackage{csquotes}

\usepackage{amsthm}
\usepackage{amsmath}
\usepackage{amssymb}
\usepackage{mathtools}
\let\olddiv\div

\usepackage{xcolor}
\usepackage{graphicx}
\usepackage{float}
\usepackage{subcaption}
\usepackage[left=23mm,right=13mm,top=35mm,columnsep=15pt]{geometry}
\usepackage{adjustbox}
\usepackage{placeins}
\usepackage{lipsum}

\usepackage[ruled,vlined,linesnumbered]{algorithm2e}
\SetKwProg{Fn}{Function}{}{}
\SetKwComment{Comment}{$\triangleright$\ }{}
\DontPrintSemicolon

\usepackage{setspace}
\newcommand{\added}[1]{#1}
\newcommand{\deleted}[1]{}
\newcommand{\removed}[1]{}

%% file: sections/introduction.tex
\section{Introduction} \label{sec:intro}

\begin{figure*}
    \includegraphics[width=0.9\linewidth]{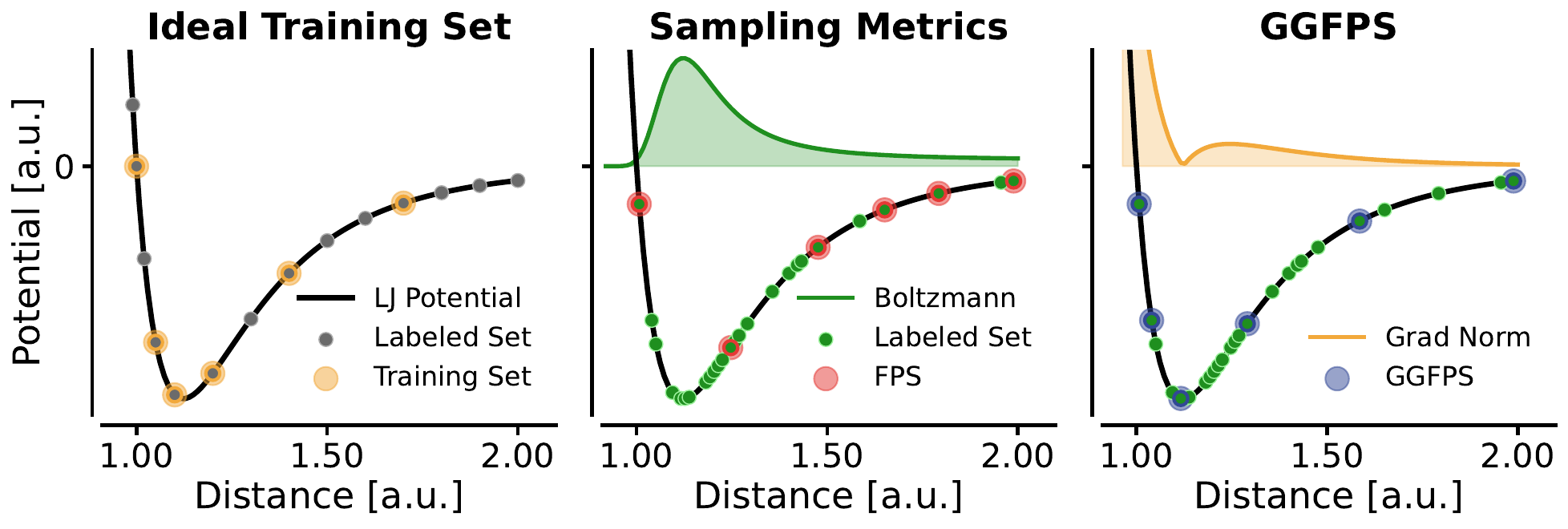}
    \caption{
    All three subplots show a toy Lennard-Jones potential in arbitrary units of energy over distance.
    Left: A subset (orange) of labeled data (grey), which represent a well-performing training set from the labeled reaction data.
    Center: Furthest point sampling (FPS, red) on the Boltzmann-distributed labeled set (green markers) is shown on the same Lennard-Jones potential, missing adequate sampling at the energy minimum. The Boltzmann distribution (green) is shown as an overlay.
    Right: Gradient Guided FPS (GGFPS, blue) is shown on the same labeled set (green), striking a balance between FPS and gradient information (Grad Norm, yellow).
    }
    \label{fig:Combined_Figure}
\end{figure*}

Chemical space is vast, encompassing an immense variety of  chemical structures and reactions.
Even confined to a single potential energy surface (PES), the cost of adequate sampling increases exponentially with the size of the molecular system.
Similarity-exploiting machine learning (ML) methods,\cite{scholkopf2018,hofmann2008,Herbrich2001,vapnik2013book}
such as kernel ridge regression (KRR) lower this cost by predicting molecular configurations outside of the sampled dataset.\cite{RuppPRL2012,Behler2007-oc,Bartok2010-ni,Von_Lilienfeld2018-wk, AssessmentMLJCTC2013,BobPaper,googlePaper2017,simm2018,proppe2019,gugler2022,gugler2025}
KRR is particularly suitable for small to intermediate-sized datasets,\cite{zhang2018,Chmiela2019-xq,Christensen2020-xs} typically in the hundreds of data points range.
However, it still requires a number of highly accurate yet expensive reference calculations to train the ML model.\cite{butler2018,himanen2019,batra2021,ramprasad2017,schmidt2019,schleder2019,dral2020,carleo2019}

One way to lower the cost of PES prediction is to generate a large, diverse initial dataset, and then select the most informative subset as training points for higher-level reference calculations.\cite{Dragoni2018, Cersonsky2021, Bernstein2019, Wengert2021}

Figure \ref{fig:Combined_Figure} (left) shows a toy example for a diatomic potential:
a large set of labeled data (grey) is needed to obtain the ground truth, the Lennard-Jones (LJ) potential (black), but the goal is to have a smaller, information-dense training set (orange).
Datasets used to train such single-PES models, such as MD17\cite{chmiela2017machine} and ISO17\cite{schnett2017, Schtt2017, DataPaper2014}, are typically generated via molecular dynamics simulations and are sampled according to the Boltzmann distribution. However, as shown in Figure \ref{fig:Combined_Figure} (middle), Boltzmann distributed datasets (green) contain redundancies and undersample higher energy molecular configurations. Therefore, the uniform random sampling of Boltzmann distributed datasets (URS \cite{mannodi2016machine, Botu2016,Bartk2017}) often reduces model robustness (increasing the incidence of outlier test errors), e.g. when predicting transition state and reactive structures.\cite{Van_der_Oord2023-ia}

Such challenges have given rise to diverse sampling techniques, which aim to generate compact training sets with even coverage of the relevant configuration space, while retaining as much of the predictive power of the larger dataset as possible.\cite{Clerse2024,Nguyen2018, Qi2024,deringer2021,Riquelme-Granada2021-wz,Wen2023-vv,He2009-mz}
These sampling techniques can be broadly broken down into unsupervised and supervised methods.

Unsupervised sampling methods do not require labeled data, operating only on molecular descriptors.
A well-known method is furthest point sampling (FPS), which selects new configurations that maximize a distance metric in descriptor space from previously selected configurations.\cite{Rosenkrantz1977-bo,fps1997eldar}
FPS is shown schematically in Figure \ref{fig:Combined_Figure} (center, red).
By enforcing even sampling over all of the original dataset, FPS often increases model robustness, usually at the cost of an increase in the overall interpolative test error due to undersampling densely packed dataset regions (in the Figure \ref{fig:Combined_Figure} Lennard-Jones potential, FPS entirely overlooks the minimum energy configuration).\cite{Bartk2017, Imbalzano2018, Rowe2020, Wengert2022, Boulangeot2024, jmi.2025.10}
Other unsupervised dataset sampling methods include FPS variants applied to local atomic environments\cite{Li2024}, entropy-based optimization\cite{MontesdeOcaZapiain2022, Karabin2020}, CUR decomposition and Pearson Correlation based data pruning\cite{Imbalzano2018}, and stratified sampling\cite{Qi2024}, all aiming to reduce data redundancy and achieve good accuracy-transferability trade-offs.

When training data regression targets (labels) are available, supervised training set sampling methods incorporate them into the sampling strategy.
One type of supervised sampling is domain expertise, where FPS training sets are augmented to contain additional strained and transition state structures, in order to bias the dataset towards a specific design principle.\cite{Rowe2020,Deringer2020,Zuo2020}
But this approach may introduce human bias, limit model robustness, and become impracticable for larger datasets and more complex models.\cite{Jia2019,MontesdeOcaZapiain2022, Karabin2020, Musil2018-ao}
In contrast, automated supervised sampling methods incorporate labels directly either as sampler loss functions or as sampling metrics.
They include using enthalpy histograms to filter CUR-sampled datasets\cite{Bernstein2019}, iteratively adding configurations when predicted energies differ significantly from reference data\cite{Smith2018-ij},
as well as a host of active learning methods applied to labeled datasets\cite{Wen2023AdaptiveSubsamplingAL_,Yin2023UncertaintyAL_Generalization, Subedi2024EmpoweringAL_3DMolGraphs}.
The active learning samplers, while effective, tend to be orders of magnitude more costly than other automated sampling methods.
Lower cost and widely used alternatives are Principle Covariates Furthest Point Sampling (PCov-FPS) and Principal Covariates CUR Decomposition (PCov-CUR), which seek to maximize distances in both descriptor and label space (in the case of PCov-FPS) and maximize sample `importance' in a joint descriptor–label space (in the case of PCov-CUR).\cite{Cersonsky2021}

In this work, we introduce a supervised sampling method that combines FPS with a Euclidean force norm bias, which we call Gradient Guided Furthest Point Sampling (GGFPS).
Forces are the negative gradient of the energy with respect to the atomic positions. They are often available in reference energy calculations, and have long been incorporated into the kernels and loss functions of ML models.\cite{HF,Schatz2000-tz, Chmiela2017-al, Faber2018-vg, Christensen2020-xs}. We use the $L^2$ norm of the gradient, as shown in Figure \ref{fig:Combined_Figure} (right, yellow overlay).
Additionally, force and Hessian information have been used successfully in dataset binning for training force fields and for active learning.\cite{Huan2017,Panknin2023}

Because molecular force norms indirectly describe the variance of molecular energy labels, using force norms as a sampling metric provides a way to cover variance not just in descriptor space, but also in label space. GGFPS contains a gradient bias hyperparameter which adjusts the sampling to the property distributions of different labeled datasets.
A GGFPS training set is shown in Figure \ref{fig:Combined_Figure} (right, blue), capturing both the energy minimum and tail ends of the dissociation curve.

Gradient norms are shown for reference (right, yellow).

Our results demonstrate that while FPS results in lower test errors than URS in low-dimensional uniformly distributed datasets, FPS often leaves large sections of non-uniformly distributed datasets like molecular trajectories unsampled, resulting in worse than random performance. We show that GGFPS training sets span the entire dataset across all tested systems and outperform both URS and FPS, as well as PCov-FPS and PCov-CUR, which were computed using the \texttt{scikit-matter} (skmatter) implementation \cite{scikit-matter}.

This paper is structured as follows: In Section \ref{section:methods} we introduce the GGFPS algorithm, and in Section \ref{section:results} we compare  GGFPS to FPS, PCov-FPS, PCov-CUR, and URS across a toy function and the molecular dynamics trajectories of the MD17 dataset.

%% file: sections/methods.tex
\section{Methods} \label{section:methods}

\subsection{Furthest Point Sampling (FPS)} \label{subsection:fps}

Let $\mathbf{X} = [\mathbf{x}_1^\top, \mathbf{x}_2^\top, \dots, \mathbf{x}_{N_\text{tot}}^\top]^\top \in \mathbb{R}^{N_\text{tot} \times d}$ be the matrix of all molecular descriptors (each row is a $d$-dimensional descriptor of a molecule $\mathbf{x}_i \in \mathbb{R}^d$). Define the labeled dataset as $\mathcal{L} = \{1, 2, \dots, N_\text{tot}\}$, i.e., a set of indices into $\mathbf{X}$.
The FPS algorithm constructs a training subset $\mathcal{T} \subset \mathcal{L}$ of size $N = |\mathcal T |$ through the following procedure:
First, the training set is initialized as $\mathcal{T} = \emptyset$ and the unsampled set as $\mathcal{A} = \mathcal{L}$.
The algorithm then selects an initial point $c \in \mathcal{A}$ randomly and updates the training subset $\mathcal{T} \leftarrow \mathcal{T} \cup \{c\}$ and the unsampled set $\mathcal{A} \leftarrow \mathcal{A} \setminus \{c\}$.
As long as the size of the training subset is smaller than $N$, more points are selected.
Thus, $\mathcal{T} \cup \mathcal{A} = \mathcal{L}$.
For this, the minimum Euclidean distances between all points $j \in \mathcal A$ and $i \in \mathcal T$ are calculated in a vector $\mathbf{d}$ with elements
    \begin{equation}
        d_j = \min_{i \in \mathcal{T}} D_{ij} \ ,
        \label{eq:D_matrix}
    \end{equation}
where $\mathbf{D} \in \mathbb{R}^{N_\text{tot} \times N_\text{tot}}$ is the pairwise distance matrix with entries
    \begin{equation}
        D_{ij} = \|\mathbf{x}_i - \mathbf{x}_j\|_2 \ .
    \end{equation}
where distances are often computed on-the-fly to avoid $\mathcal{O}(N_\text{tot}^2)$ memory.
We select a new point to add to the training set that is furthest away among all these minimal distances, i.e. select
    \begin{equation}
        j^{*} = \operatorname{arg}\max_{j \in \mathcal{A}} d_j \ .
        \label{eq:min-dist-selection}
    \end{equation}
If multiple points maximize $d_j$, one is selected arbitrarily.
Now the training subset $\mathcal{T}$ and unsampled set $\mathcal{A}$ are updated as before.

In this work, we used the \texttt{fps\_sampling} function from the \texttt{fpsample} Python package for easy reproduction of our timing comparisons\cite{fpsample}.

\subsection{Gradient Guided Furthest Point Sampling (GGFPS)} \label{subsection:ggfps}

GGFPS extends FPS by including gradient norm information
$\mathbf{g} = (g_1,\dots,g_{N_\text{tot}})^\top$
    with $g_i= \{\|\mathbf{F}_i\|_2\}$
    where $\mathbf{F}_i \in \mathbb{R}^{M \times 3}$ are forces in three spatial directions for configuration $i$, and $M$ is the number of atoms of the system.
The algorithm balances geometric spread (FPS part) with  interpolated biases between low and high-gradient regions (GG part).

The GGFPS algorithm can be structured to accept a full distance matrix $\mathbf{D}$ over all points $j \in \mathcal L$, which is useful in many kernel applications because $\mathbf{D}$ can be reused. Alternatively, GGFPS can calculate distance values on-the-fly, which is faster for single-shot sampling purposes. The full distance matrix version is described below, while the pseudo-code for on-the-fly GGFPS can be found in SI (Algorithm~\ref{salg:GGFPS_on_the_fly}).

The algorithm proceeds as follows:
The gradient norms $\mathbf{g}$ and distance matrix $\mathbf{D}$ are computed.
Training set $\mathcal{T}$ and unsampled set $\mathcal{A}$ are initialized as in FPS and the minimum distances are set to an arbitrarily large number.
Among all points $j \in \mathcal L$, we sample an initial point $c$ with the probability
\begin{equation}
\label{eq:GGFPSprob1}
p_j = \frac{(g_j+\epsilon)^{\beta_0}}{\sum_{\ell\in\mathcal L}(g_\ell+\epsilon)^{\beta_0}},
\end{equation}
where $\epsilon$ is a small value used to avoid errors when $g_j=0$, and $\beta_0$ is the first value of the gradient norm biasing $\beta$-schedule defined below.

 Alternative initialization strategies include selecting a configuration with the highest gradient norm, the lowest gradient norm, or a random configuration.

After sampling $c$, we update $\mathcal{T} \leftarrow \mathcal{T}\cup\{c\}$ and $\mathcal{A} \leftarrow \mathcal{A}\setminus\{c\}$.
The minimum-distance vector is initialized as
\begin{equation}
d_j \leftarrow D_{c j}, \quad \forall j \in \mathcal{A}.
\end{equation}

To guide each subsequent selection, GGFPS computes a weighted score $s_j$ for every candidate $j \in \mathcal{A}$, where at sample selection step $k = 1, \dots, N-1 $,
\begin{equation}\label{eq:Dgradscore}
s_j = \bigl(g_j+\epsilon\bigr)^{\beta_k}\, d_j \ .
\end{equation}
Here, $\beta_k$ is a gradient norm biasing parameter and $d_j$ is the current minimum distance of point $j$ to the existing set $\mathcal{T}$. When $\beta_k$ is positive, high-gradient configurations are favored; when $\beta_k$ is negative, low-gradient configurations are favored; when $\beta_k = 0$, standard FPS is recovered.

\begin{figure}[t]
    \centering
    \includegraphics[width=0.9\linewidth]{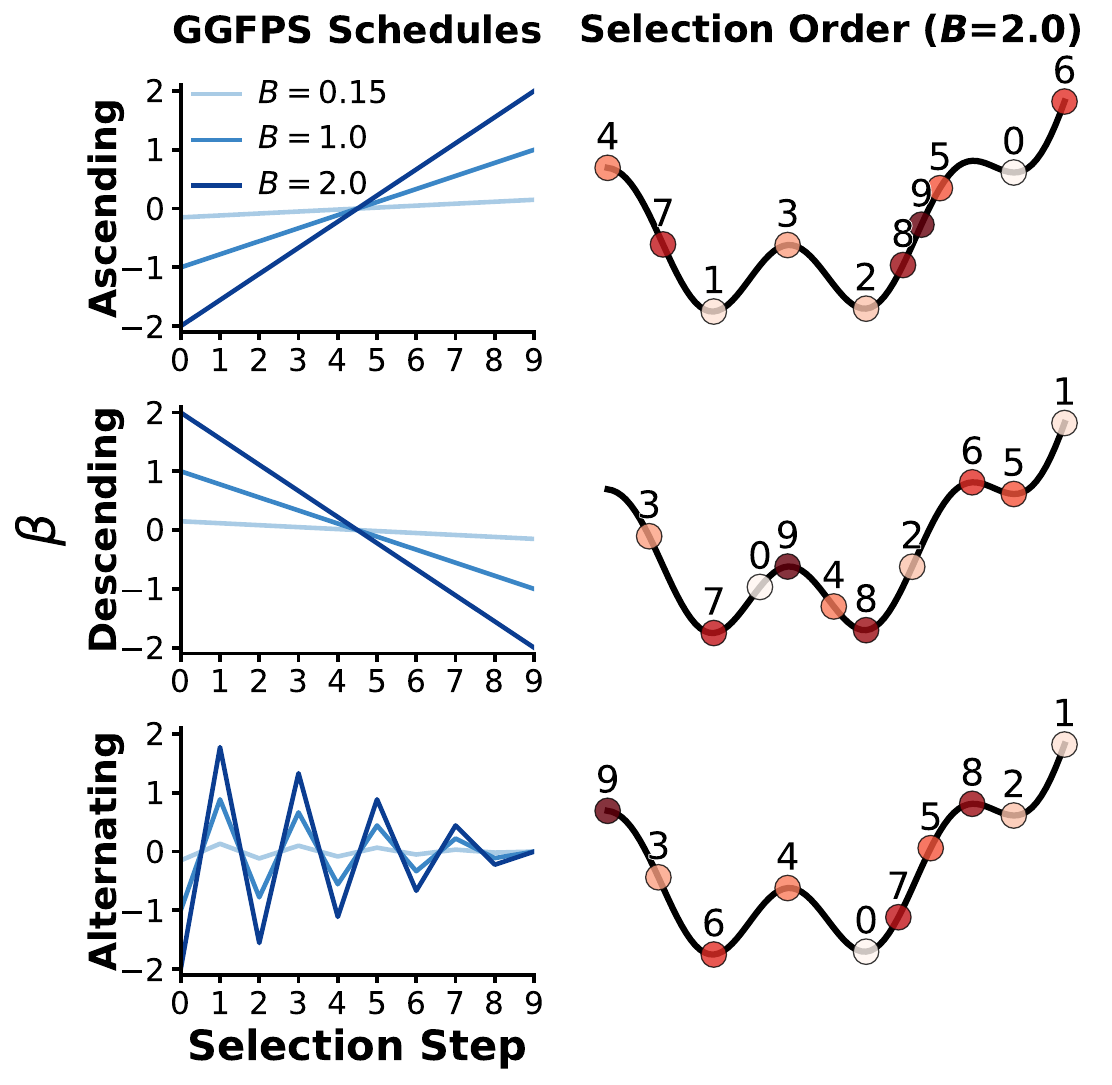}
    \caption{ 
    Left: GGFPS $\beta$ values w.r.t. 10 training point selection steps. The x-axis denotes the selection step $k \in \{0,\dots,9\}$ and the y-axis denotes the schedule value $\beta_k \in [-B,B]$. Color encodes the sweep range $B$, from light to dark blue: $B=0.15$, $B=1.0$, $B=2.0$. Panels (top to bottom) show the ascending, descending, and alternating schedules used throughout the paper.
    Right: The corresponding GGFPS training point selection steps across the same 3 schedules (rows), plotted on a $1D$ toy function at $B=2.0$. The selection order is printed above the training points and is encoded in their colors (light to dark red).
    }
    \label{fig:Methods}
\end{figure}
\FloatBarrier

To prevent overfitting to a specific gradient norm value, $\beta$ interpolates between symmetric gradient-bias bounds.
The bounds hyperparameter $B>0$ defines a range $\left[-B, B\right]$ that is partitioned into $N$ values (one per selection step), in a descending, $\{\beta^{(N)}, \beta^{(N-1)}, ..., \beta^{(1)}\}$, ascending, $\{\beta^{(1)}, \beta^{(2)}, ..., \beta^{(N)} \}$, or alternating, $\{\beta^{(N)}, \beta^{(1)}, \beta^{(N-1)}, \beta^{(2)}, ... \}$, fashion. We re-index this as $\{\beta_k\}_{k=0}^{N-1}$ and call it a \textit{schedule}.
The GGFPS $\beta$ schedules and their corresponding training set selection orders (visualized on a $1D$ toy function) are shown in Figure \ref{fig:Methods}. Further selection orders across a range of $B$ values can be found in SI Figure~\ref{fig:SI_ggfps_beta_1d_schedules}.

The second column of Figure \ref{fig:Methods} also shows that the ascending schedule oversamples high gradient norm function regions, while the descending schedule does the opposite.
Unless explicitly stated otherwise, we use the descending schedule in our MD17 experiments and ascending in the ST experiments, as explained in Section \ref{section:results}.
In addition, one may treat $B$ itself as a hyperparameter and evaluate a small set of candidate $B$ values (e.g., a grid of four $B$ values) via cross-validation, selecting the best-performing $B$ for a given dataset and training set size.

At each iteration, analogously to Eq. \ref{eq:min-dist-selection}, the candidate point with the highest score
\begin{equation}\label{eq:candpoint}
j^{*} = \operatorname{arg}\max_{j \in \mathcal{A}} s_j
\end{equation}
is selected,
and $\mathcal{T}$, $\mathcal{A}$, and $\mathbf{d}$ are updated by
\begin{equation}
d_j \leftarrow \min\!\left(d_j, D_{jj^{*}}\right), \quad \forall j \in \mathcal{A}.
\end{equation}
These steps repeat until the desired number of training points $N$ is reached or until $N_\text{tot}$ is exhausted. The GGFPS algorithm and schedule pseudocode are provided in the SI (Algorithm~\ref{salg:GGFPS_fullD}),
and the runnable GGFPS code is available in the GGFPS repository\cite{ggfps_code}.

\subsection{Kernel Ridge Regression}\label{subsection:KRR}

We use Kernel Ridge Regression (KRR) as the predictive model to evaluate the performance of our GGFPS method.
KRR establishes a nonlinear mapping from molecular descriptors, $\mathbf{x} \in \mathbb{R}^d$, to target properties $y \in \mathbb{R}$ through the kernel trick, which implicitly projects inputs into a reproducing kernel Hilbert space.\cite{hofmann2008,scholkopf2002learning,krige1951statistical,vapnik2013book}

For a query molecule with descriptor $\mathbf{x}_q$, the predicted property $\hat{y}_q$ is expressed as
\begin{equation}
\hat{y}_q =
    \sum_{i=1}^{N} \alpha_i k(\mathbf{x}_i, \mathbf{x}_q)
\end{equation}
where
    $k: \mathbb{R}^d \times \mathbb{R}^d \rightarrow \mathbb{R}$ is a positive-definite kernel function, and
    $\boldsymbol{\alpha} \in \mathbb{R}^{N}$ are the dual coefficients learned during training.
We use a different kernel function for each of our systems, as described in Section \ref{subsection:repsandtargets}.

The matrix formulation for predictions on a test set $\{\mathbf{x}_q\}_{q=1}^{N_{\text{test}}}$ becomes
\begin{equation}
    \hat{\mathbf{y}}_{\text{test}}
    = \mathbf{K}_{\text{test}} \boldsymbol{\alpha}
\end{equation}
with the kernel between train and test points, $\mathbf{K}_{\text{test}} = k(\mathbf{x}_i, \mathbf{x}_q) \in \mathbb{R}^{N \times N_{\text{test}}}$.

The dual coefficients are obtained through regularized least-squares minimization,
\begin{equation}
    \boldsymbol{\alpha}
    = \left(\mathbf{K}_{\text{train}} + \lambda \mathbf{I}\right)^{-1} \mathbf{y}_{\text{train}}
\end{equation}
where $\mathbf{K}_{\text{train}} = k(\mathbf{x}_i, \mathbf{x}_j) \in \mathbb{R}^{N \times N}$ is the training kernel matrix and $\lambda > 0$ is the regularization parameter, while $\mathbf{y}_{\text{train}}$ contains quantum chemical reference values.

\subsection{Representations and kernel functions} \label{subsection:repsandtargets}

We tested our algorithm on the 2D Styblinski-Tang function and the aspirin MD17 trajectory.
In this section, we discuss their form and representation.

\subsubsection{The Styblinski-Tang Function}\label{subsubsection:ST function methods}

\begin{figure*}
    \includegraphics[width=\linewidth]{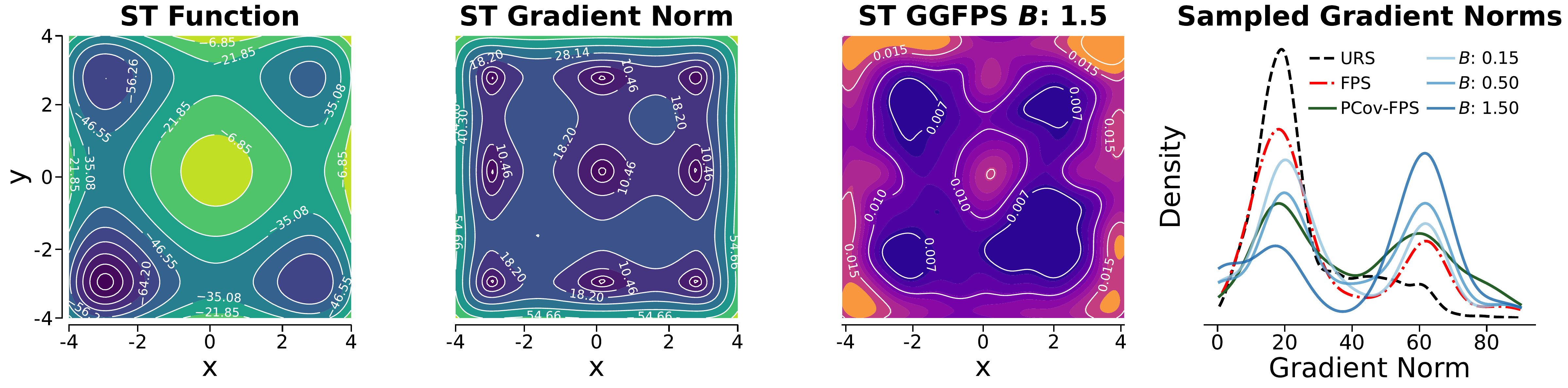}
    \caption{Left: A contour plot of the ST function surface in two dimensions.
        Center-left: The ST function gradient norm surface.
        Center-right: a heatmap of 50 ST function training sets generated by GGFPS, each with 100 data points, with $B=1.5$.
        In each of the three, a lighter color denotes a higher value.
        Right: ST function gradient norm distributions of training sets selected via
        URS (black dashed line),
        FPS (red dash-dotted line),
        PCov-FPS (solid green line), and GGFPS over increasing $B$ values (solid blue lines with increasing hue for $B \in \{0.15, 0.5, 1.5\}$).
        Training sets were sampled from 50 uniform randomly labeled sets of 1000 data points each.}
    \label{fig:STGRADHEATMAP}
\end{figure*}

The Styblinski-Tang (ST) function is a multi-modal, $d$-dimensional benchmark function used to test optimization algorithms.\cite{jamil2013literature, styblinski1990experiments}
It contains a mixture of wells surrounded by steep walls and is defined for $d=2$ as
\begin{equation}
    f(\mathbf{x}) = \frac{1}{2} \sum_{i=1}^{2} \left( x_i^4 - 16x_i^2 + 5x_i \right)
\end{equation}
where $\mathbf{x} = [x_1, x_2]^T \in \mathbb{R}^2$. The global minimum of the function is located at $\mathbf{x}^* = [-2.903534, -2.903534]$, with a function value of $f({\mathbf{x}}^*) = -78.33198$.
The configurations are uniformly sampled from the domain $[-4, 4]$ on each coordinate. The Cartesian coordinates are used directly as input descriptors $\mathbf{x}$.

We use a Gaussian kernel function as the similarity metric,
\begin{equation}
k(\mathbf{x}_i, \mathbf{x}_j) = \exp \left( - \frac{\|\mathbf{x}_i - \mathbf{x}_j\|_2^2}{2\sigma^2} \right)
\end{equation}
where $\sigma$ is the bandwidth hyperparameter of the kernel function.

\subsubsection{The MD17 Trajectories}
We represent the configurations of the MD17 aspirin, toluene, malonaldehyde, naphthalene, paracetamol, and uracil trajectories\cite{chmiela2017machine} with FCHL19 \cite{FCHL19}, a faster albeit slightly less accurate version of the original Faber–Christensen–Huang–Lilienfeld (FCHL) representation.\cite{FCHL} FCHL19 is a smooth local representation that contains radial and angular distributions of atoms across a given molecule's atomic environments. We use a local Gaussian kernel to compute similarities between configurations, where each kernel element represents the pairwise summation over kernel similarities between the atomic environments of the configurations\cite{bpkc2010},

\begin{equation}
    k(\mathbf{x}_{A}, \mathbf{x}_{B}) = \sum_{i \in A} \sum_{j \in B} \delta_{Z_i Z_j} \exp \left( - \frac{\|\mathbf{x}_i - \mathbf{x}_j\|_2^2}{2\sigma^2} \right) \ ,
\end{equation}
for atoms $i$ and $j$ in molecules $A$ and $B$ (encoded here as FCHL representations), $Z_{i}$ and $Z_{j}$ are their respective nuclear charges, and $\delta$ is the Kronecker delta.

%% file: sections/results.tex
\section{Results and Discussion}
\label{section:results}

\subsection{Sampling the Styblinski-Tang function}\label{subsection:ST results}

Here we apply URS, FPS, PCov-FPS, PCov-CUR, and GGFPS (single-$B$), to the ST function in 2 dimensions.
The ST function values and gradient norms are shown in Figure \ref{fig:STGRADHEATMAP} (left and center-left, respectively).
In both plots, lighter colors denote higher values.

To illustrate the result of the GGFPS sampling, a heatmap of 50 training sets, each with 100 data points, is shown in Figure \ref{fig:STGRADHEATMAP} (center-right), where lighter colors indicate a higher sampling density.

Here, GGFPS uses probabilistic initialization, as motivated by the ablation shown in SI Figure \ref{fig:SI_st_init_schedule_ablation}, along with the ascending $\beta$ schedule and with a $B$ value of $1.5$. The schedule choice is supported by $B$-sensitivity ablations across $\beta$ schedules (SI Figure \ref{fig:SI_st_b_dependencies}), and by the $B$ hyperparameter-count ablation across both $\beta$ schedules and $B$ grids (SI Figures \ref{fig:SI_st_b_grid_density_mae} and  \ref{fig:SI_st_b_grid_density_variance}). We investigated the $B$ value both by itself (SI Figure \ref{fig:SI_st_b_sweep_errors_vs_norms}), and compared to a $4B$ hyperparameter grid (SI Figure \ref{fig:SI_st_single_vs_multi_b}).

Figure \ref{fig:STGRADHEATMAP} (right) shows that for 1{,}000 samples, URS (black dashed) peaks around gradient norms of 20, while FPS (red dash-dotted) shows a bimodal distribution at  20 and 60.
PCov-FPS (green, solid) exhibits a flatter bimodal behavior.

GGFPS is shown with 3 different $B$ values (solid blue colored lines). At $B=0.15$, GGFPS almost recovers FPS. Increasing $B$ values smoothly increase the sampling of function gradient extrema.
This oversampling of high ST function gradient norms is specific to the ascending GGFPS schedule, as the descending schedule over-samples low gradient norms, and the alternating schedule sits somewhere in the middle (SI Figure \ref{fig:SI_st_train_grad_norm_dists}).
Because the ST function is uniformly distributed, data points along its high gradient norm edge regions number much higher than if the function were Boltzmann distributed. The ascending schedule therefore outperforms the other GGFPS schedules in this context.

\begin{figure}
    \centering
    \includegraphics[width=0.9\linewidth]{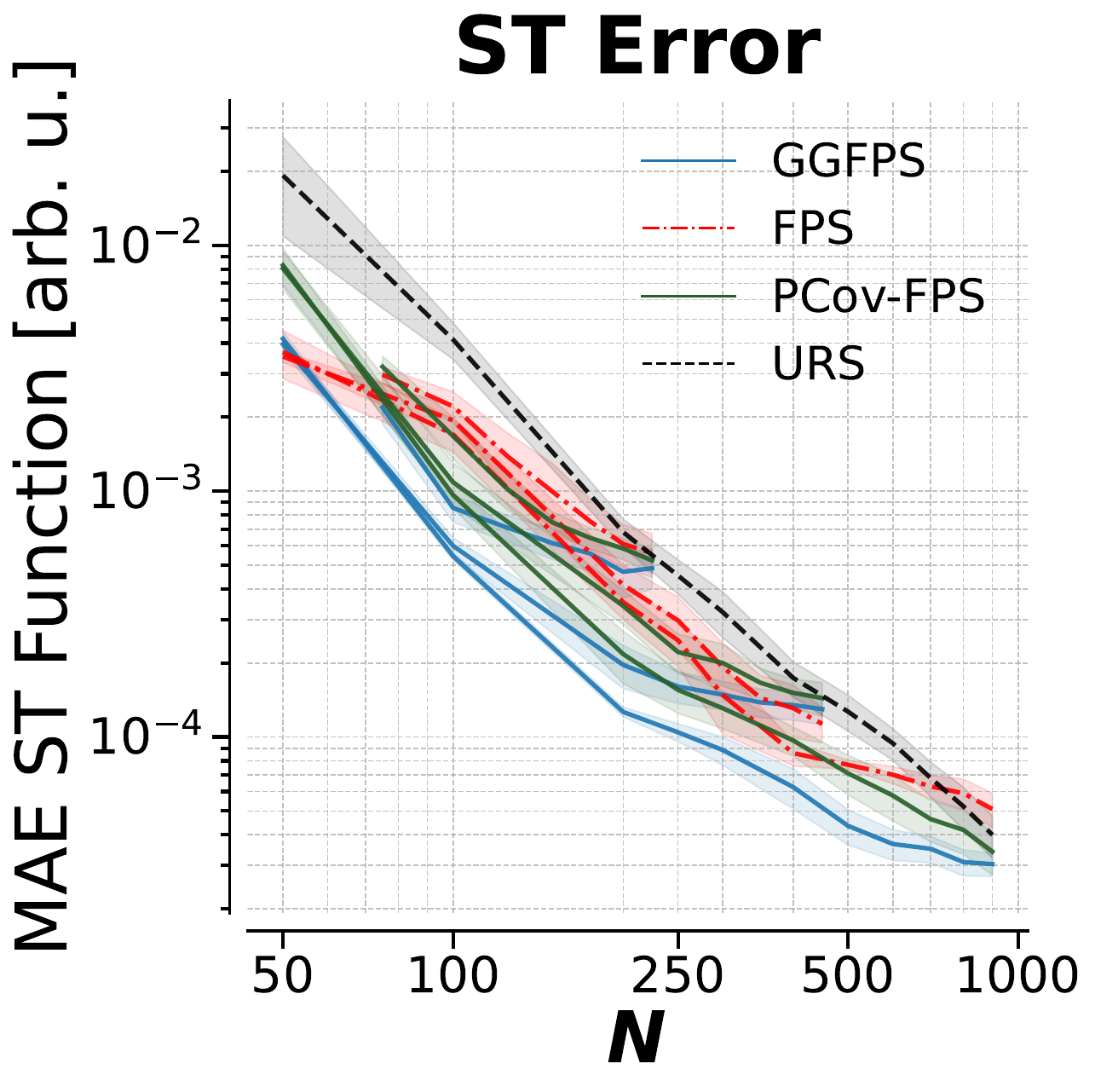}
    \caption{
    The MAE of the 2D ST function surface with respect to training set size for GGFPS (single-$B$), FPS, PCov-FPS, and URS (blue solid, red dash-dotted, green solid and black dashed lines).
    The curves are read from right to left in the plot, i.e. from a labeled set size of 1{,}000 data points (whose error is shown \textit{via} URS), FPS, PCov-FPS, and GGFPS sub-select from 950 to 50 training points.
    }
    \label{fig:ST_LC}
\end{figure}

With the $\beta$ range set, we generate ST function data and train models on training sets generated by URS, FPS, PCov-FPS, and GGFPS. PCov-CUR models performed poorly and are omitted here.
Labeled sets of sizes $\{250, 500, 1{,}000\}$ are selected \textit{via} URS.
GGFPS, FPS, and PCov-FPS training sets are sub-sampled from these labeled sets, with the remainder used for testing.
All sets are bootstrapped 100 times.
The KRR hyperparameters are generated using 5-fold grid-search cross-validation (CV).
As shown in SI Figures \ref{fig:SI_st_b_grid_density_mae}, \ref{fig:SI_st_b_grid_density_variance}, and \ref{fig:SI_st_single_vs_multi_b}, for the ascending schedule single and multi-$B$ GGFPS performance broadly converge, allowing single-shot GGFPS to be used, lowering the sampling cost to that of vanilla FPS.

\begin{figure}
    \centering
    \includegraphics[width=1\linewidth]{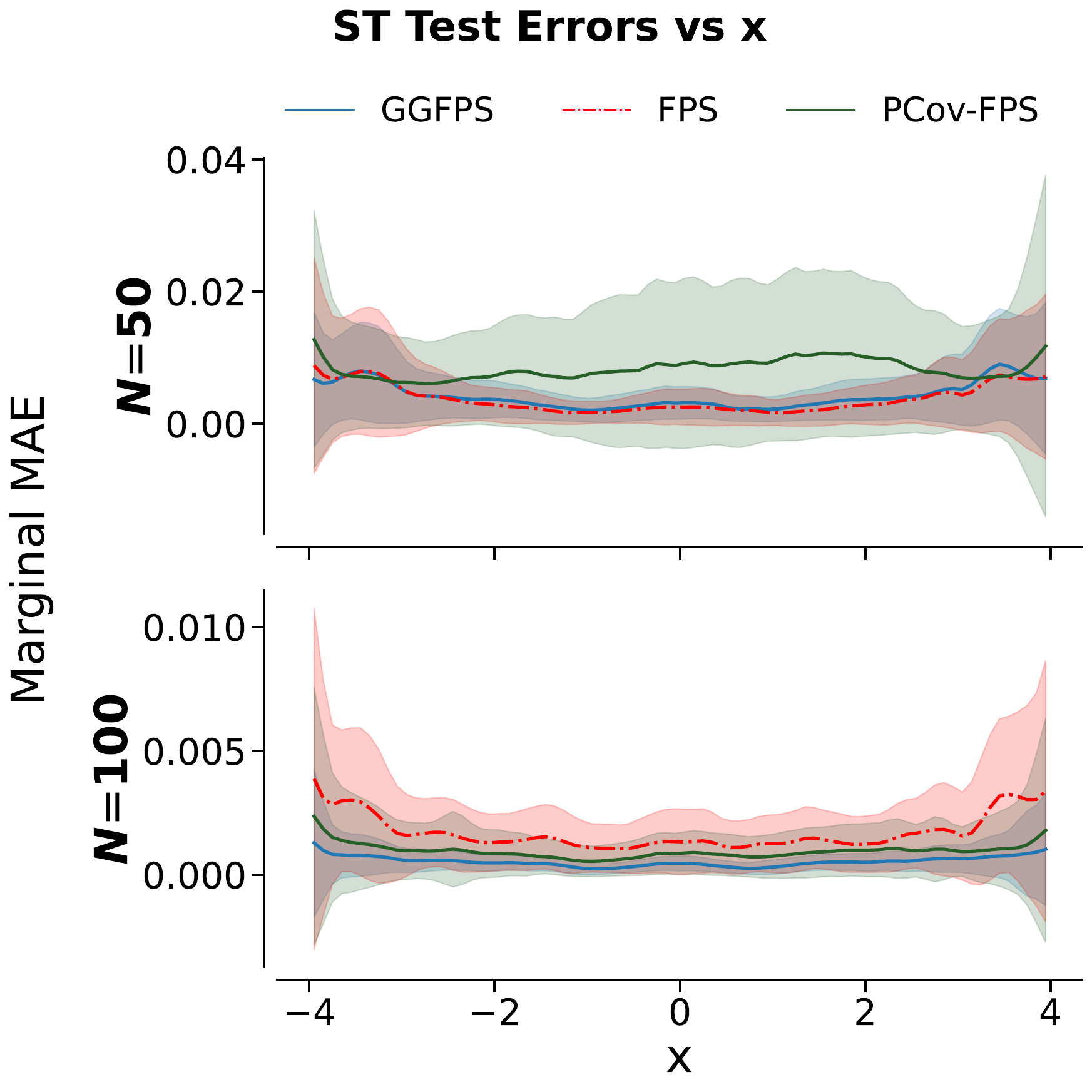}
    \caption{ST function test errors marginalized \textit{via} mean-binning onto the function $x$ axis, comparing GGFPS to FPS and PCov-FPS for representative training set sizes ($N=50$ and $N=100$ out of $N_\text{tot}=1,000$).}
    \label{fig:st_error_marginal}
\end{figure}

The ST function learning curves for URS (dashed black), GGFPS (solid blue), FPS (red dash-dotted), and PCov-FPS (solid green) are shown in log-log scale in Figure \ref{fig:ST_LC}.
The FPS and GGFPS training sets sampled from the labeled set of 1{,}000 data points start at an $N$ of 950. These 950 data points are the ``best'' performing out of the initial 1{,}000. We continue this sampling process until we reach a training set size of 50. The corresponding learning curves are therefore read from right to left in the plot. The same is true for initial labeled set sizes of 500 and 250. The URS learning curves, by virtue of uniform sampling, are invariant to the initial labeled set size.

Across all labeled set sizes, FPS requires on average 1.5 times fewer training points than URS to achieve the same MAE. The performance gap between FPS and URS increases with lower training set sizes for all labeled set sizes.
PCov-FPS outperforms FPS at all labeled set sizes and training set sizes, except for $N < 70$.
GGFPS improves upon the predictive performance of FPS and PCov-FPS by up to a factor of 3 for the same number of training points and the same labeled set size. It also improves predictive efficiency by up to a factor of 2 compared to FPS, achieving the same predictive accuracy with half the number of training points.
Compared to URS, GGFPS training sets match the MAE of the entire labeled set with, on average, half the total training set size across all labeled set sizes.
For smaller labeled sets and training set sizes, the performance gap between GGFPS and FPS shrinks.

Figure \ref{fig:st_error_marginal} shows the GGFPS, FPS, and PCov-FPS test errors at $N=50$ and $100$, out of $N_\text{tot} = 1,000$. The test errors are marginalized \textit{via} mean-binning onto the ST function x-axis. At $N=50$, GGFPS and FPS perform identically across the function space, while PCov-FPS poorly fits the function center. The corresponding $N=50$ subplots in SI Figures \ref{fig:SI_st_b_dependencies} and \ref{fig:SI_st_b_sweep_errors_vs_norms} show that GGFPS predictive performance improves as $B$ tends to $0$, indicating that FPS is the best option in the low-data regime.
However, at $N=100$, FPS fails to properly fit the steep function edges, while GGFPS outperforms both FPS and PCov-FPS across the function space. These results are expanded upon over the entire ST function surface in SI Figure \ref{fig:SI_st_error_heatmap}.

\FloatBarrier
\subsection{Sampling the MD17 Trajectories}\label{subsection:md17 results}

\begin{figure*}
    \centering
        \includegraphics[width=\linewidth]{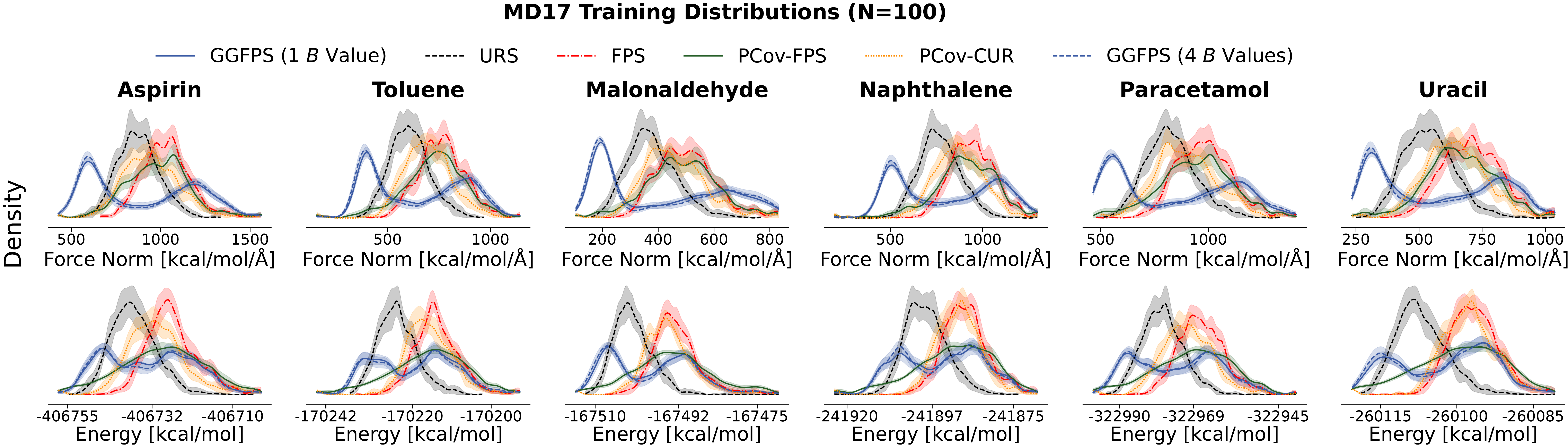}
        \caption{
        Force norm (top row) and energy (bottom row) training-set distributions for $N=100$ selected configurations across six MD17 molecules (columns). All panels show density histograms. Curves compare URS, FPS, PCov-FPS, PCov-CUR, GGFPS with a single $B$ value, and GGFPS with a four-$B$ hyperparameter search.
        }
    \label{fig:md17distributions}
\end{figure*}

While samples from the ST function are uniformly distributed in descriptor space, samples from chemical potential energy surfaces (PESs) are usually Boltzmann distributed.

Boltzmann distributed datasets introduce the additional difficulty in automated training set selection of adapting not only to the variance of the dataset labels, but also the variance of the dataset density.
Here we apply  URS, FPS, PCov-FPS, PCov-CUR, and GGFPS (single-$B$ and $4B$) to the MD17 aspirin, toluene, malonaldehyde, naphthalene, paracetamol, and uracil trajectories.\cite{chmiela2017machine} with training set sizes, $N$, ranging from 50 to 1{,}000 and labeled sets $N_\text{tot} = 10{,}000$ generated with sGDML\cite{Chmiela2019-xq} in order to mirror the property distributions of the entire trajectories (bootstrapped 30 times).

On the aspirin trajectory, GGFPS performance was insensitive to initialization strategy, schedule type (ascending/descending/alternating), and $B$ values across MAE, $B$-sensitivity, and force norm distributions (SI Figures~\ref{fig:SI_aspirin_init_schedules}, \ref{fig:SI_aspirin_init_strats_wrt_b}, \ref{fig:SI_aspirin_init_schedules_test_error_force_norms}, and \ref{fig:SI_aspirin_init_schedules_training_force_norm_distr}). We therefore fix probabilistic initialization (Eq.~\ref{eq:GGFPSprob1}) for all subsequent experiments.

The $B$ sweep range $[0,2]$ was selected via grid-search CV.
Using $\sim 4$ $B$ values yields a good compute/accuracy trade-off (SI Figures~\ref{fig:SI_best_mae_vs_ggfps_b_grid_size}, \ref{fig:SI_variance_vs_ggfps_g_grid_size}).
MAE versus force norm curves show poor performance at low $B$ for low-force configurations, with optimal single-$B$ performance at $B=1.55$ (SI Figure~\ref{fig:SImd17_b_sweep_errors_vs_force_norms}).
Comparing fixed $B=1.55$ against four $B$ values shows minimal test MAE gain but reduced variance (SI Figure~\ref{fig:SImd17_single_vs_multi_b}).
Ablating schedule strategies reveals the descending schedule delivers the most robust performance across $B$ and training set sizes for all molecules, especially at $B=1.55$ (SI Figure~\ref{fig:SImd17_beta_dependencies}).
This is because the descending schedule preferentially samples low-force configurations relative to ascending/alternating schedules (SI Figures~\ref{fig:SImd17_sched_error_vs_force_norms_4b}, \ref{fig:SImd17_sched_training_force_norm_distr_4b}), offsetting the FPS high force norm bias.

Timing benchmarks show GGFPS (with on-the-fly distances) is $\sim$10$\times$ faster than the standard FPS implementation and supervised PCov baselines on MD17 aspirin (SI Figures~\ref{fig:SI_md17_sampler_timing_by_n}--\ref{fig:SI_ggfps_timing_distributions}), while the $4B$ training/CV pipeline is only $2\times$ slower than single-$B$ (SI Figure~\ref{fig:SI_ggfps_eval_time_vs_n}). GGFPS timings are also contextualized within the broader learning workflow (SI Figure \ref{fig:SI_ggfps_timing_distributions}).

Figure \ref{fig:md17distributions} shows how different selection strategies redistribute sampling mass across force-norm (top row) and energy (bottom row) space for $N=100$ selected molecules. In addition to URS and FPS, we include supervised PCov-FPS/PCov-CUR baselines as well as GGFPS with a single $B$ value and with a four-$B$ hyperparameter search.

The FPS distributions bias towards both high energy and high force norm configurations across all molecules, under-sampling low to medium energy and force norm structures. Rather than simply under-sample the densest regions of the MD17 configuration space, FPS consistently `shifts' right, implying that the configurations which are `furthest' from each other in representation space correspond to strained structures.
This observation is consistent with the geometric behavior of PESs nearing dissociation, as the anharmonic tails `flatten out' over large distances. FPS will therefore preferentially sample these geometrically distant configurations, despite them being energetically rare. The supervised PCov baselines mitigate this effect by incorporating target information, while GGFPS explicitly uses force-norm information to ensure coverage of both low- and high-force-norm regimes.

Figure \ref{fig:md17distributions} shows that while PCov-CUR and PCov-FPS both sample more relaxed structures than FPS, GGFPS is unique in forming almost inverted Boltzmann training set distributions, under-sampling the densest regions of configuration space while over-representing the sparsest.

\begin{figure*}[!t]
    \centering
    
        \includegraphics[width=\linewidth]{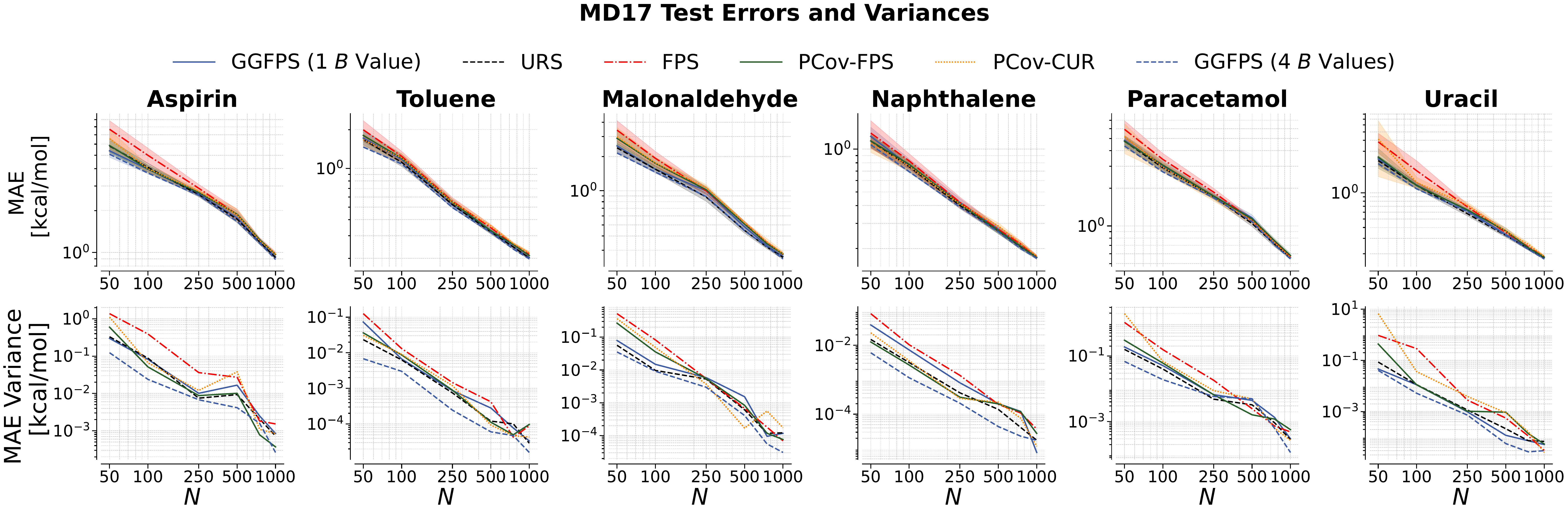}
    
    \caption{
    Top row: MD17 MAE learning curves (kcal/mol) with respect to training set size $N$ across MD17 molecules. Bottom row: corresponding MAE variances. Curves compare URS, FPS, PCov-FPS, PCov-CUR, GGFPS with a single $B$ value, and GGFPS with a four-$B$ hyperparameter search.}
    \label{fig:md17LCs}
\end{figure*}

Figure \ref{fig:md17LCs} shows the MAE learning curves (top row) and MAE variances (bottom row) for URS, FPS, PCov-FPS, PCov-CUR, and GGFPS. The results give credence to the observation that FPS poorly samples MD17 trajectories, as both the FPS mean predictive error and predictive variance are consistently higher than their URS counterparts. While the FPS mean errors all broadly converge with their corresponding URS errors at higher training set sizes, their variances do not. Additionally, at lower training set sizes, FPS predictive variances are up to an order of magnitude greater than URS predictive variances.
For completeness, we also report auxiliary PCov-FPS/PCov-CUR variants that incorporate \emph{force-label} information (rather than the standard energy-based definition) to provide a like-for-like supervised comparison (SI Figures~\ref{fig:SI_md17_pcov_force_labels_train_dists_n100} and \ref{fig:SI_md17_pcov_force_labels_lcs}.)

In contrast, the single-$B$ GGFPS learning curves show lower or equal predictive errors than the other sampling methods across all training set sizes for every MD17 molecule, while the $4B$ GGFPS MAEs are consistently the lowest.  Single-$B$ GGFPS models show lower or equal predictive variance compared to the other samplers, while $4B$ GGFPS has up to a 5-fold reduction in variance beyond its single-$B$ counterpart. While both PCov-FPS and PCov-CUR typically improve upon FPS, PCov-FPS shows more consistent MAE and MAE variance reductions.

It is intuitive that models trained on GGFPS training sets have lower predictive variances than their URS counterparts. Because the GGFPS training sets include more uncommon configurations, they have fewer catastrophic errors. Less intuitive is the fact that GGFPS training sets result in the same or lower mean prediction errors than URS training sets. The medium force norm configurations in the middle of the MD17 Boltzmann distributions comprise the vast majority of both the labeled and unlabeled (test) data. GGFPS drastically under-samples these configurations compared to URS, and (invoking the bias/variance tradeoff) one would expect a concurrent increase in GGFPS mean test error, even if the GGFPS test errors for rare configurations are, on average, lower.

\begin{figure*}
    \centering

        \includegraphics[width=0.7\linewidth]{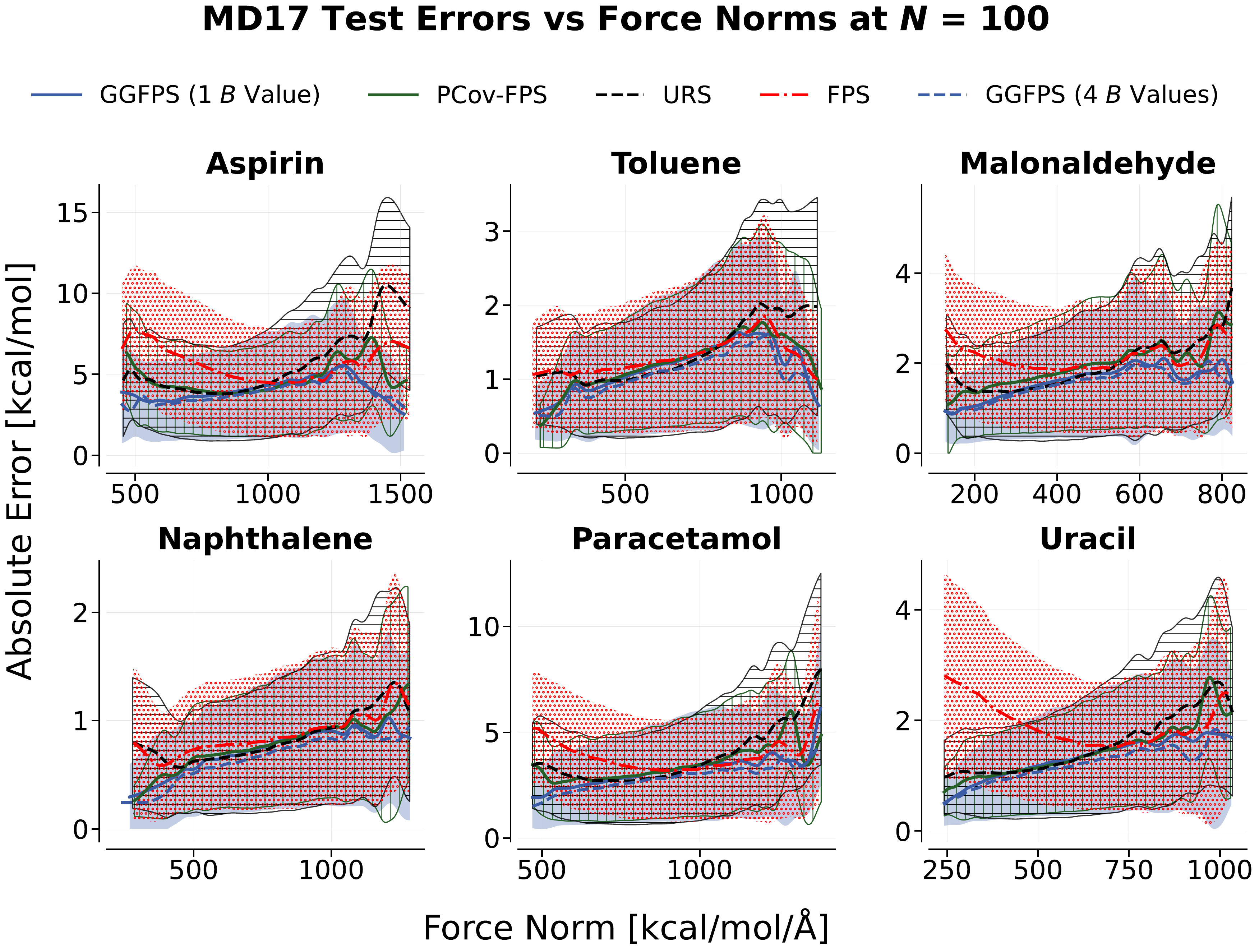}
        \caption{
        MD17 absolute test errors versus force norms at $N=100$ training configurations across MD17 molecules. Curves compare URS, FPS, PCov-FPS, PCov-CUR, and GGFPS variants.
        }
    \label{fig:MD17_test_errors}
\end{figure*}

We now show how these test errors are distributed across the configuration space in Figure \ref{fig:MD17_test_errors} by fixing the training set size to 100 configurations (other training set sizes are shown in the SI in Figures~\ref{fig:md17_test_errors_tss50}, \ref{fig:md17_test_errors_tss100}, \ref{fig:md17_test_errors_tss250}, \ref{fig:md17_test_errors_tss500}, \ref{fig:md17_test_errors_tss1000}, \ref{fig:md17_test_errors_tss50_force}, \ref{fig:md17_test_errors_tss100_force}, \ref{fig:md17_test_errors_tss250_force}, \ref{fig:md17_test_errors_tss500_force}, and \ref{fig:md17_test_errors_tss1000_force}). We bin the predicted test configurations with respect to their force norm values and plot the mean absolute errors and variances of the bins against their force norms. Each bin contains up to 30 configurations, with on average ($30 \text{ bootstraps} \times 10{,}000$ test configurations $\olddiv$ 30 configurations per bin) 10{,}000 bins per MD17 molecule. We omit PCov-CUR from these plots to reduce visual clutter because its performance is consistently worse than the other supervised sampling methods.
Importantly, the erratic behavior of  all methods at the upper end of the force norms comes from the sparsity of available samples in that regime (see Figure \ref{fig:md17distributions}, top right), meaning some spikes in error, despite the 50 bootstraps, correspond to only a couple dozen test error instances.

The URS predictive errors and variances are highest at force norm values for which there is limited labeled data. URS errors consistently increase for high force norm configurations, for every MD17 molecule.
The errors dip lower for the medium force norm configurations that form the bulk of the Boltzmann distribution. Naphthalene, whose trajectory includes many relaxed structures far from the bulk of the trajectory dataset, sees a large increase in test errors at low force norms. Conversely, uracil and toluene have low test errors and variances at low force norms, consistent with the paucity of low force norm structures far from the bulk datasets. The other molecules see a slight increase in test errors at low force norms.

FPS training sets result in dramatically worse predictive performance for low force norm configurations across most of MD17, with mean test errors and predictive variances up to twice as high as URS. The two exceptions to this behavior are naphthalene and toluene, which show a slightly higher variance, but the same mean error. One explanation is that naphthalene and toluene are more rigid and less functionalized than the other molecules, resulting in a higher number of similar low force norm configurations.

FPS and URS test errors and variances tend to converge for all systems at force norm values corresponding to the FPS training set distribution peaks in Figure \ref{fig:md17distributions}. Following the increased FPS coverage of high force norm configurations, FPS test error and variance are generally better than URS at higher force norms, with the exceptions of malonaldehyde and naphthalene.

GGFPS outperforms URS,  FPS, and PCov-FPS across all force norm values.
The most strained configuration energies are predicted on average twice as accurately with GGFPS training sets compared to URS. Also, GGFPS predictive variances for strained structures are on average 4 times smaller than URS variances, and twice as small as FPS variances.

The mean absolute GGFPS errors and error variances are slightly lower than their URS and FPS counterparts for medium force norm configurations, despite the paucity of medium force norm configurations sampled by GGFPS.
For toluene, malonaldehyde, and naphthalene, equilibrium and low force norm configurations are predicted on average twice as accurately and with half the variance with GGFPS training sets. GGFPS outperforms URS by an average factor of 1.5 for the remaining molecules.

$4B$ GGFPS outperforms single-$B$ GGFPS across all force norms, but the slightly lower errors around the medium force norm configurations are responsible for the dramatically lower MAE variances observed in Figure \ref{fig:md17LCs}, given the density of that region of the configuration space.

GGFPS consistently outperforms URS and FPS across training set sizes; the improvement is most distinct in the low data regime, before the configuration space becomes saturated. As shown in SI Figure \ref{fig:md17_test_errors_tss50}, at 50 training configurations GGFPS dramatically outperforms URS and FPS, with up to a fourfold decrease in test error for low and high force norm configurations compared to URS. GGFPS training sets still show a marked improvement in test error and variance over URS and FPS for high and low force norm configurations at 500 training points, as seen in Figure \ref{fig:md17_test_errors_tss500}. By 1{,}000 training points, while the GGFPS test errors broadly converge with the URS and FPS test errors, the GGFPS variances remain up to twice as small as URS variances for high force norm configurations.

Comparisons to PCov-FPS and PCov-CUR variants that incorporate force label information are shown in SI Figures~\ref{fig:md17_test_errors_tss50_force}, \ref{fig:md17_test_errors_tss100_force}, \ref{fig:md17_test_errors_tss250_force}, \ref{fig:md17_test_errors_tss500_force}, and \ref{fig:md17_test_errors_tss1000_force}.

FPS performs poorly on MD17 not just because it under-samples dense regions of the configuration spaces, but also because it systematically under-samples low force norm regions. In contrast, GGFPS also under-samples dense configuration space regions, but through proper coverage over all of the PES, it significantly outperforms FPS across all training set sizes and MD17 molecules.

%% file: sections/conclusions.tex
\section{Conclusions}
\label{section:conclusions}

We have introduced Gradient Guided Furthest Point Sampling (GGFPS), a supervised sampling method that combines gradient norms with furthest point sampling (FPS) to balance predictive accuracy and robustness.
Applied to the 2D Styblinski-Tang (ST) function and the MD17 molecular trajectories, GGFPS demonstrates superior performance over FPS and uniform random sampling (URS). For the ST function, GGFPS achieves the MAE of the full dataset with 50~\% fewer training points, and has on average half the test error of FPS for the same number of training points.

On the MD17 trajectories, GGFPS reduces MAE by factors of up to 2 (vs. FPS) and 3 (vs. URS) in high-force norm regions, critical for capturing transition states and strained configurations. At a given training set size, GGFPS also improves predictive accuracy for relaxed and equilibrium structures. GGFPS training sets of 50 configurations match the high-force-norm accuracy of URS and FPS training sets of up to 500 configurations. They also match the low-force-norm accuracy of URS and FPS training sets of up to 100 and 250 configurations, respectively.
The performance increase is most pronounced in the low-data regime, suitable for building fast models. However, significant benefits over FPS are observed up to a training set size of 750 configurations.

GGFPS lowers predictive variance by up to 50~\% compared to URS and up to an order of magnitude compared to FPS, ensuring robustness across Boltzmann-distributed and high-dimensional data.

By incorporating force/gradient-norm information and using a robust descending schedule with a single-$B$ default ($B=1.55$) or a small four-$B$ hyperparameter search when compute permits, GGFPS systematically reduces both mean errors and error variances while improving coverage of equilibrium (low-force) and strained (high-force) configurations. We also benchmark against supervised FPS-style selectors (PCov-FPS/PCov-CUR), and find that GGFPS remains competitive or superior while retaining a simple, single-pass selection procedure.

While we have successfully used GGFPS for interpolative learning relevant to molecular dynamics applications where the train and test data are sampled from the same PES, a key next step is extending it to extrapolative learning, for example across chemical compound space.
Another use of GGFPS could be to select diverse systems for generating synthetic data to fine-tune chemical foundation models.

%% file: sections/acknowledgements.tex
\section*{Acknowledgements} \label{sec:acknowledgements}
S.G. was supported by the Postdoc.Mobility fellowship by the Swiss National Science Foundation (project no. 225476).

%% file: sections/data_availability.tex
\section*{Data Availability}
GGFPS data are available at \url{https://doi.org/10.5281/zenodo.18653521}.

%% file: sections/algo_appendix.tex
\clearpage
\onecolumngrid
\section{Algorithms Appendix} \label{sec:algo_appendix}

\begin{algorithm}
\caption{Gradient-Guided Furthest Point Sampling (GGFPS) with selectable schedule.}
\label{salg:GGFPS_fullD}
\SetAlgoLined
\KwIn{
    $\mathbf{g} \in \mathbb{R}^{N_\text{tot}}$ (gradient norms),
    $\mathbf{D} \in \mathbb{R}^{N_\text{tot}\times N_\text{tot}}$ (distance matrix),
    $N$ (target subset size),
    $B>0$ (schedule bound; $\beta\in[-B,B]$),
    $\epsilon>0$,
    \texttt{schedule\_type} $\in \{\texttt{ascending},\texttt{descending},\texttt{alternating}\}$\;
}
\KwOut{
    $\mathcal{T} \subset \mathcal{L}$ (selected indices), $|\mathcal{T}|=N$\;
}

\BlankLine
\textbf{Schedule definition:} for step $k\in\{0,\dots,N-1\}$, set $\beta_k \gets \textsc{Schedule}(k,N,B,\texttt{schedule\_type})$, where\;
\[
\textsc{Schedule}(k,N,B,\texttt{ascending}) =
\begin{cases}
-B + 2B\,\dfrac{k}{N-1}, & N>1\\[6pt]
+B, & N=1
\end{cases}
\]
\[
\textsc{Schedule}(k,N,B,\texttt{descending}) =
\begin{cases}
+B - 2B\,\dfrac{k}{N-1}, & N>1\\[6pt]
+B, & N=1
\end{cases}
\]
\[
\textsc{Schedule}(k,N,B,\texttt{alternating}) =
\begin{cases}
-B + \dfrac{kB}{N-1}, & N>1\ \text{and } k \text{ even}\\[8pt]
\;\;\,B - \dfrac{(k-1)B}{N-1}, & N>1\ \text{and } k \text{ odd}\\[8pt]
+B, & N=1
\end{cases}
\]
\BlankLine

Initialize $\mathcal{T} \gets \emptyset$, $\mathcal{A} \gets \mathcal{L}$, and $d_j \gets \infty\ \forall j\in\mathcal{L}$\;

\BlankLine
\textbf{Initialization (probabilistic):}\;
$\beta_0 \gets \textsc{Schedule}(0,N,B,\texttt{schedule\_type})$\;
Compute $p_j = (g_j+\epsilon)^{\beta_0}\Big/\sum_{\ell\in\mathcal{L}}(g_\ell+\epsilon)^{\beta_0}\ \forall j\in\mathcal{L}$\;
Sample $c \sim \operatorname{Categorical}(\mathbf{p})$\;
Update $\mathcal{T}\gets\mathcal{T}\cup\{c\}$, $\mathcal{A}\gets\mathcal{A}\setminus\{c\}$\;
Initialize $d_j \gets D_{c j}\ \forall j\in\mathcal{A}$\;

\BlankLine
\For{$k = 1$ \KwTo $N-1$}{
    $\beta_k \gets \textsc{Schedule}(k,N,B,\texttt{schedule\_type})$\;
    \ForEach{$j \in \mathcal{A}$}{
        $s_j \gets (g_j+\epsilon)^{\beta_k}\, d_j$\;
    }
    $c \gets \operatorname{argmax}_{j\in\mathcal{A}} s_j$\;
    Update $\mathcal{T}\gets\mathcal{T}\cup\{c\}$, $\mathcal{A}\gets\mathcal{A}\setminus\{c\}$\;
    \ForEach{$j \in \mathcal{A}$}{
        $d_j \gets \min\!\left(d_j, D_{c j}\right)$\;
    }
}
\Return $\mathcal{T}$\;
\end{algorithm}

\FloatBarrier

\begin{algorithm}
\caption{Gradient-Guided Furthest Point Sampling (GGFPS) with on-the-fly distances.}
\label{salg:GGFPS_on_the_fly}
\SetAlgoLined
\KwIn{
    $\mathbf{g} \in \mathbb{R}^{N_\text{tot}}$ (gradient norms),
    $\mathbf{X} = [\mathbf{x}_1^\top,\dots,\mathbf{x}_{N_\text{tot}}^\top]^\top \in \mathbb{R}^{N_\text{tot}\times d}$ (descriptors/points),
    $N$ (target subset size),
    $B>0$ (schedule bound; $\beta\in[-B,B]$),
    $\epsilon>0$,
    \texttt{schedule\_type} $\in \{\texttt{ascending},\texttt{descending},\texttt{alternating}\}$\;
}
\KwOut{
    $\mathcal{T} \subset \mathcal{L}$ (selected indices), $|\mathcal{T}|=N$\;
}

\BlankLine
\textbf{Schedule definition:} for step $k\in\{0,\dots,N-1\}$, set $\beta_k \gets \textsc{Schedule}(k,N,B,\texttt{schedule\_type})$, where\;
\[
\textsc{Schedule}(k,N,B,\texttt{ascending}) =
\begin{cases}
-B + 2B\,\dfrac{k}{N-1}, & N>1\\[6pt]
+B, & N=1
\end{cases}
\]
\[
\textsc{Schedule}(k,N,B,\texttt{descending}) =
\begin{cases}
+B - 2B\,\dfrac{k}{N-1}, & N>1\\[6pt]
+B, & N=1
\end{cases}
\]
\[
\textsc{Schedule}(k,N,B,\texttt{alternating}) =
\begin{cases}
-B + \dfrac{kB}{N-1}, & N>1\ \text{and } k \text{ even}\\[6pt]
\;\;\,B - \dfrac{(k-1)B}{N-1}, & N>1\ \text{and } k \text{ odd}\\[6pt]
+B, & N=1
\end{cases}
\]

\BlankLine
Initialize $\mathcal{T}\gets\emptyset$, $\mathcal{A}\gets\mathcal{L}$, and $d_j\gets\infty\ \forall j\in\mathcal{L}$\;

Precompute squared norms $n_j \gets \|\mathbf{x}_j\|_2^2\ \forall j\in\mathcal{L}$\;

\BlankLine
\textbf{Initialization (probabilistic):}\;
$\beta_0 \gets \textsc{Schedule}(0,N,B,\texttt{schedule\_type})$\;
Compute $p_j = (g_j+\epsilon)^{\beta_0}\Big/\sum_{\ell\in\mathcal{L}}(g_\ell+\epsilon)^{\beta_0}\ \forall j\in\mathcal{L}$\;
Sample $c \sim \operatorname{Categorical}(\mathbf{p})$\;
Update $\mathcal{T}\gets\mathcal{T}\cup\{c\}$, $\mathcal{A}\gets\mathcal{A}\setminus\{c\}$\;

Compute distances from $c$ to all points using
\[
\delta_j^2 \gets n_j + n_c - 2\,\mathbf{x}_j^\top \mathbf{x}_c \quad \forall j\in\mathcal{L},
\qquad
\delta_j \gets \sqrt{\max(\delta_j^2,0)}.
\]
Update $d_j \gets \min(d_j,\delta_j)\ \forall j\in\mathcal{L}$ and set $d_c\gets 0$\;

\BlankLine
\For{$k = 1$ \KwTo $N-1$}{
    $\beta_k \gets \textsc{Schedule}(k,N,B,\texttt{schedule\_type})$\;

    \ForEach{$j \in \mathcal{A}$}{
        $s_j \gets (g_j+\epsilon)^{\beta_k}\, d_j$\;
    }
    $c \gets \operatorname{argmax}_{j\in\mathcal{A}} s_j$\;
    Update $\mathcal{T}\gets\mathcal{T}\cup\{c\}$, $\mathcal{A}\gets\mathcal{A}\setminus\{c\}$\;

    Compute distances from $c$ to all points:
    \[
    \delta_j^2 \gets n_j + n_c - 2\,\mathbf{x}_j^\top \mathbf{x}_c \quad \forall j\in\mathcal{L},
    \qquad
    \delta_j \gets \sqrt{\max(\delta_j^2,0)}.
    \]
    Update $d_j \gets \min(d_j,\delta_j)\ \forall j\in\mathcal{L}$ and set $d_c\gets 0$\;
}
\Return $\mathcal{T}$\;
\end{algorithm}

%% file: sections/appendix1.tex
\clearpage
\onecolumngrid
\floatplacement{figure}{!hbp}
\newif\ifSIFirstSubsection
\SIFirstSubsectiontrue
\newcommand{\SISubsection}[2]{%
    \FloatBarrier
    \ifSIFirstSubsection
        \global\SIFirstSubsectionfalse
    \else
        \clearpage
    \fi
    \subsection{#1}%
    \label{#2}%
    \suppressfloats[t]%
}
\section{Appendix} \label{sec:appendix}

\SISubsection{GGFPS Methods Plots}{sec:SI_methods}

\begin{figure}
    \centering
    \includegraphics[width=0.5\linewidth]{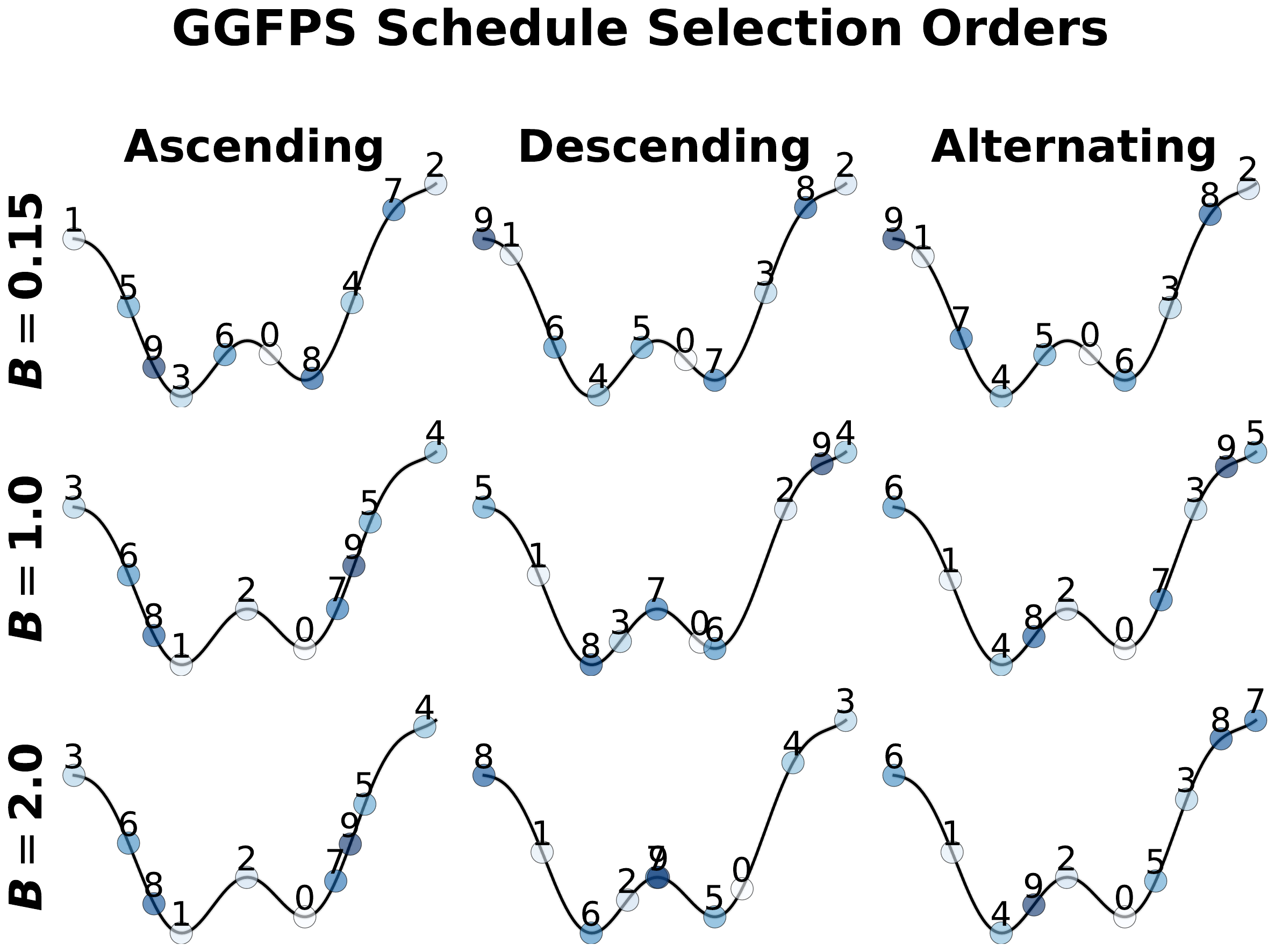}
    \caption{\added{Illustration of how the GGFPS $\beta$ schedule affects the \emph{order} of sampling on a 1D dummy double-well function. Rows correspond to sweep ranges $B=0.15$, $B=1$, and $B=2$ (top to bottom). Columns correspond to schedule strategies (ascending, descending, alternating). Numbers next to selected points indicate the selection order under the corresponding schedule.}}
    \label{fig:SI_ggfps_beta_1d_schedules}
\end{figure}

\SISubsection{Styblinski--Tang (ST) additional ablations}{sec:SI_st_additional}

\begin{figure}
    \centering
    \includegraphics[width=0.98\linewidth]{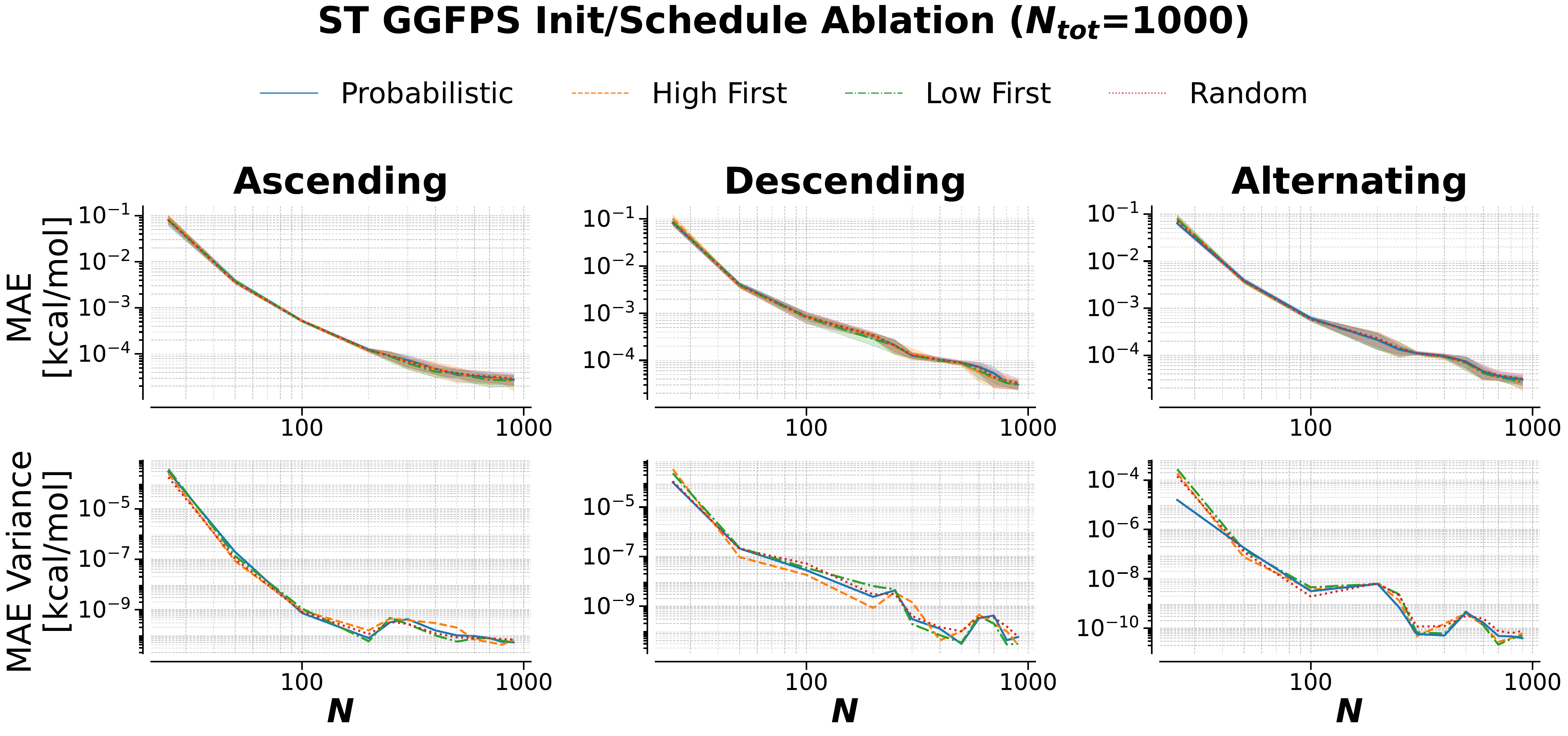}
    \caption{\added{ST: initialization $\times$ schedule ablation at $N_\mathrm{tot}=1000$. Columns correspond to schedule strategies (ascending, descending, alternating), and colors compare initialization strategies (probabilistic, high-first, low-first, random). The figure reports both MAE and MAE variance across training set sizes.}}
    \label{fig:SI_st_init_schedule_ablation}
\end{figure}

\begin{figure}
    \centering
    \includegraphics[width=0.98\linewidth]{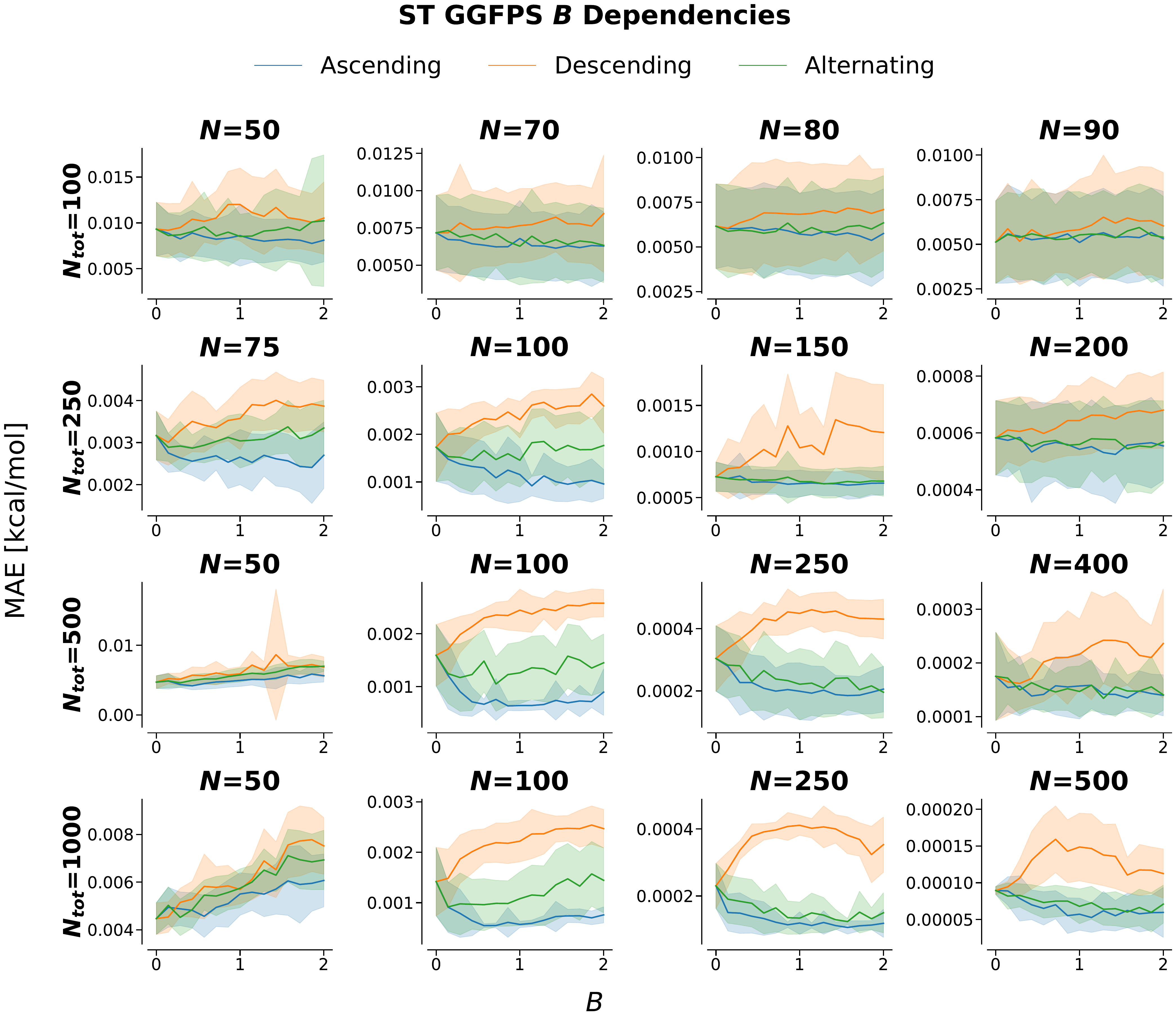}
    \caption{\added{ST: $B$-dependence of performance across schedule strategies and dataset sizes. Panels report MAE as a function of $B$ for multiple training set sizes $N$ and labeled set sizes $N_\mathrm{tot}$, comparing ascending, descending, and alternating schedules.}}
    \label{fig:SI_st_b_dependencies}
\end{figure}

\begin{figure}
    \centering
    \includegraphics[width=0.7\linewidth]{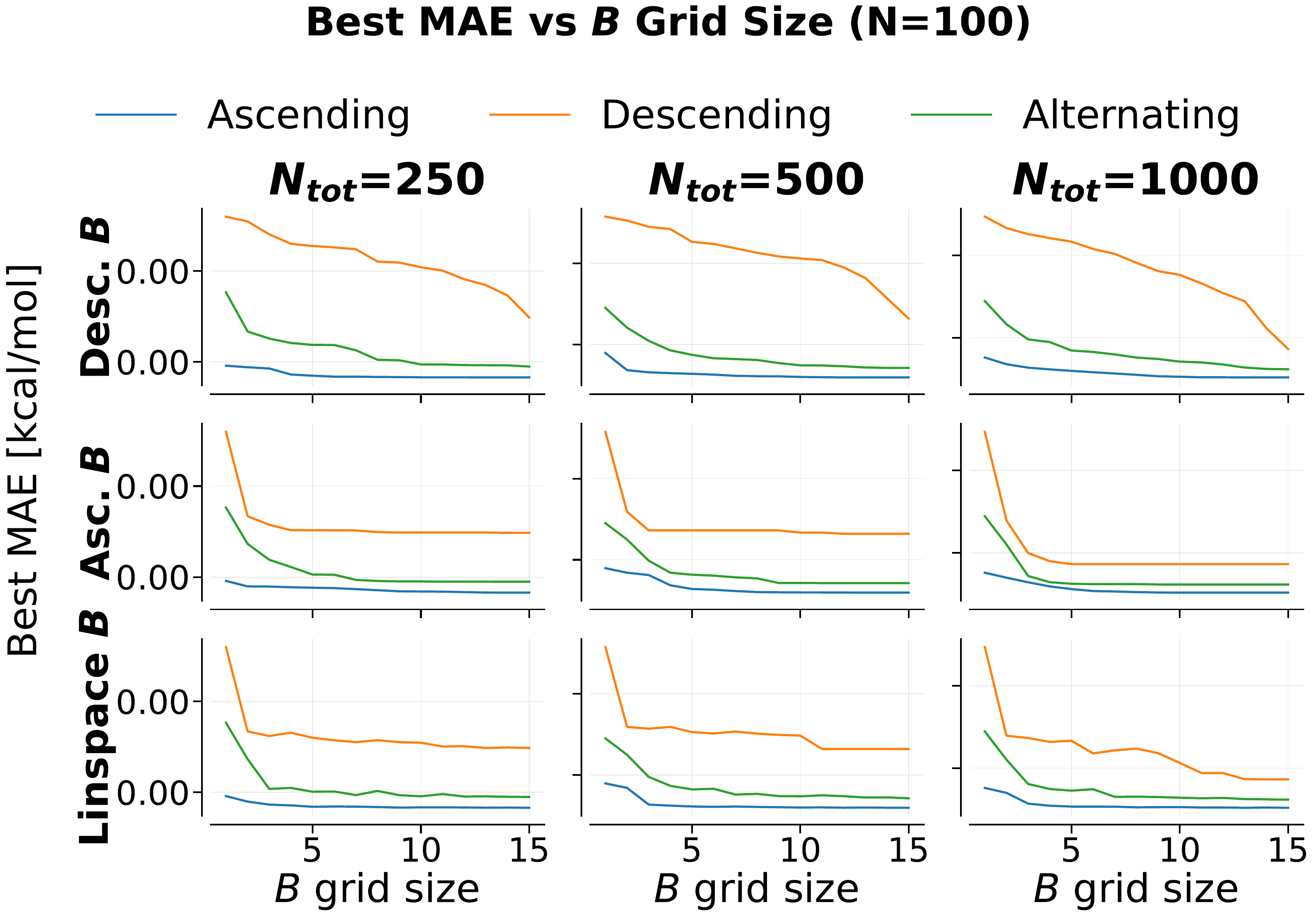}
    \caption{\added{ST: ablation of the $B$-grid size used when treating $B$ as a hyperparameter, measured via best MAE (here at $N=100$). Curves show how performance approaches an asymptote as the number of candidate $B$ values increases, across labeled set sizes $N_\mathrm{tot}$ and schedule strategies. Descending $B$ refers to adding $B$ values starting at $B=2.0$ and moving downward, ascending $B$ means the opposite, and linspace $B$ refers to building an even grid of $B$ values between $0$ and $2$, starting at $B=2.0$.}}
    \label{fig:SI_st_b_grid_density_mae}
\end{figure}

\begin{figure}
    \centering
    \includegraphics[width=0.7\linewidth]{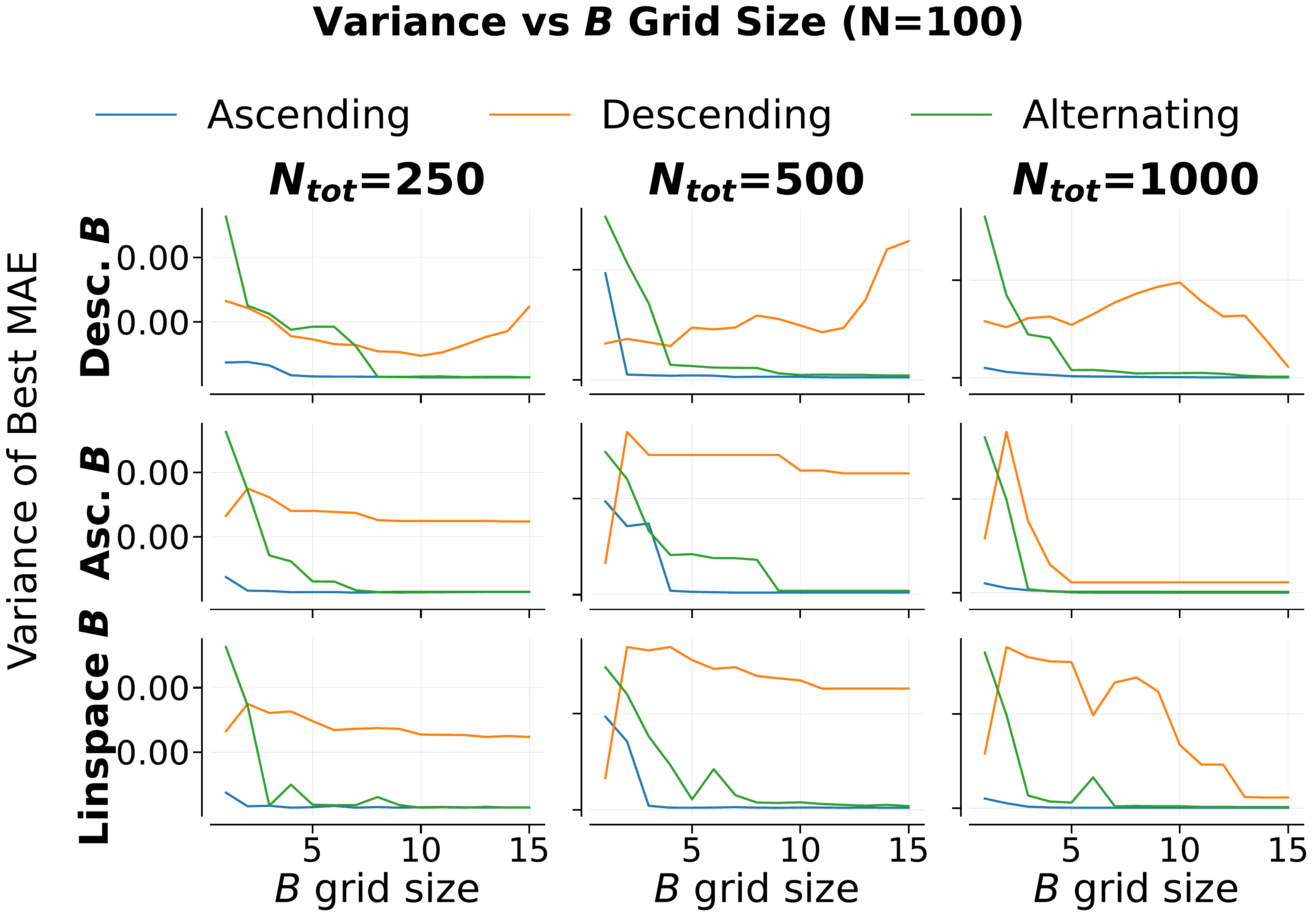}
    \caption{\added{ST: ablation of the $B$-grid size used when treating $B$ as a hyperparameter, measured via best variance (here at $N=100$). Curves show how performance approaches an asymptote as the number of candidate $B$ values increases, across labeled set sizes $N_\mathrm{tot}$ and schedule strategies. Descending $B$ refers to adding $B$ values starting at $B=2.0$ and moving downward, ascending $B$ means the opposite, and linspace $B$ refers to building an even grid of $B$ values between $0$ and $2$, starting at $B=2.0$.}}
    \label{fig:SI_st_b_grid_density_variance}
\end{figure}

\begin{figure}
    \centering
    \includegraphics[width=0.98\linewidth]{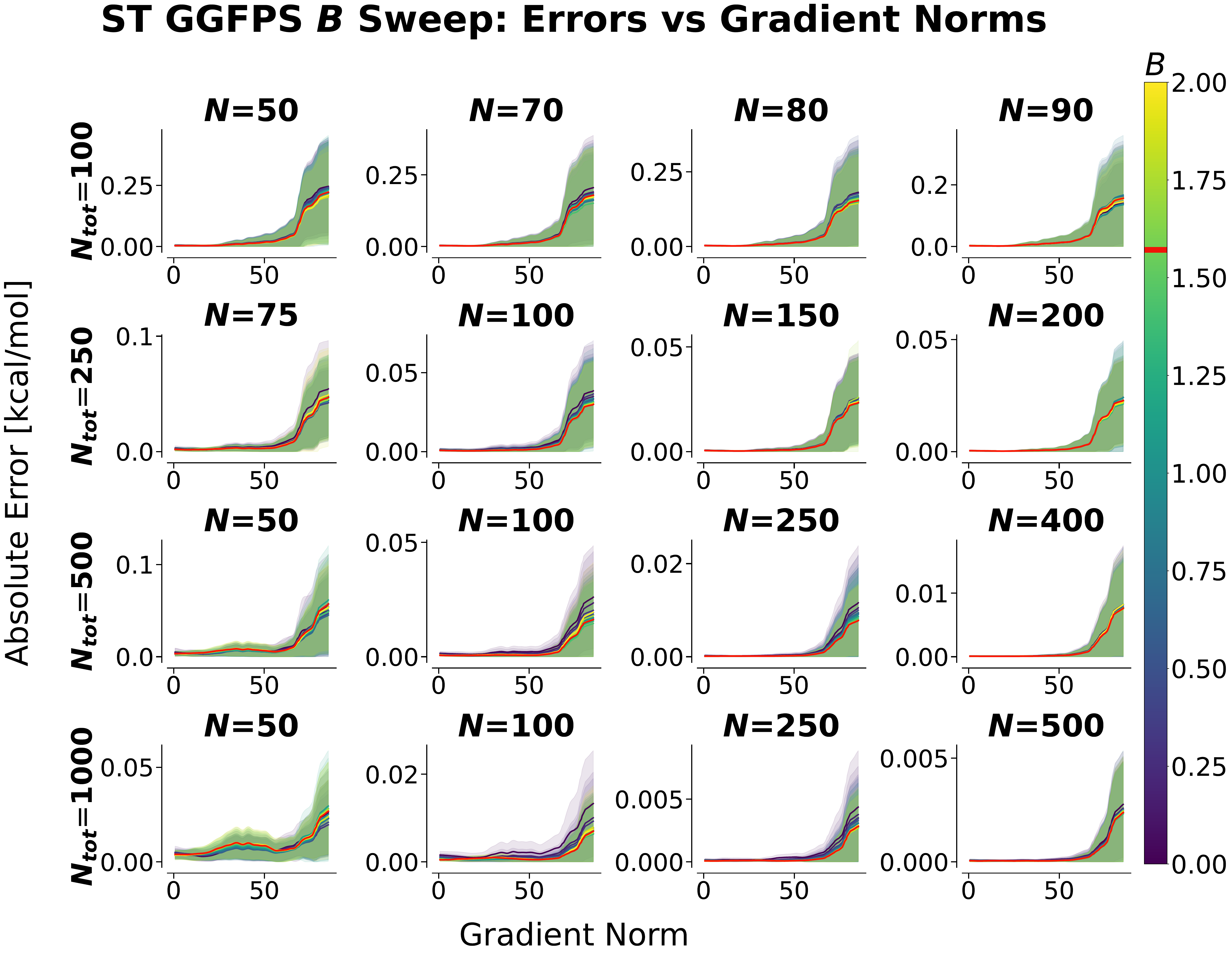}
    \caption{\added{ST: sweep over the gradient-bias range parameter $B$, reporting absolute test error versus gradient norm for multiple training set sizes $N$ (per column) and labeled set sizes $N_\mathrm{tot}$ (per row). The optimal $B$ value is indicated in red, while the others follow a color gradient as indicated by the color map on the right, from blue to yellow.}}
    \label{fig:SI_st_b_sweep_errors_vs_norms}
\end{figure}

\begin{figure}
    \centering
    \includegraphics[width=0.98\linewidth]{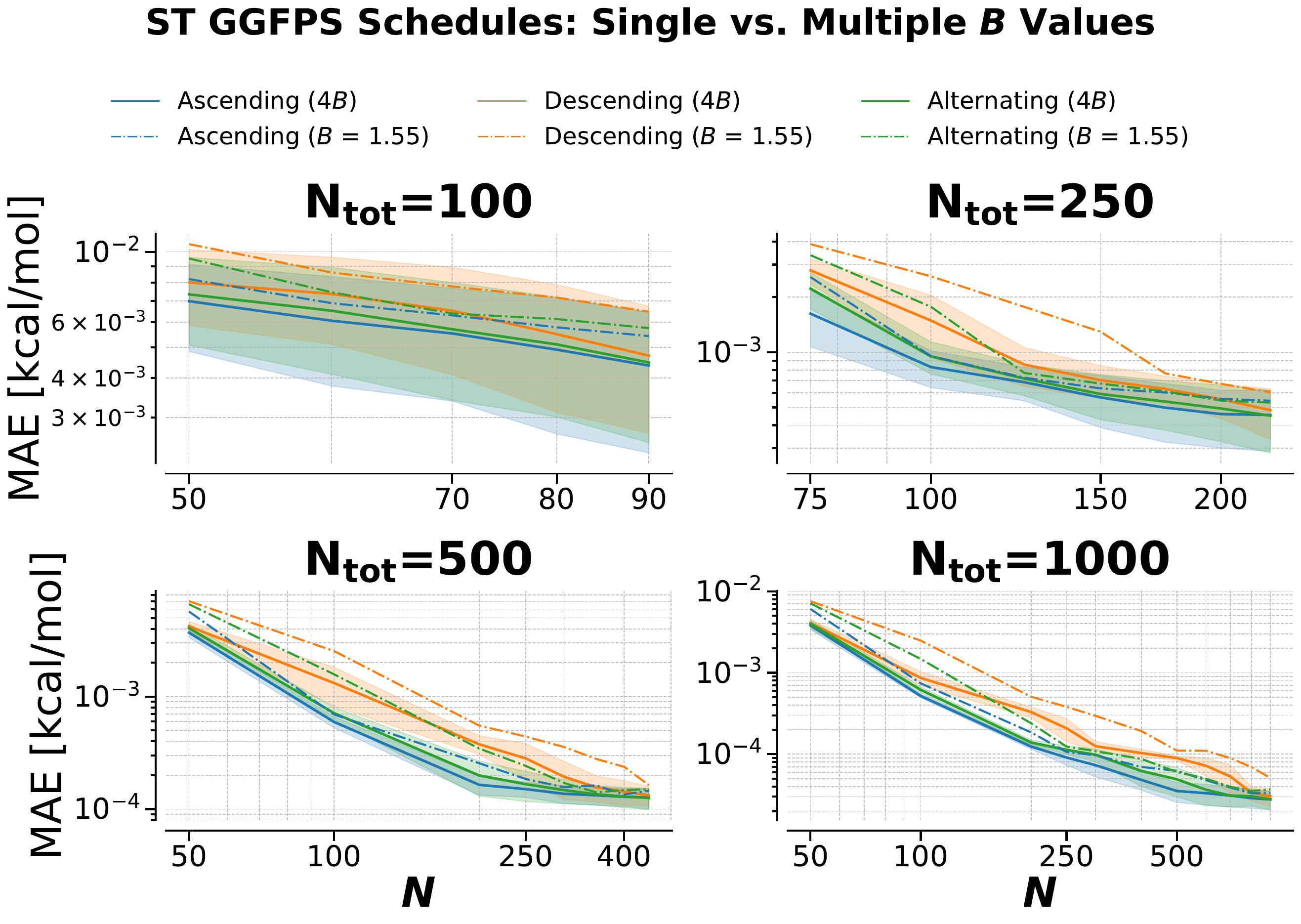}
    \caption{\added{ST: learning curves comparing a fixed single-$B$ setting versus treating $B$ as a hyperparameter (multiple candidate $B$ values), across schedule strategies and labeled set sizes $N_\mathrm{tot}$.}}
    \label{fig:SI_st_single_vs_multi_b}
\end{figure}

\begin{figure}
    \centering
    \includegraphics[width=0.98\linewidth]{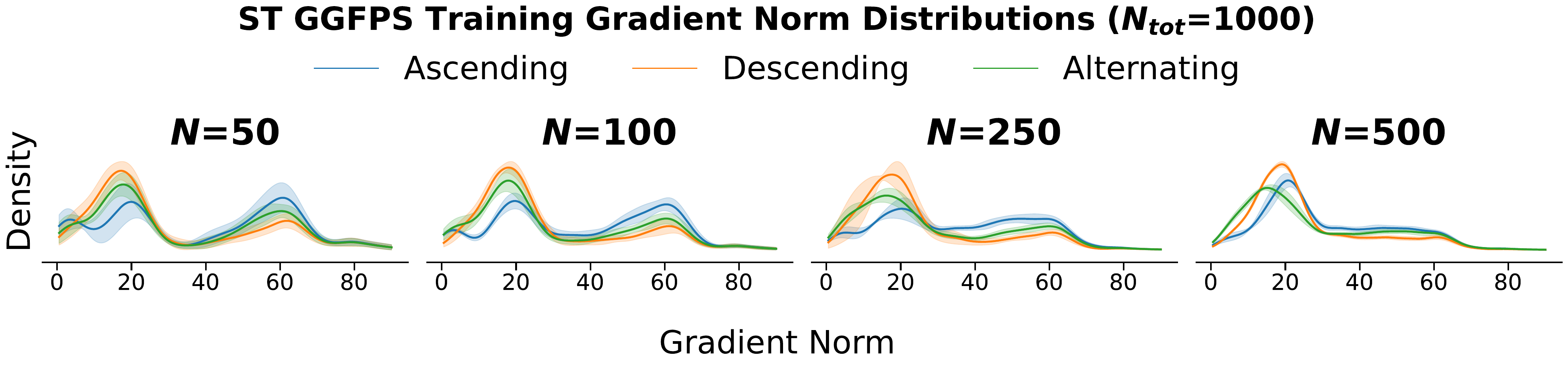}
    \caption{\added{ST: training gradient-norm distributions induced by different schedules (ascending, descending, alternating) at $N_\mathrm{tot}=1000$ and $B=1.55$, shown for multiple selected training set sizes $N$.}}
    \label{fig:SI_st_train_grad_norm_dists}
\end{figure}

\begin{figure}
    \centering
    \includegraphics[width=0.5\linewidth]{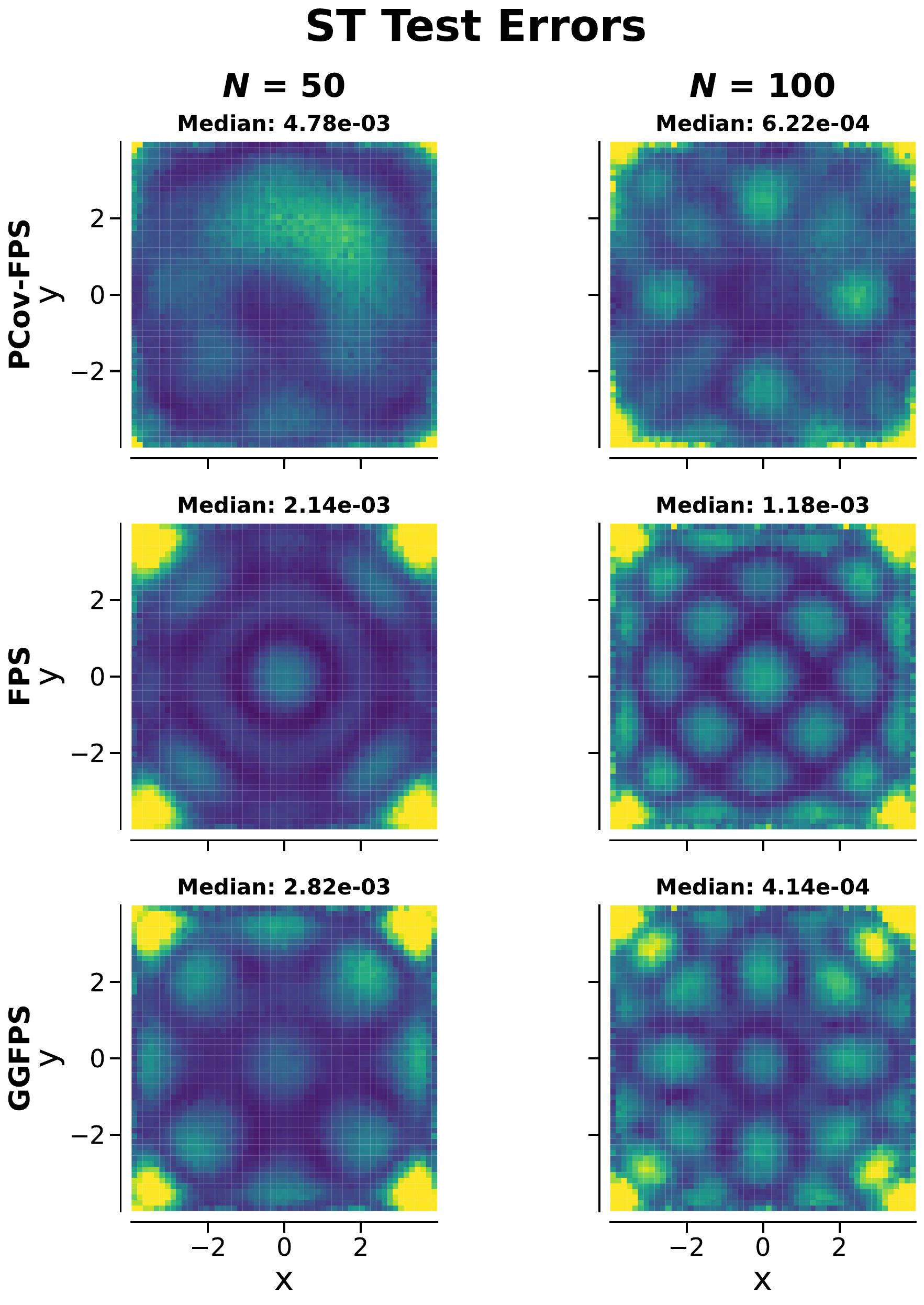}
    \caption{
    \added{Scatter plots of the 2D ST function absolute test errors corresponding to models trained on FPS (top row) and GGFPS (bottom row), with training set sizes, $N = 50$ (left column) and $N = 100$ (right column). 
    All models start with an initial labeled set of 1{,}000 points.
    Median absolute test error values are shown above each scatter plot.}}
    \label{fig:SI_st_error_heatmap}
\end{figure}

\SISubsection{MD17 Aspirin GGFPS Initialization Strategies}{sec:SI_Aspirin_Inits}

\begin{figure}
    \centering
    \includegraphics[width=0.9\linewidth]{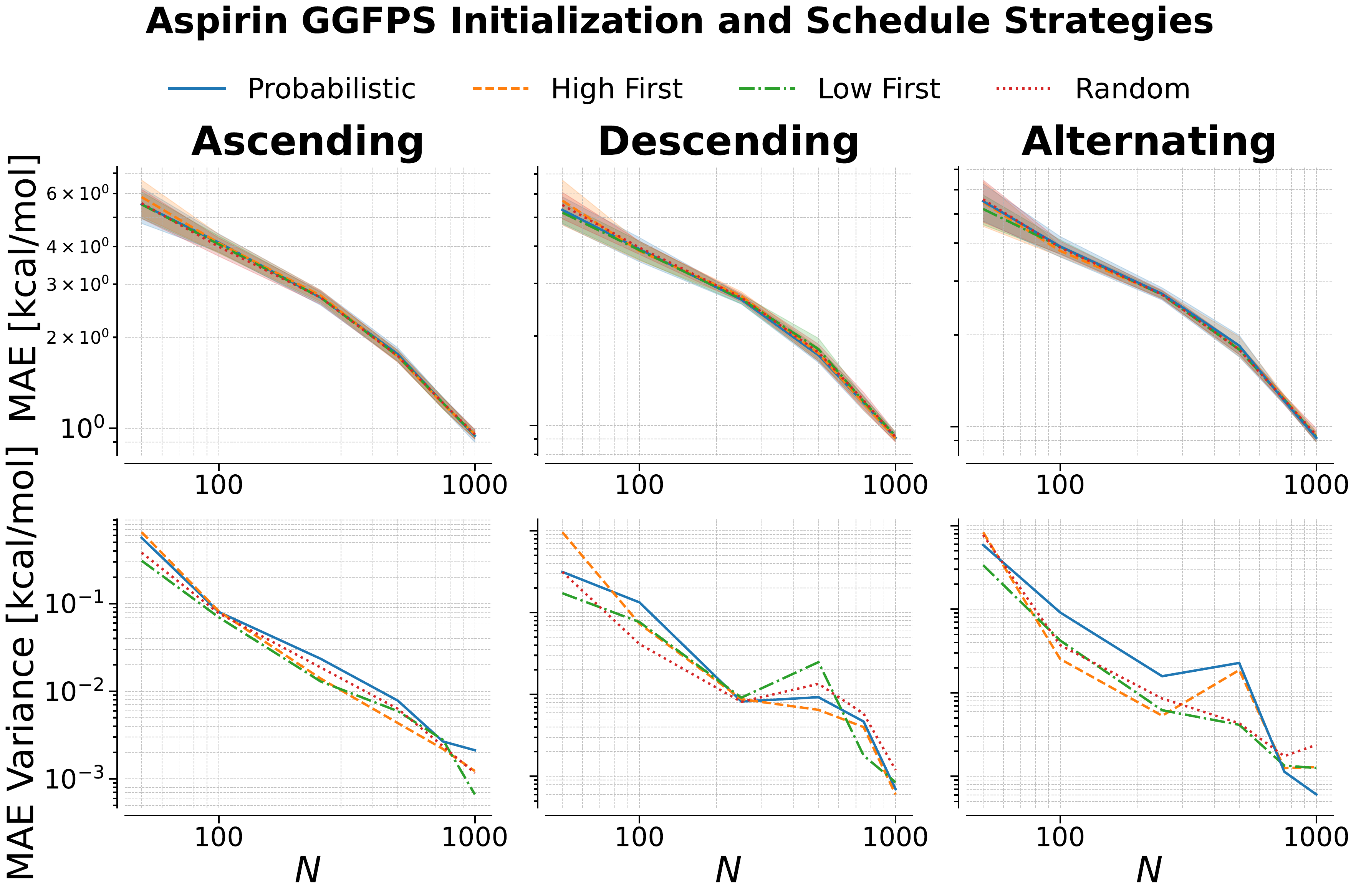}
    \caption{\added{MD17 aspirin Initialization and schedule ablation for GGFPS over a hyperparameter search of 4 different $B$ values. Columns correspond to training set sizes $N \in \{50, 100, 500, 1000\}$, rows correspond to schedule strategies (ascending, descending, alternating), and colors compare four initialization strategies (probabilistic, high-first, low-first, random). The figure reports both MAE and MAE variance (as indicated in the panels).}}
    \label{fig:SI_aspirin_init_schedules}
\end{figure}

\begin{figure}
    \centering
    \includegraphics[width=0.9\linewidth]{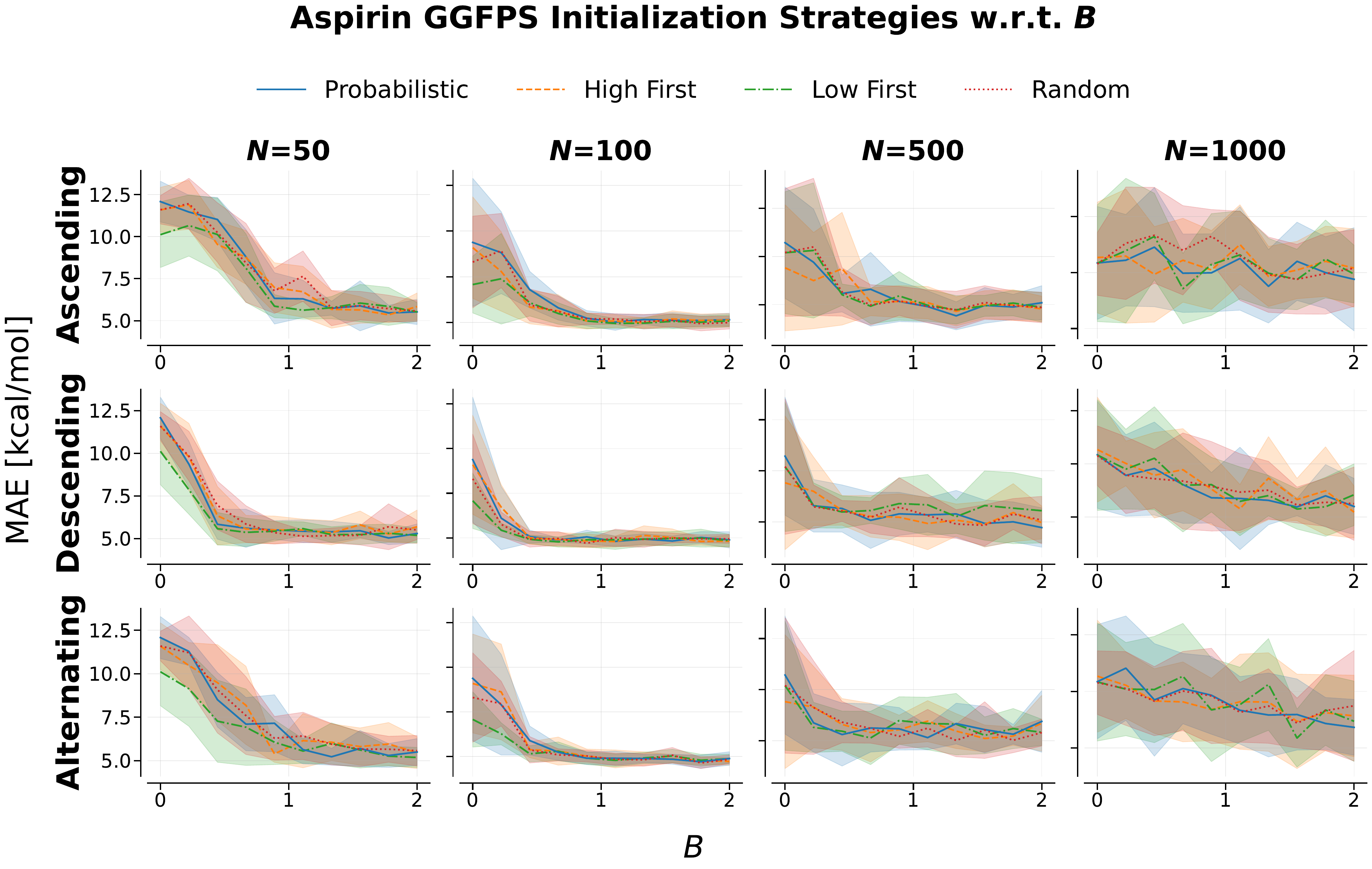}
    \caption{\added{MD17 aspirin GGFPS performance as a function of the gradient-bias range parameter $B$ (x-axis). $B$ constrains the schedule values via $\beta \in [-B, B]$, with $B=0$ recovering FPS. Columns correspond to training set sizes $N \in \{50, 100, 500, 1000\}$, rows correspond to schedule strategies (ascending, descending, alternating), and colors compare four initialization strategies (probabilistic, high-first, low-first, random). Performance improves from $B=0$ and saturates for $B$ in the $1$–$2$ range, motivating the fixed single-$B$ choice $B=1.55$ used in subsequent MD17 experiments.}}
    \label{fig:SI_aspirin_init_strats_wrt_b}
\end{figure}

\begin{figure}
    \centering
    \includegraphics[width=0.95\linewidth]{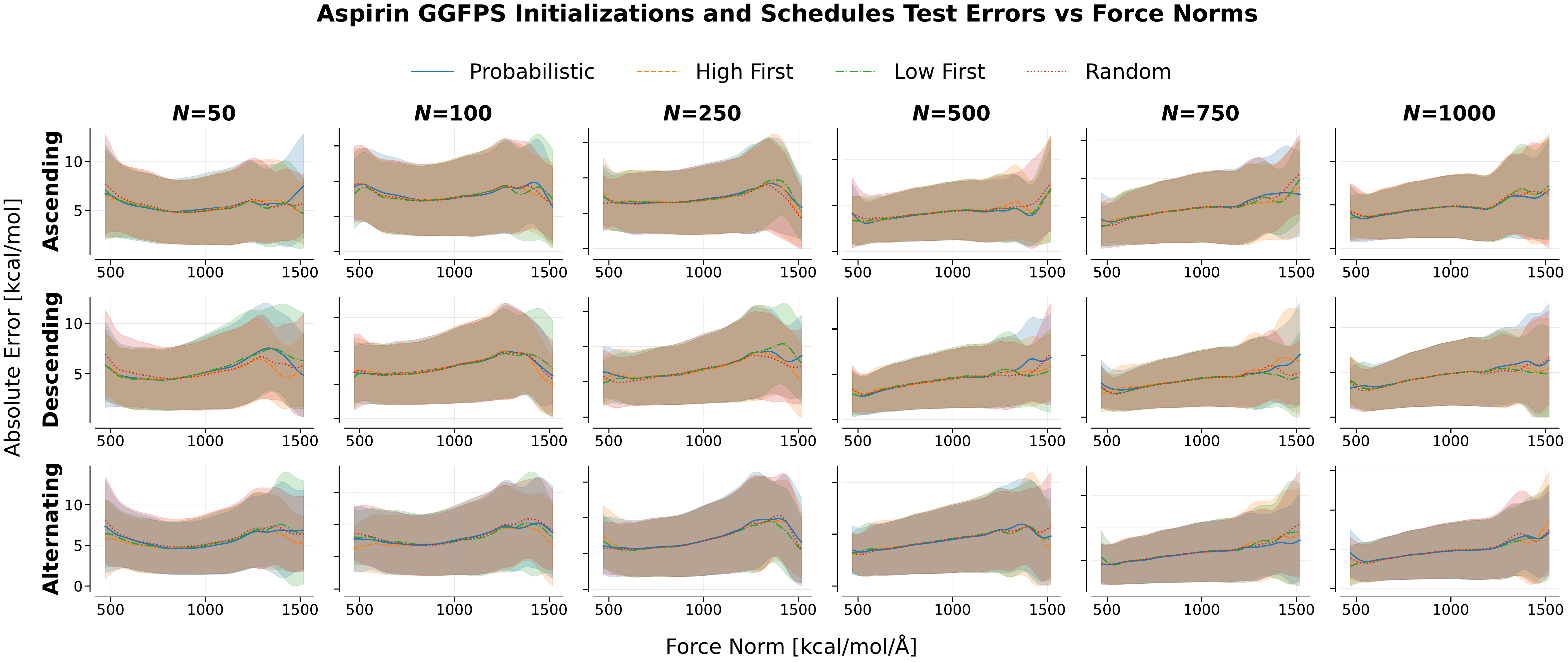}
    \caption{\added{MD17 aspirin absolute test error versus force norm for GGFPS under different schedule and initialization strategies. Rows correspond to schedule strategies (ascending, descending, alternating), columns correspond to training set sizes $N \in \{50, 100, 250, 500, 750, 1000\}$, and colors compare four initialization strategies (probabilistic, high-first, low-first, random). Across all schedules and training set sizes, the curves are very similar, indicating that initialization has little effect on the error-vs-force-norm behavior.}}
    \label{fig:SI_aspirin_init_schedules_test_error_force_norms}
\end{figure}

\begin{figure}
    \centering
    \includegraphics[width=0.95\linewidth]{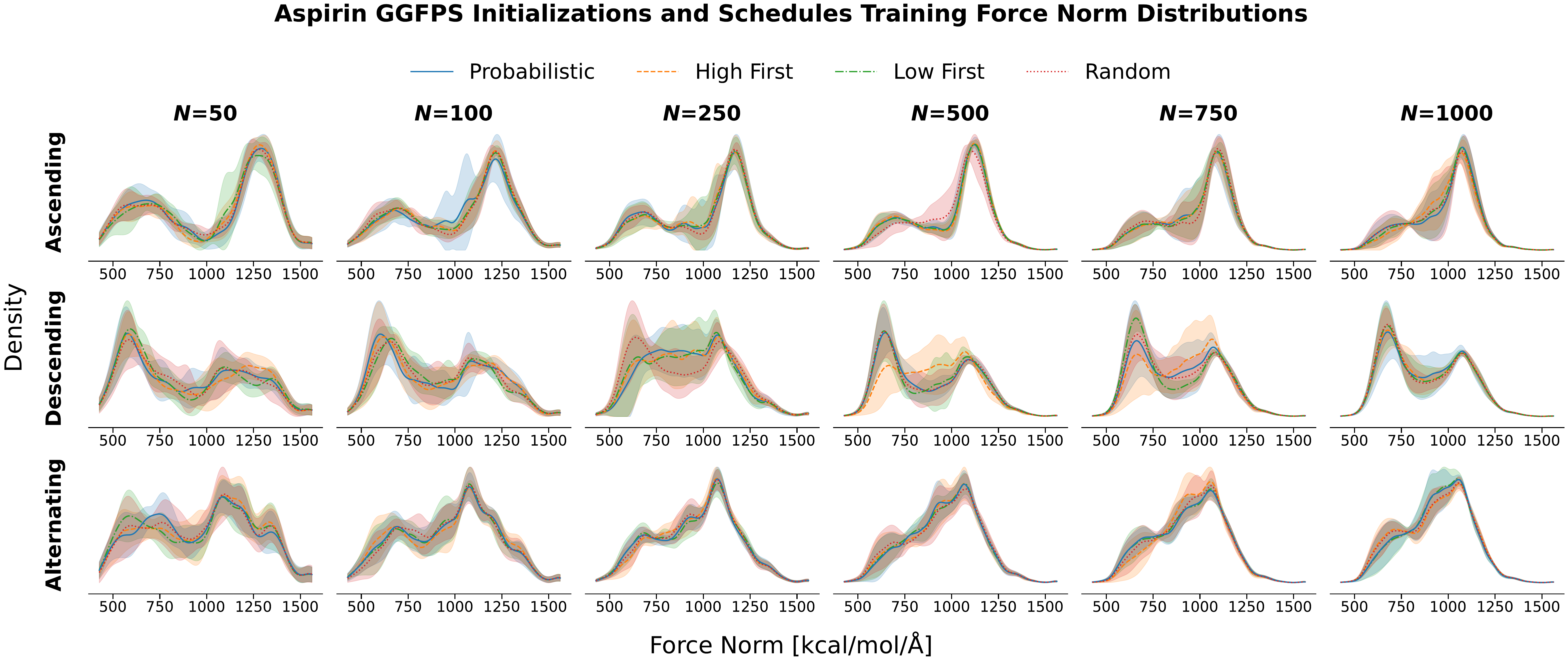}
    \caption{\added{MD17 aspirin training force-norm distributions of GGFPS-selected training sets under different schedule and initialization strategies. Rows correspond to schedule strategies (ascending, descending, alternating), columns correspond to training set sizes $N \in \{50, 100, 250, 500, 750, 1000\}$, and colors compare four initialization strategies (probabilistic, high-first, low-first, random). The resulting distributions are nearly unchanged across initialization choices for all schedules and training set sizes, demonstrating that initialization does not materially affect the induced training-set distribution.}}
    \label{fig:SI_aspirin_init_schedules_training_force_norm_distr}
\end{figure}

\SISubsection{MD17 Aspirin GGFPS Schedules}{sec:SI_MD17_schedules}

\begin{figure}
    \centering
    \includegraphics[width=0.95\linewidth]{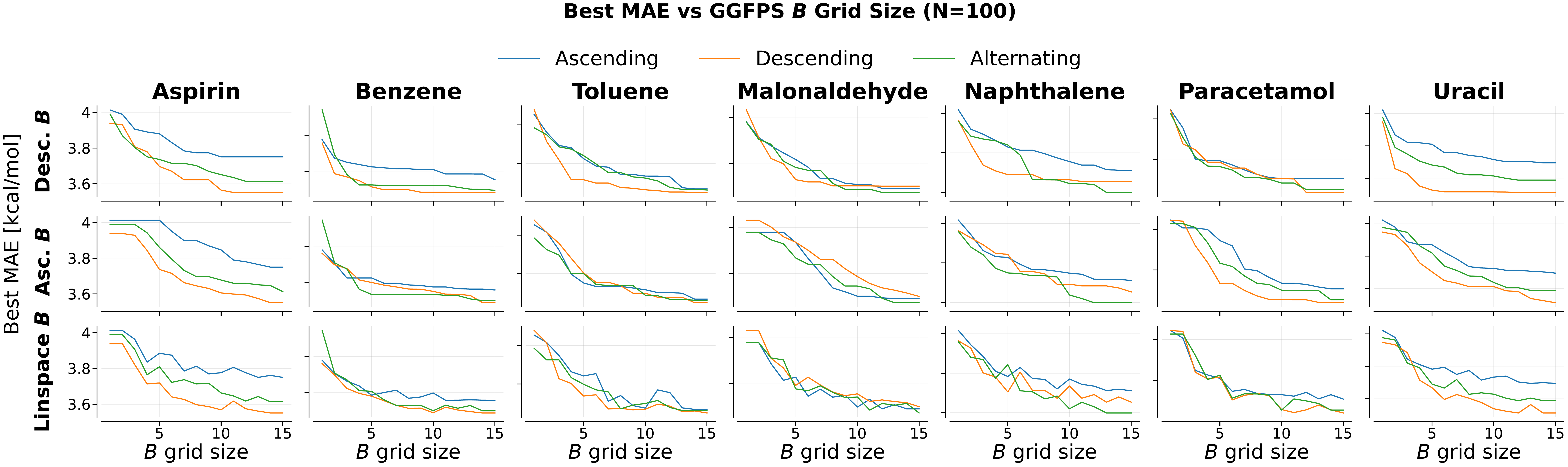}
    \caption{\added{Ablation of how many $B$ values are needed when treating $B$ as a hyperparameter. The plot reports the best MAE as a function of the $B$-grid size (number of candidate $B$ values evaluated). Descending $B$ refers to adding $B$ values starting at $B=2.0$ and moving downward, ascending $B$ means the opposite, and linspace $B$ refers to building an even grid of $B$ values between $0$ and $2$, starting at $B=2.0$. Performance approaches an asymptote by $\sim 4$ $B$ values, motivating the use of four $B$ values in compute-constrained hyperparameter searches.}}
    \label{fig:SI_best_mae_vs_ggfps_b_grid_size}
\end{figure}

\begin{figure}
    \centering
    \includegraphics[width=0.95\linewidth]{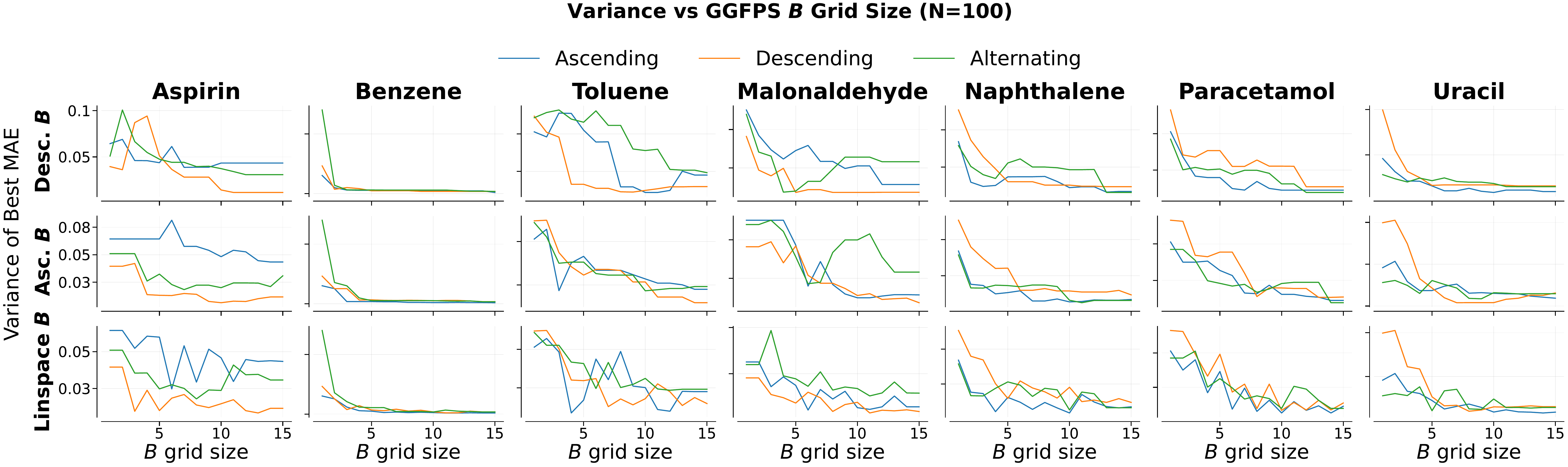}
    \caption{\added{Ablation of how many $B$ values are needed when treating $B$ as a hyperparameter, measured via predictive variance. The plot reports the variance metric as a function of the $B$-grid size (number of candidate $B$ values evaluated). Descending $B$ refers to adding $B$ values starting at $B=2.0$ and moving downward, ascending $B$ means the opposite, and linspace $B$ refers to building an even grid of $B$ values between $0$ and $2$, starting at $B=2.0$. Variance performance similarly approaches an asymptote by $\sim 4$ $B$ values, corroborating the four-$B$ choice.}}
    \label{fig:SI_variance_vs_ggfps_g_grid_size}
\end{figure}

\begin{figure}
    \centering
    \includegraphics[width=0.95\linewidth]{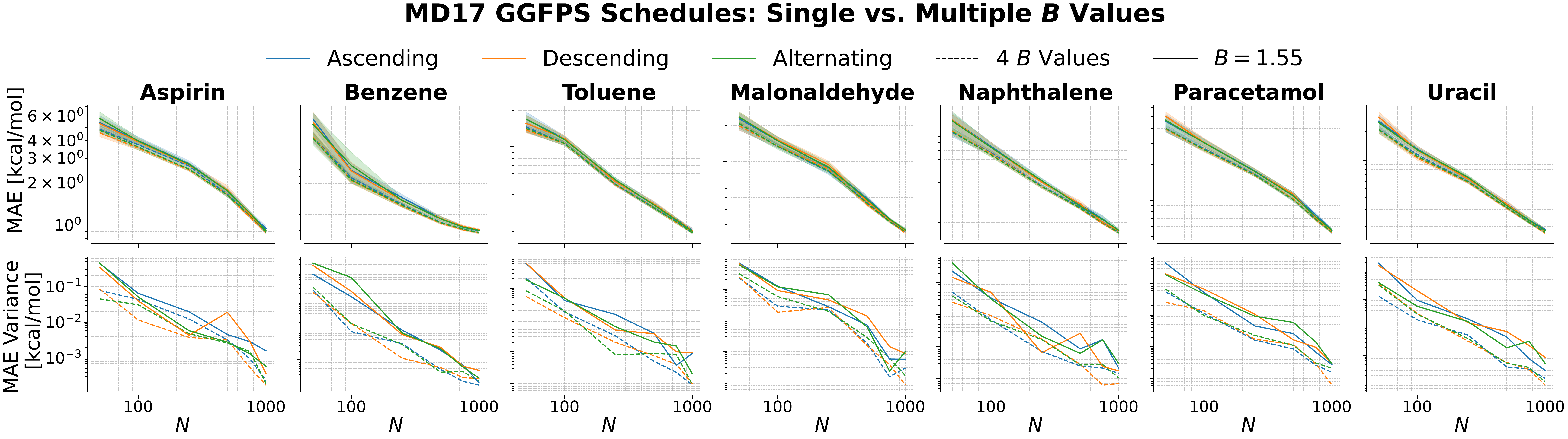}
    \caption{\added{MD17 learning curves comparing a single-$B$ setting versus treating $B$ as a hyperparameter. Columns correspond to the MD17 molecules. The top row shows MAE and the bottom row shows MAE variance as a function of training set size. Colors denote schedule strategies (ascending, descending, alternating). Line style denotes the $B$ strategy: dashed curves correspond to selecting from a set of four $B$ values (hyperparameter search), while solid curves correspond to a fixed single value $B=1.55$ (chosen as the best single-$B$ setting). Using four $B$ values yields a marginal improvement, while the single-$B$ setting remains competitive and is preferred when compute is constrained.}}
    \label{fig:SImd17_single_vs_multi_b}
\end{figure}

\begin{figure}
    \centering
    \includegraphics[width=0.9\linewidth]{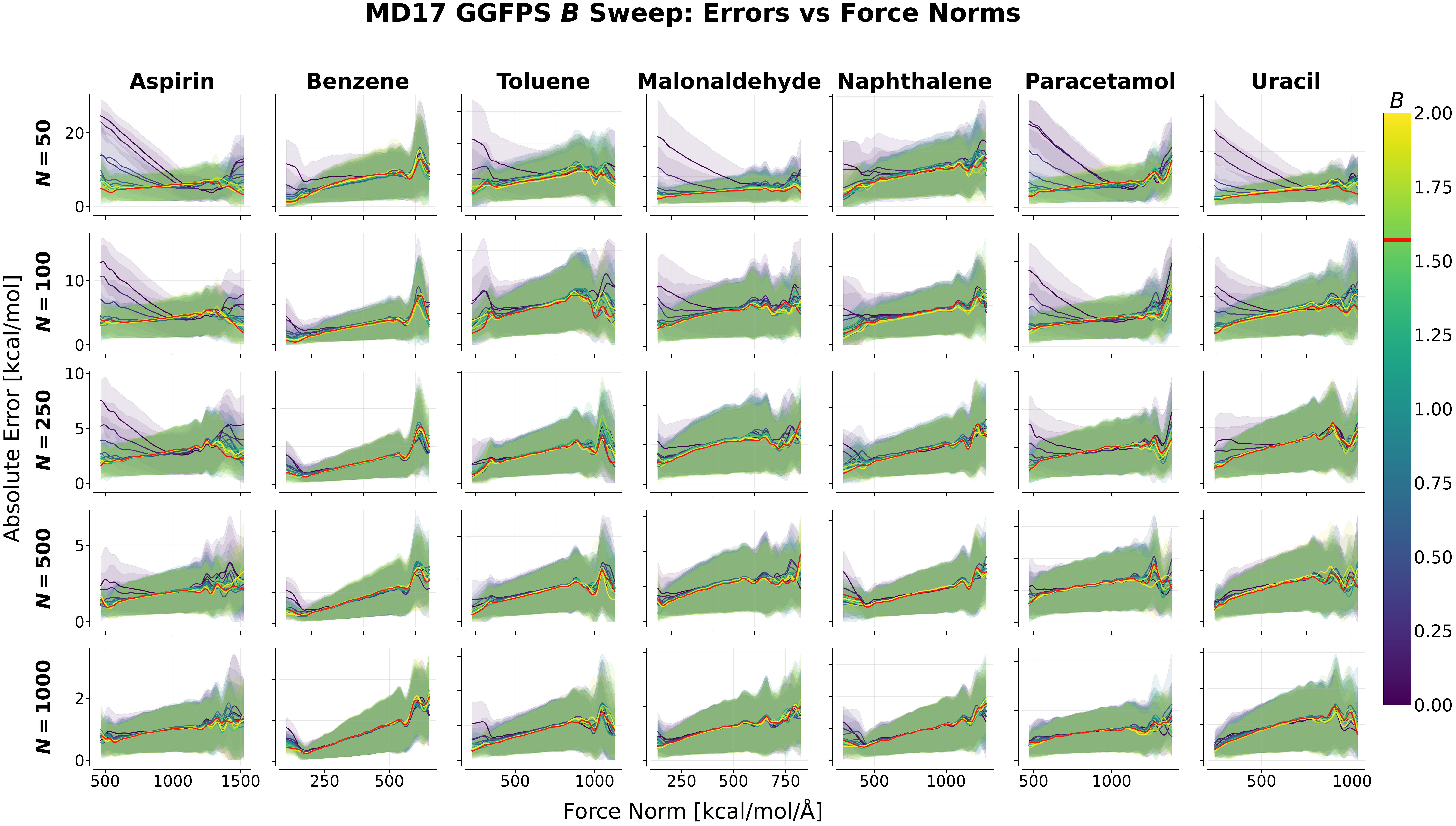}
    \caption{\added{MD17 absolute test errors versus force norms for a sweep over the gradient-bias range parameter $B$. Columns correspond to MD17 molecules and rows correspond to training set sizes $N \in \{50, 100, 250, 500, 1000\}$. Colors correspond to different $B$ values swept from $B=0$ (blue; equivalent to FPS) to $B=2$ (yellow), with uncertainty intervals shown around each curve. Low $B$ values lead to elevated errors for low force norms, while the best single-$B$ trade-off across molecules is achieved at $B=1.55$ (highlighted in red).}}
    \label{fig:SImd17_b_sweep_errors_vs_force_norms}
\end{figure}

\begin{figure}
    \centering
    \includegraphics[width=0.95\linewidth]{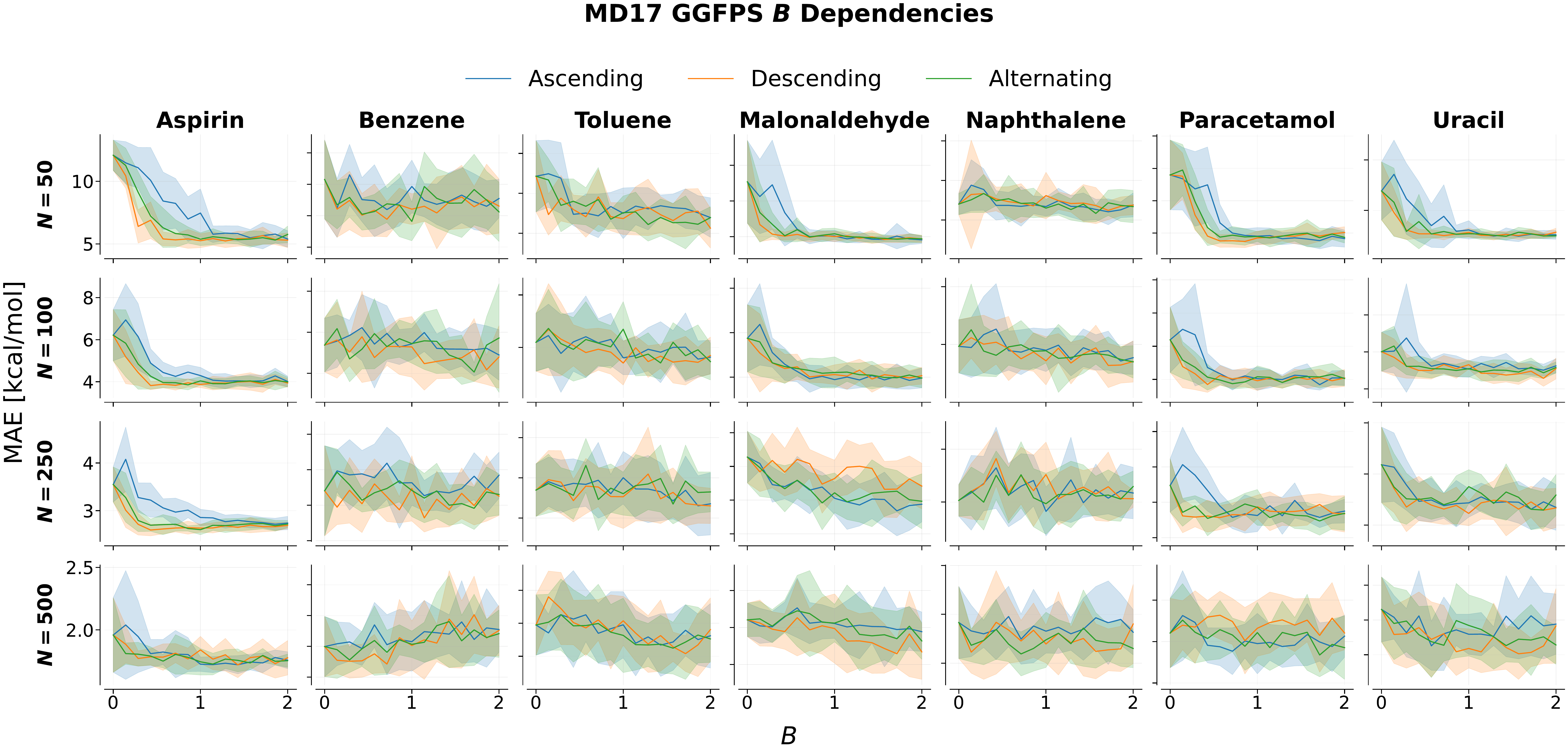}
    \caption{\added{MD17 MAE as a function of the gradient-bias range parameter $B$ for different schedule strategies. Columns correspond to the MD17 molecules and rows correspond to training set sizes $N \in \{50, 100, 250, 500\}$. Curves correspond to schedule strategies (ascending, descending, alternating). Overall, the descending schedule yields the most consistent performance across $B \in [0,2]$, with $B=1.55$ providing a robust single-$B$ choice.}}
    \label{fig:SImd17_beta_dependencies}
\end{figure}

\begin{figure}
    \centering
    \includegraphics[width=0.95\linewidth]{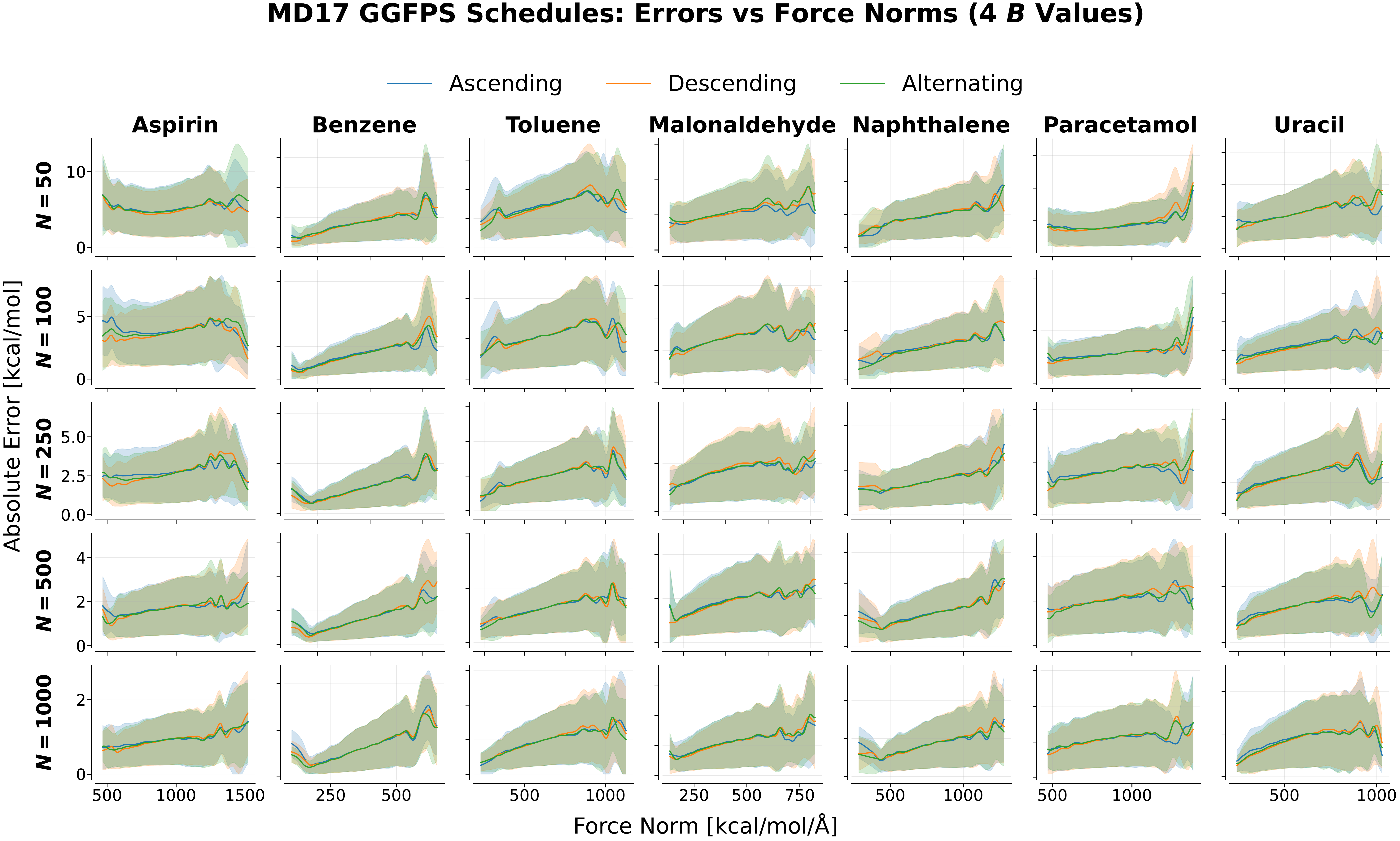}
    \caption{\added{MD17 absolute test errors versus force norms for different schedule strategies when $B$ is treated as a hyperparameter (four $B$ value grid). Columns correspond to MD17 molecules and rows correspond to training set sizes $N \in \{50, 100, 250, 500, 1000\}$. Curves correspond to schedule strategies (ascending, descending, alternating). The descending schedule is consistently comparable or slightly better across most force-norm regimes.}}
    \label{fig:SImd17_sched_error_vs_force_norms_4b}
\end{figure}

\begin{figure}
    \centering
    \includegraphics[width=0.95\linewidth]{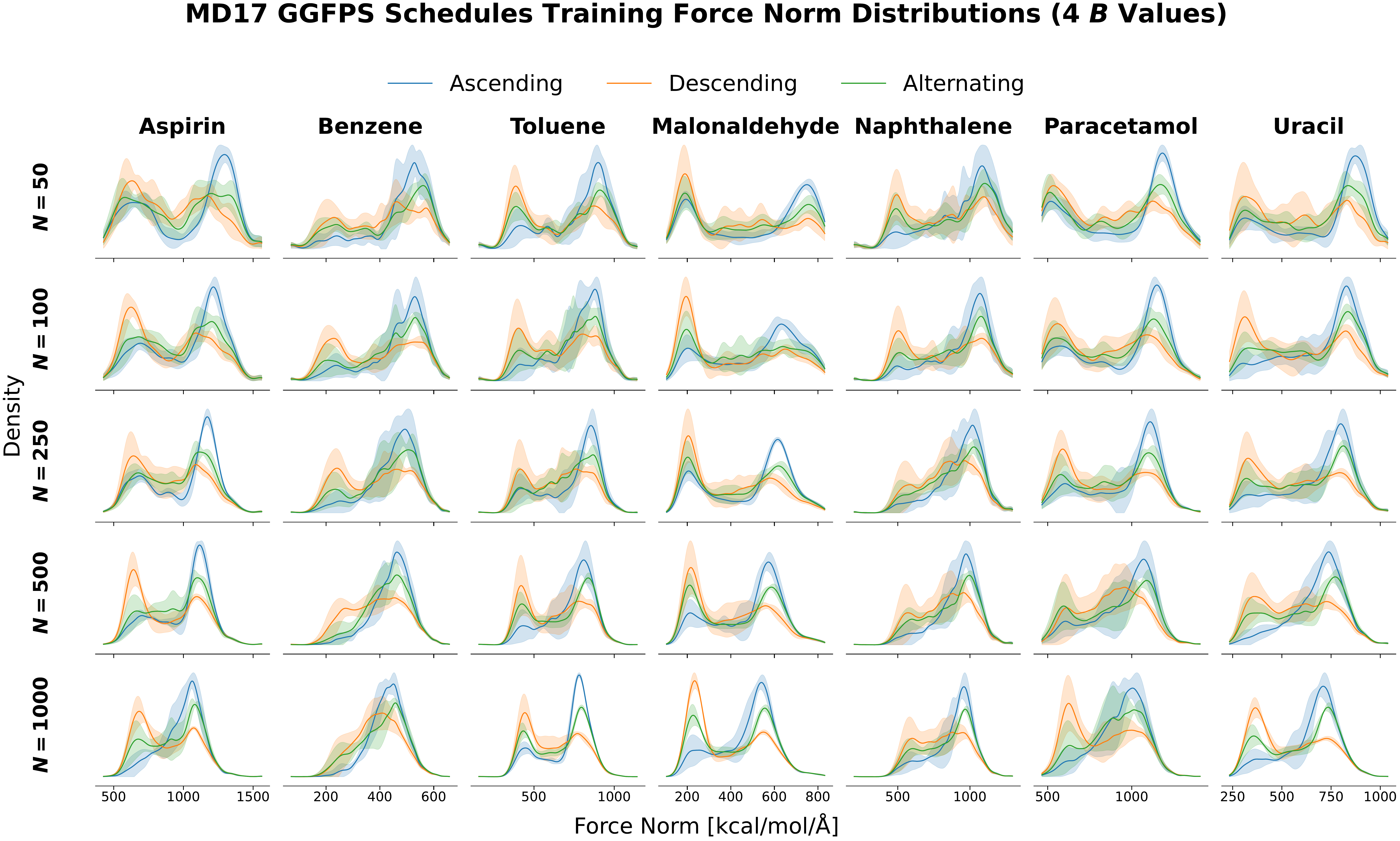}
    \caption{\added{MD17 training force-norm distributions under different schedule strategies when $B$ is treated as a hyperparameter (four $B$ value grid). Columns correspond to MD17 molecules and rows correspond to training set sizes $N \in \{50, 100, 250, 500, 1000\}$. Curves correspond to schedule strategies (ascending, descending, alternating). While the alternating schedule provides the best compromise between low- and high-force-norm coverage (i.e., balancing the low- and high-force bumps) across molecules and training set sizes, the descending schedule tends to result in more stable models.}}
    \label{fig:SImd17_sched_training_force_norm_distr_4b}
\end{figure}

\SISubsection{MD17 Additional Test Error vs. Force Norm Plots}{sec:SI_test_E_vs_Forces}

\begin{figure}
    \centering
    \includegraphics[width=0.8\linewidth]{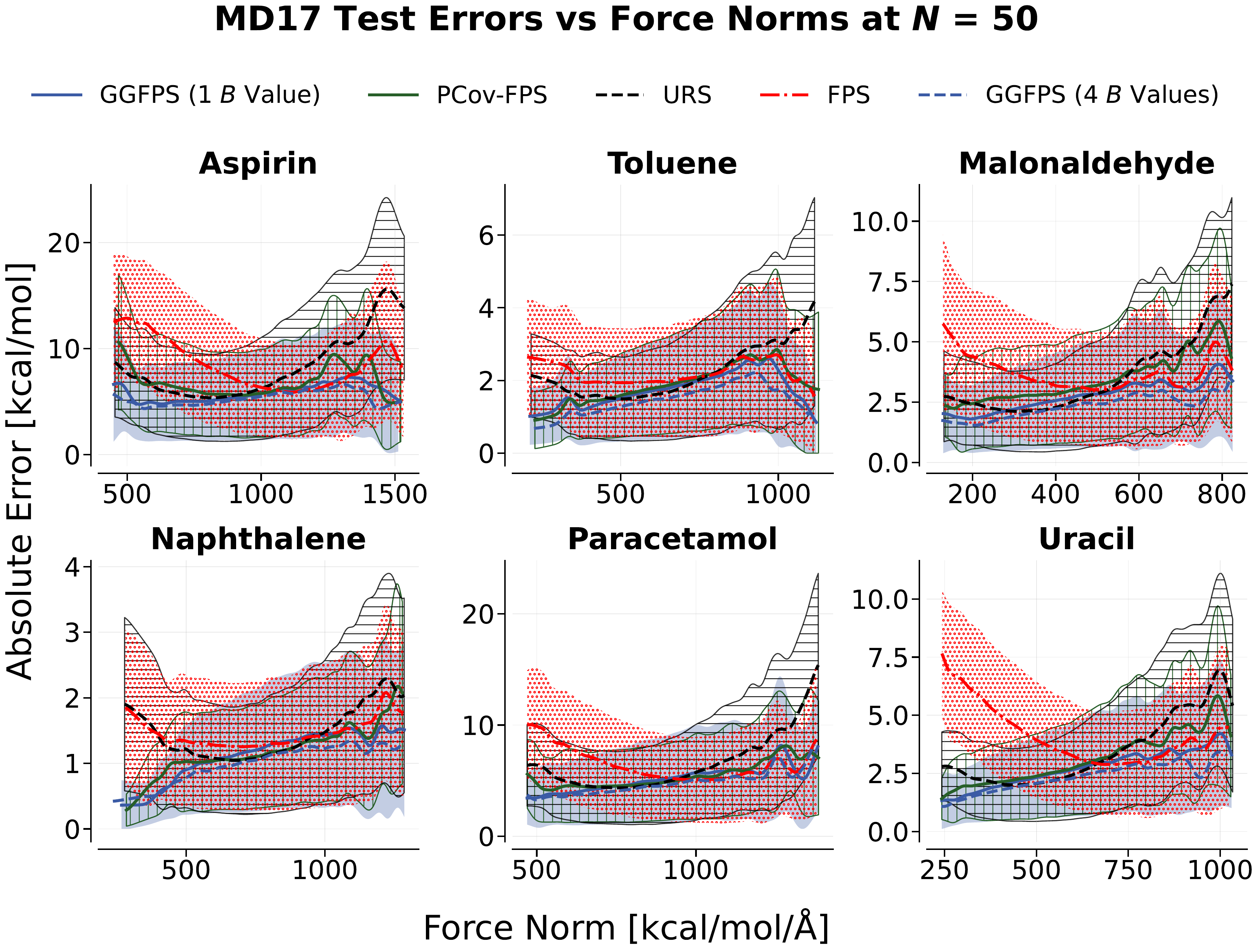}
    \caption{\added{MD17 trajectory test errors versus force norms for GGFPS (single-$B$ and $4B$), PCov-FPS, URS, and FPS  training sets (blue solid and blue dashed, green, black, and red), with a training set size ($N$) of 50 configurations. Variance bands for GGFPS $4B$ are not shown to reduce visual clutter.}}
    \label{fig:md17_test_errors_tss50}
\end{figure}

\begin{figure}
    \centering
    \includegraphics[width=0.8\linewidth]{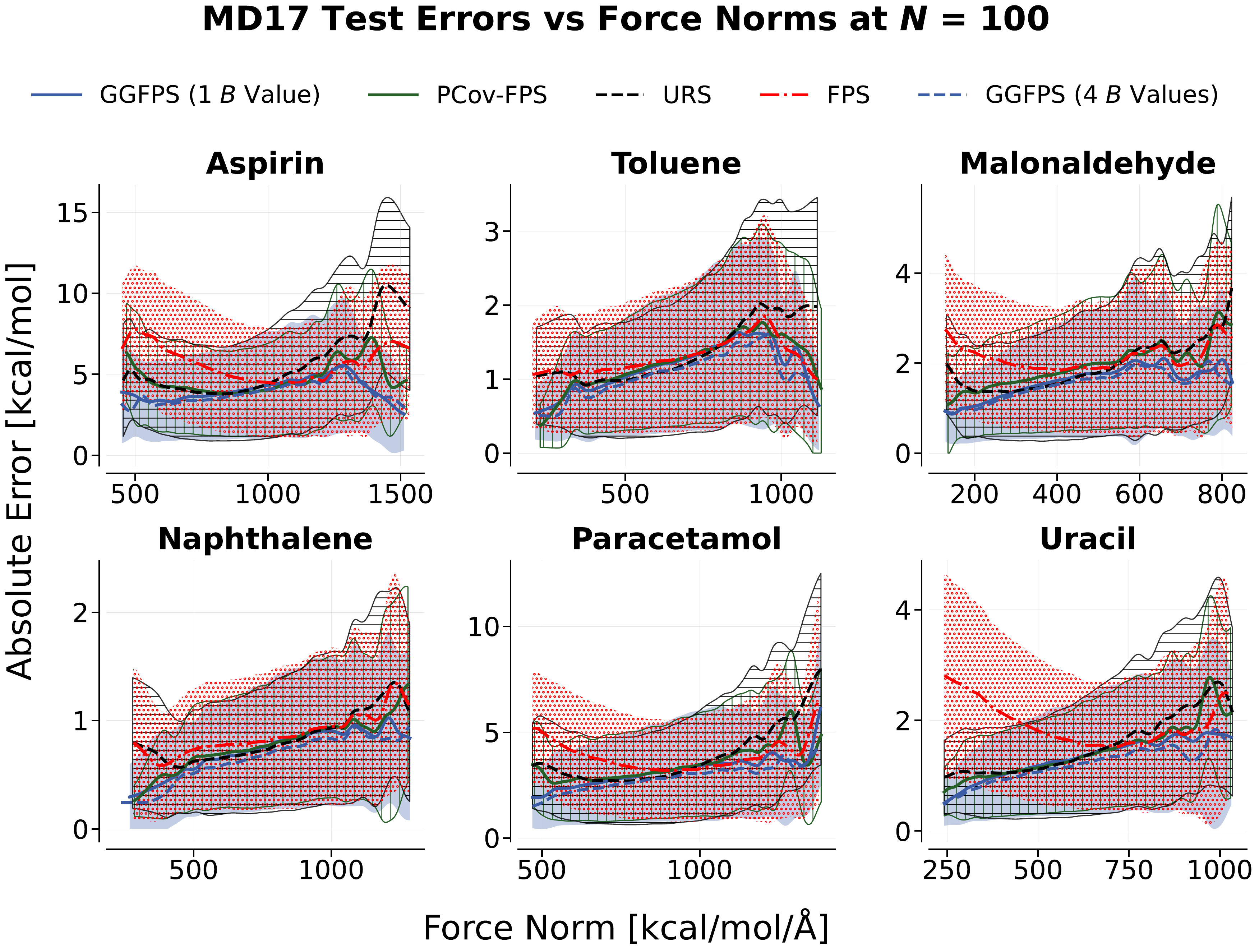}
    \caption{\added{MD17 trajectory test errors versus force norms for GGFPS (single-$B$ and $4B$), PCov-FPS, URS, and FPS  training sets (blue solid and blue dashed, green, black, and red), with a training set size ($N$) of 100 configurations. Variance bands for GGFPS $4B$ are not shown to reduce visual clutter.}}
    \label{fig:md17_test_errors_tss100}
\end{figure}

\begin{figure}
    \centering
    \includegraphics[width=0.8\linewidth]{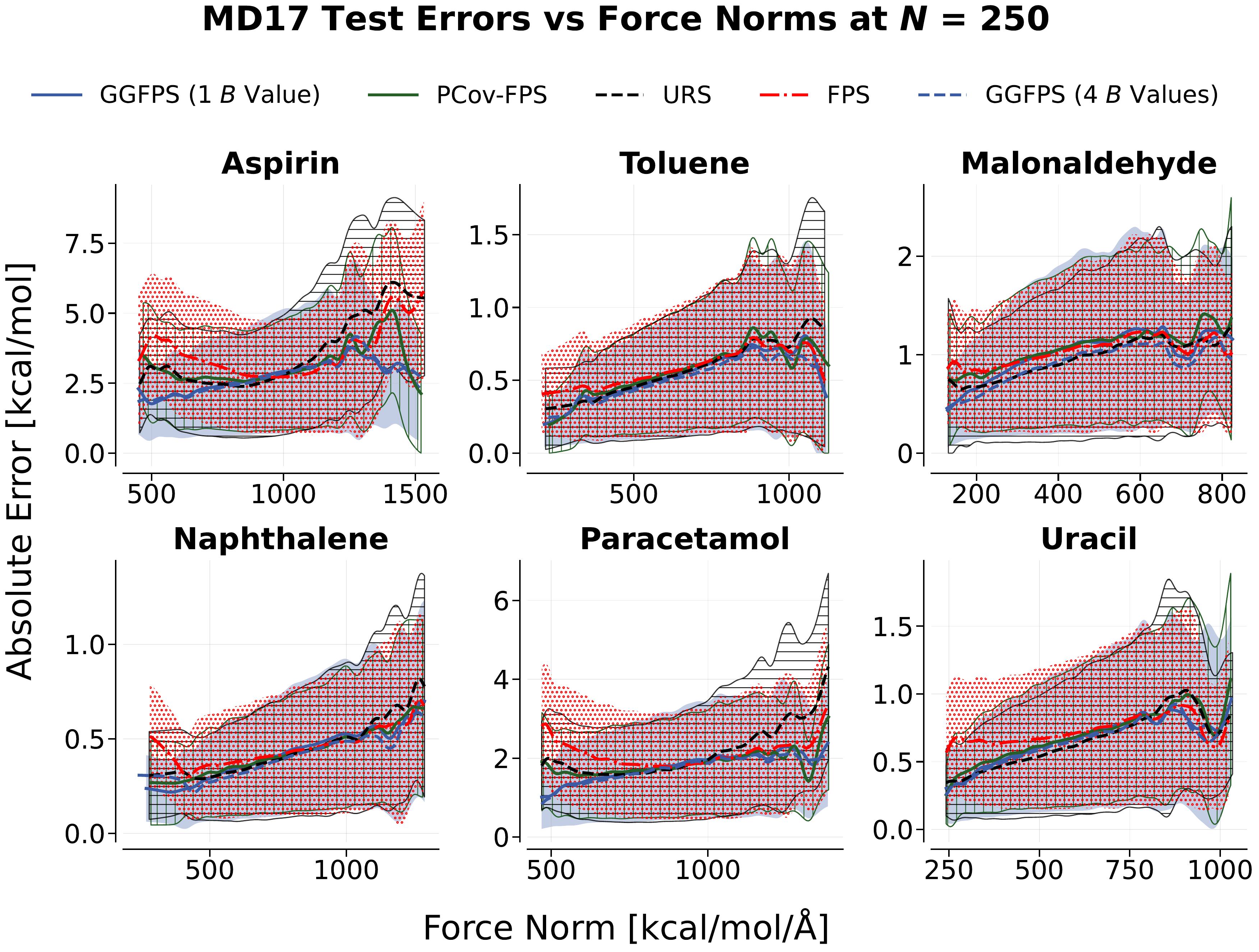}
    \caption{\added{MD17 trajectory test errors versus force norms for GGFPS (single-$B$ and $4B$), PCov-FPS, URS, and FPS  training sets (blue solid and blue dashed, green, black, and red), with a training set size ($N$) of 250 configurations. Variance bands for GGFPS $4B$ are not shown to reduce visual clutter.}}
    \label{fig:md17_test_errors_tss250}
\end{figure}

\begin{figure}
    \centering
    \includegraphics[width=0.8\linewidth]{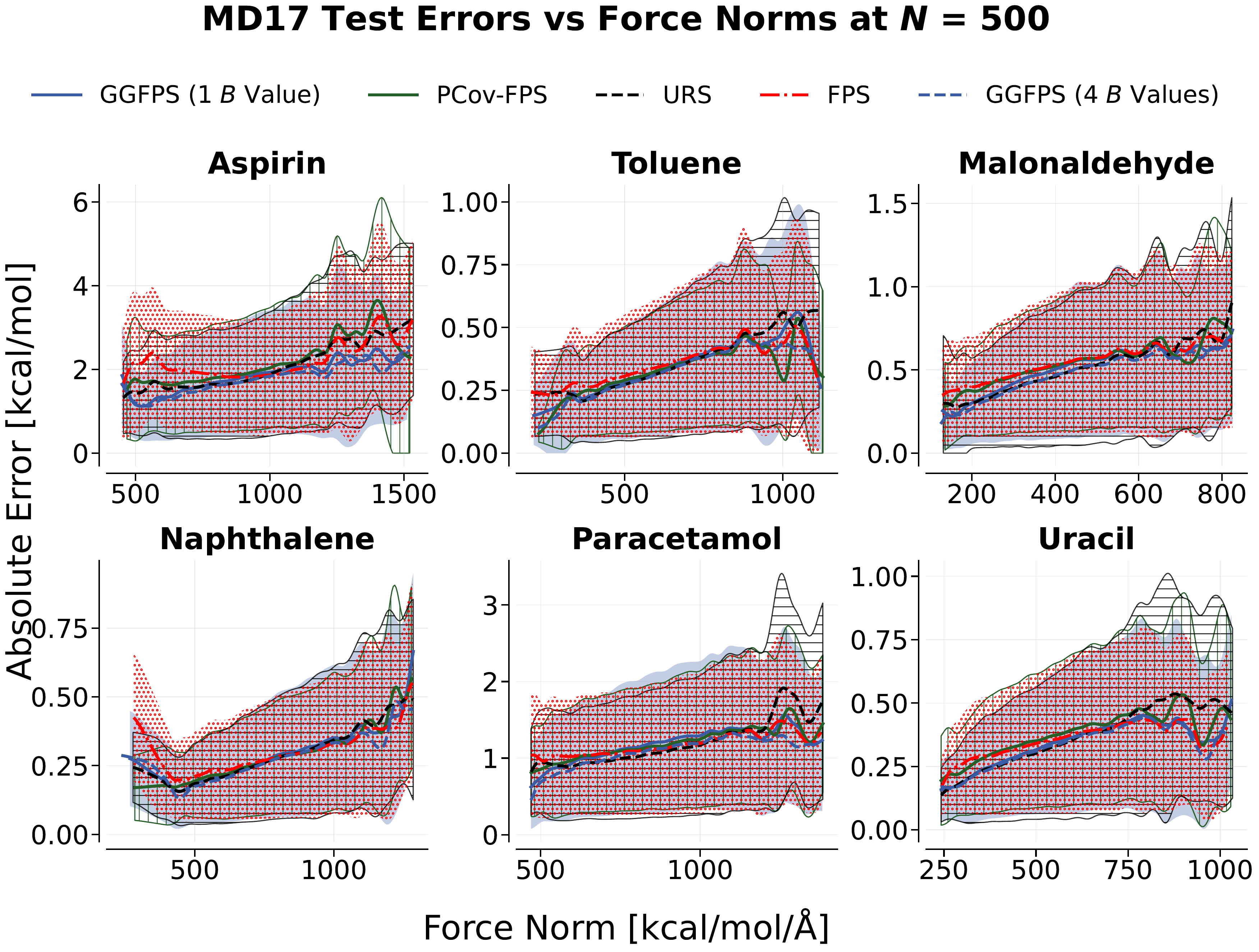}
    \caption{\added{MD17 trajectory test errors versus force norms for GGFPS (single-$B$ and $4B$), PCov-FPS, URS, and FPS  training sets (blue solid and blue dashed, green, black, and red), with a training set size ($N$) of 500 configurations. Variance bands for GGFPS $4B$ are not shown to reduce visual clutter.}}
    \label{fig:md17_test_errors_tss500}
\end{figure}

\begin{figure}
    \centering
    \includegraphics[width=0.8\linewidth]{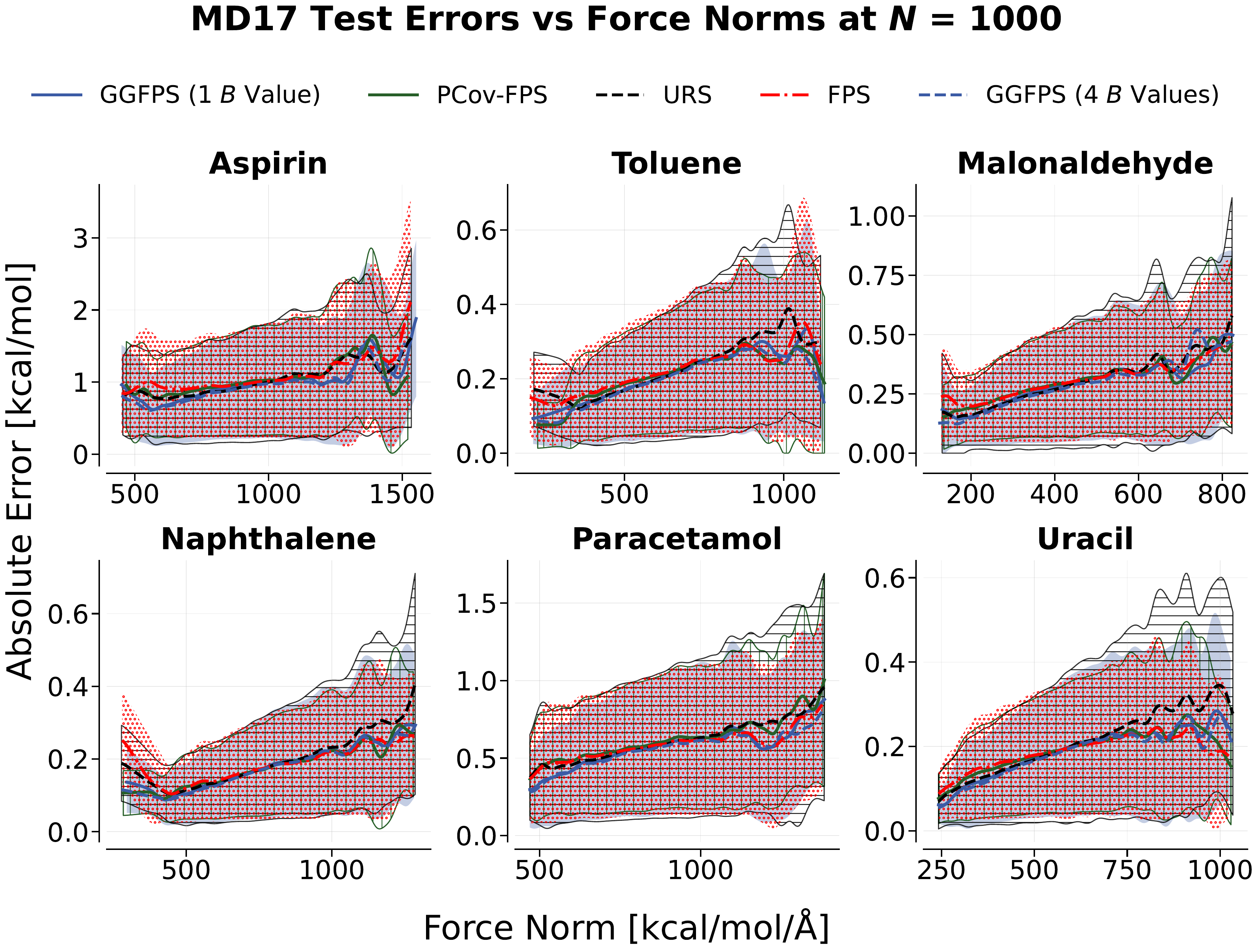}
    \caption{\added{MD17 trajectory test errors versus force norms for GGFPS (single-$B$ and $4B$), PCov-FPS, URS, and FPS  training sets (blue solid and blue dashed, green, black, and red), with a training set size ($N$) of 1000 configurations. Variance bands for GGFPS $4B$ are not shown to reduce visual clutter.}}
    \label{fig:md17_test_errors_tss1000}
\end{figure}

\SISubsection{PCov baselines using force labels (comparison)}{sec:SI_pcov_force_labels}

\begin{figure}
    \centering
    \includegraphics[width=0.98\linewidth]{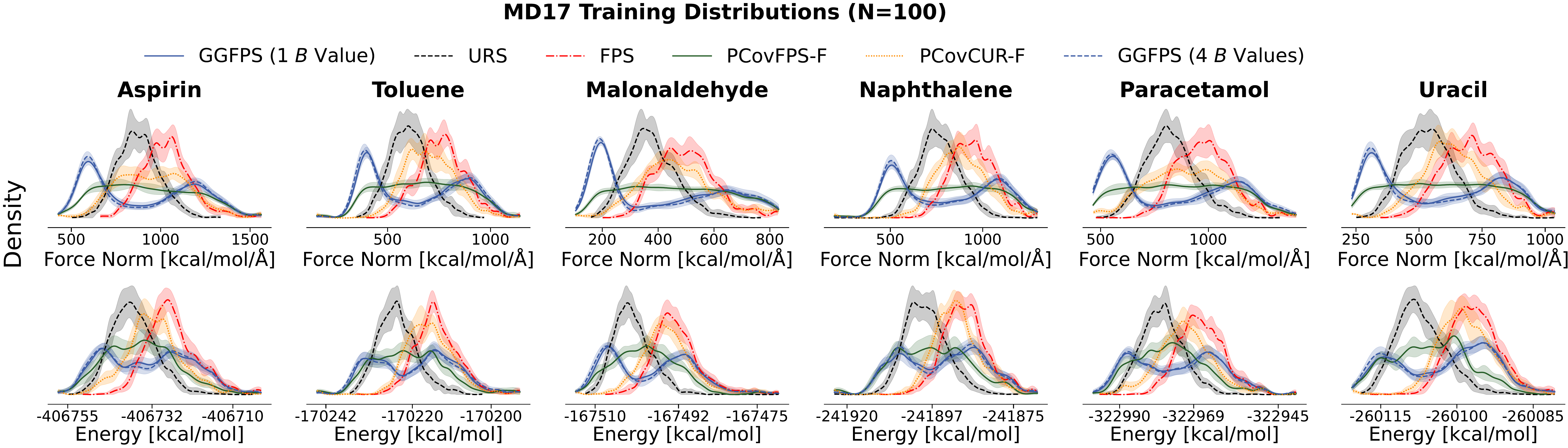}
    \caption{\added{MD17 training-set force-norm distributions for $N=100$ selected configurations when using PCov-FPS/PCov-CUR variants that incorporate \emph{force-label} information. This experiment is included only as a like-for-like supervised comparison (since the standard PCov baselines are typically defined with respect to energies rather than forces).}}
    \label{fig:SI_md17_pcov_force_labels_train_dists_n100}
\end{figure}

\begin{figure}
    \centering
    \includegraphics[width=0.98\linewidth]{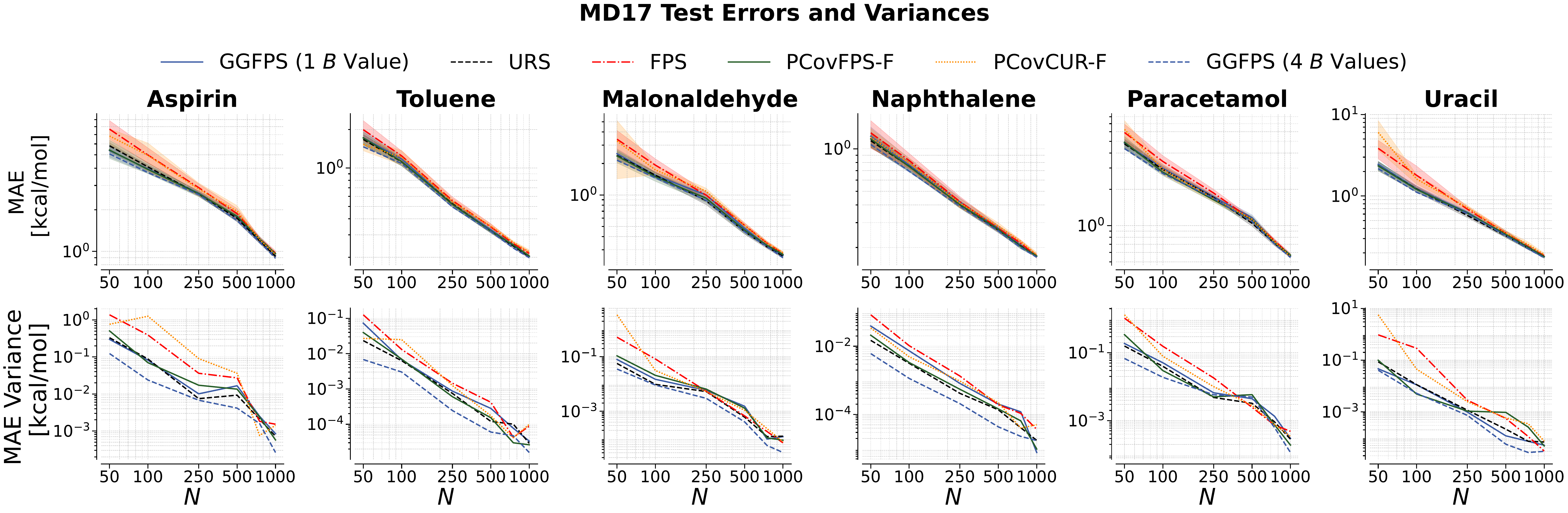}
    \caption{\added{MD17 learning curves (MAE and MAE variance) for PCov-FPS/PCov-CUR variants that incorporate \emph{force-label} information. This experiment is included only as a like-for-like supervised comparison; using force labels inside PCov is not the standard baseline definition.}}
    \label{fig:SI_md17_pcov_force_labels_lcs}
\end{figure}

\begin{figure}
    \centering
    \includegraphics[width=0.8\linewidth]{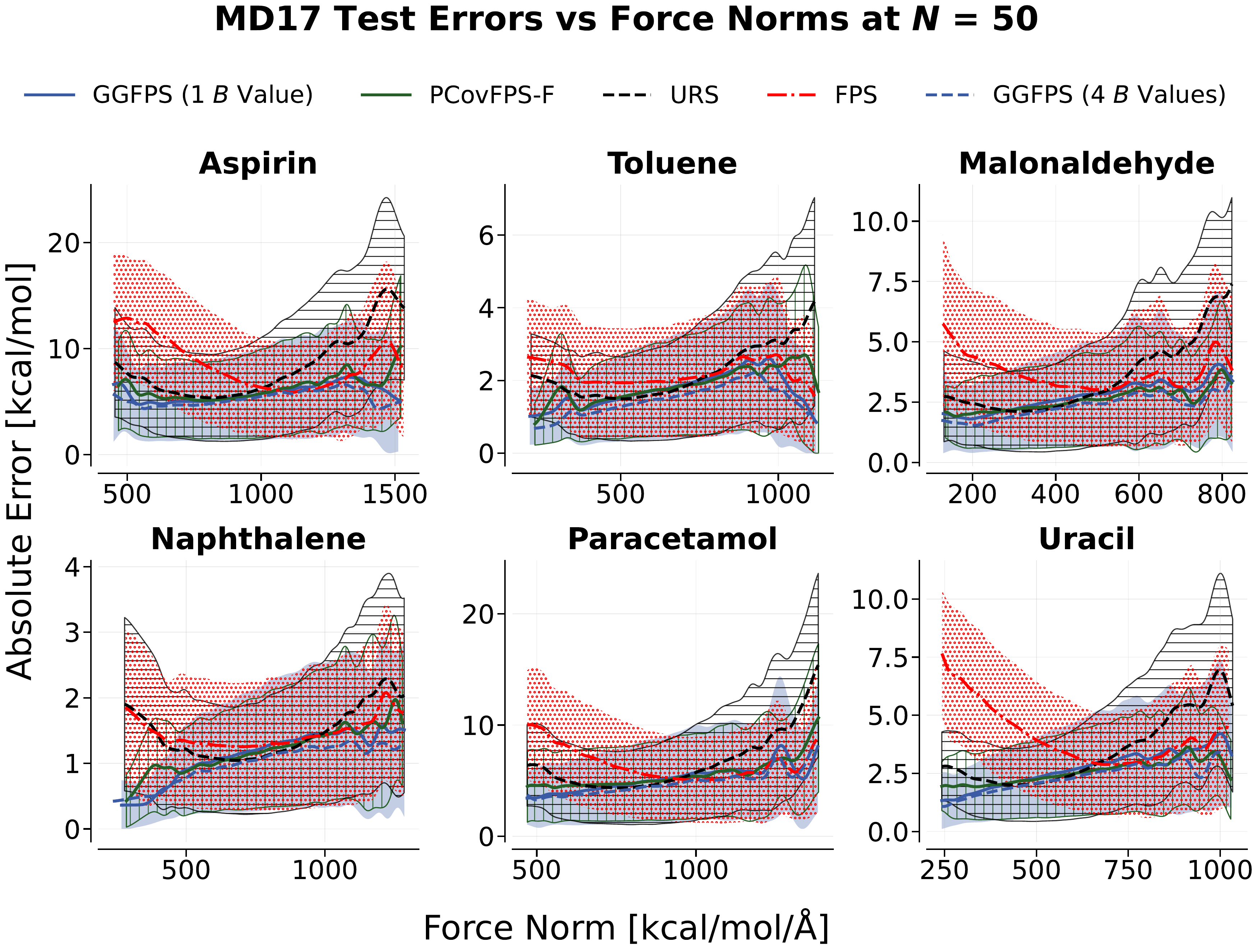}
    \caption{\added{MD17 absolute test errors versus force norms at $N=50$ when using PCov-FPS/PCov-CUR variants that incorporate \emph{force-label} information. This experiment is included only as a like-for-like supervised comparison; using force labels inside PCov is not the standard baseline definition.}}
    \label{fig:md17_test_errors_tss50_force}
\end{figure}

\begin{figure}
    \centering
    \includegraphics[width=0.8\linewidth]{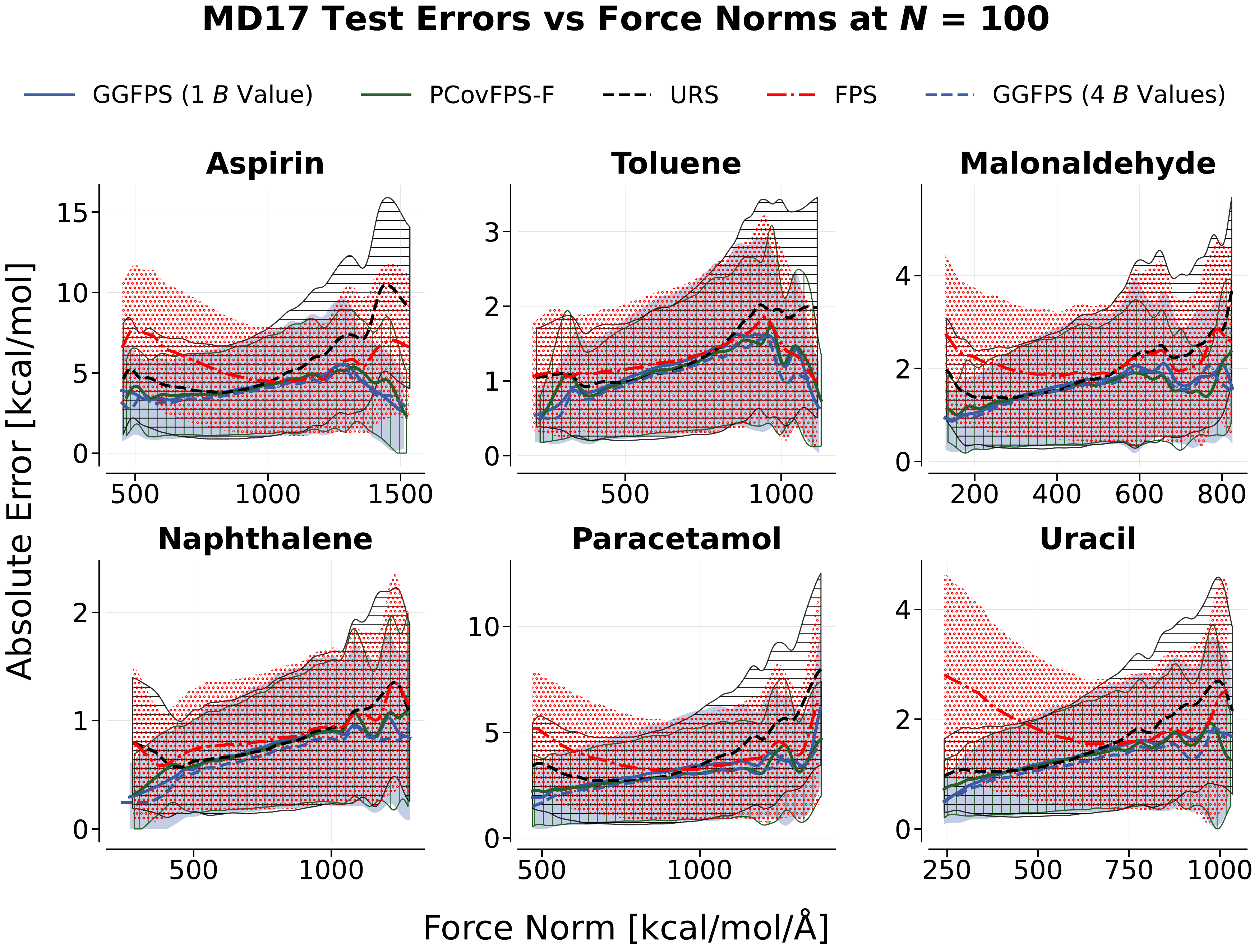}
    \caption{\added{MD17 absolute test errors versus force norms at $N=100$ when using PCov-FPS/PCov-CUR variants that incorporate \textit{force-label} information. This experiment is included only as a like-for-like supervised comparison; using force labels inside PCov is not the standard baseline definition.}}
    \label{fig:md17_test_errors_tss100_force}
\end{figure}

\begin{figure}
    \centering
    \includegraphics[width=0.8\linewidth]{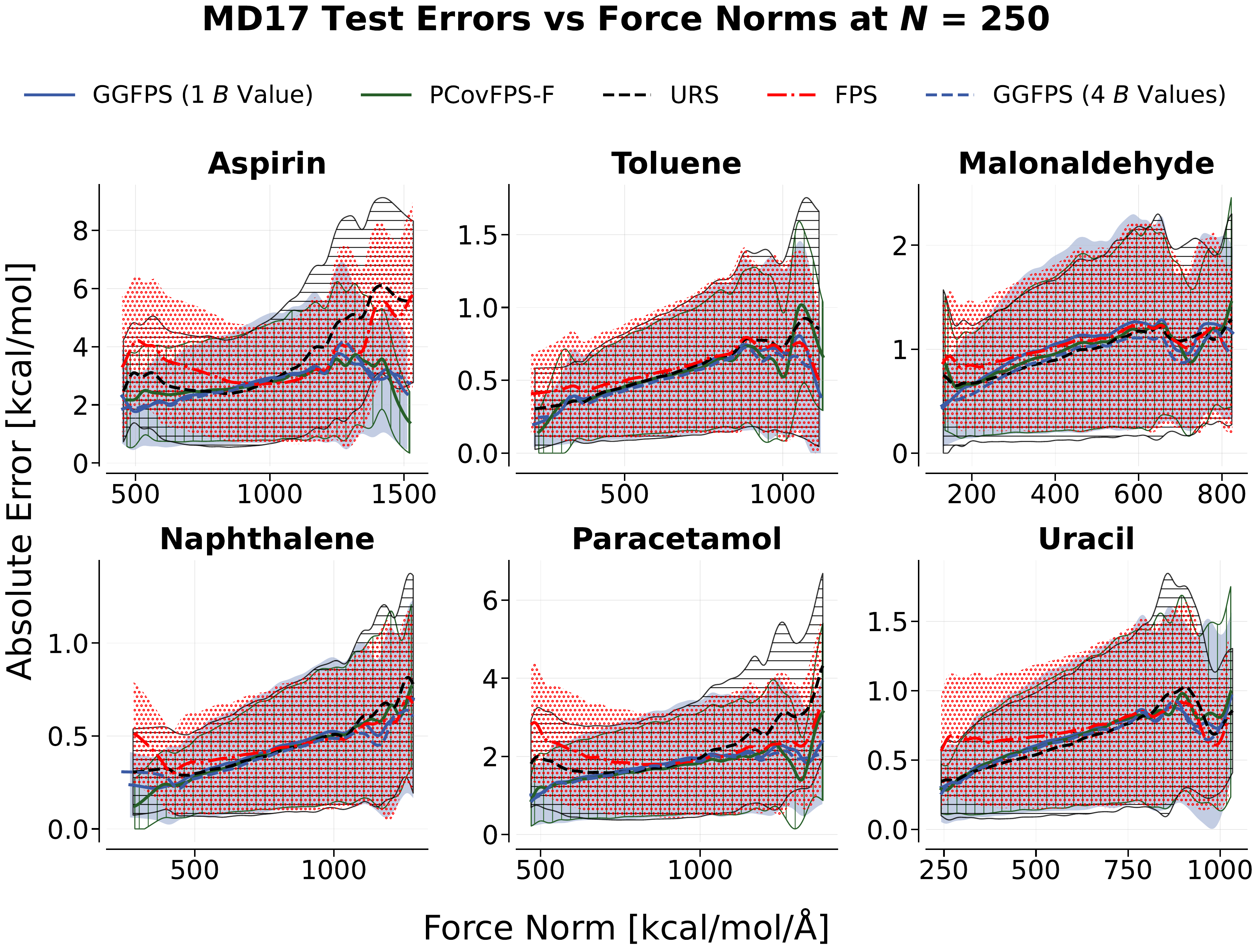}
    \caption{\added{MD17 absolute test errors versus force norms at $N=250$ when using PCov-FPS/PCov-CUR variants that incorporate \emph{force-label} information. This experiment is included only as a like-for-like supervised comparison; using force labels inside PCov is not the standard baseline definition.}}
    \label{fig:md17_test_errors_tss250_force}
\end{figure}

\begin{figure}
    \centering
    \includegraphics[width=0.8\linewidth]{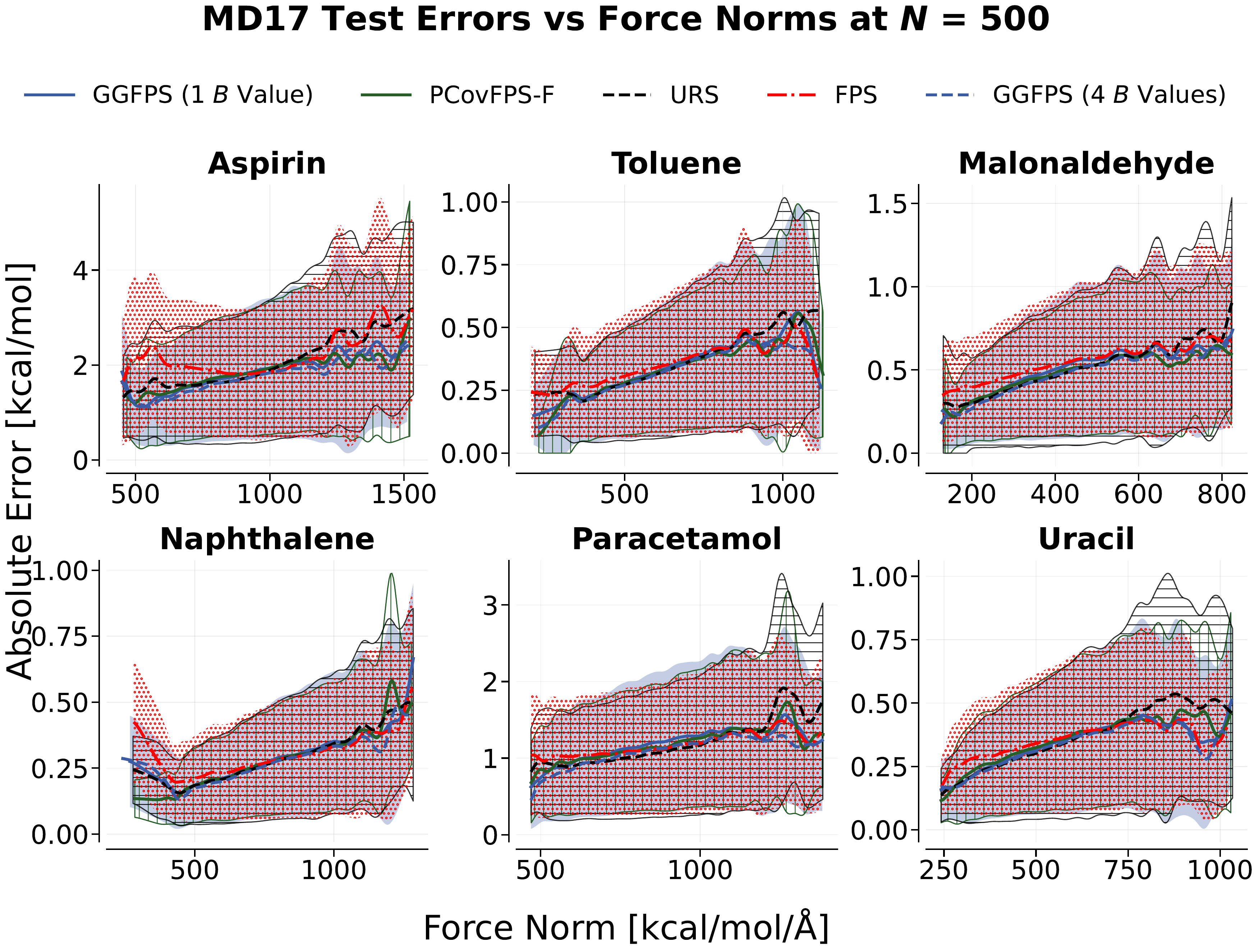}
    \caption{\added{MD17 absolute test errors versus force norms at $N=500$ when using PCov-FPS/PCov-CUR variants that incorporate \emph{force-label} information. This experiment is included only as a like-for-like supervised comparison; using force labels inside PCov is not the standard baseline definition.}}
    \label{fig:md17_test_errors_tss500_force}
\end{figure}

\begin{figure}
    \centering
    \includegraphics[width=0.8\linewidth]{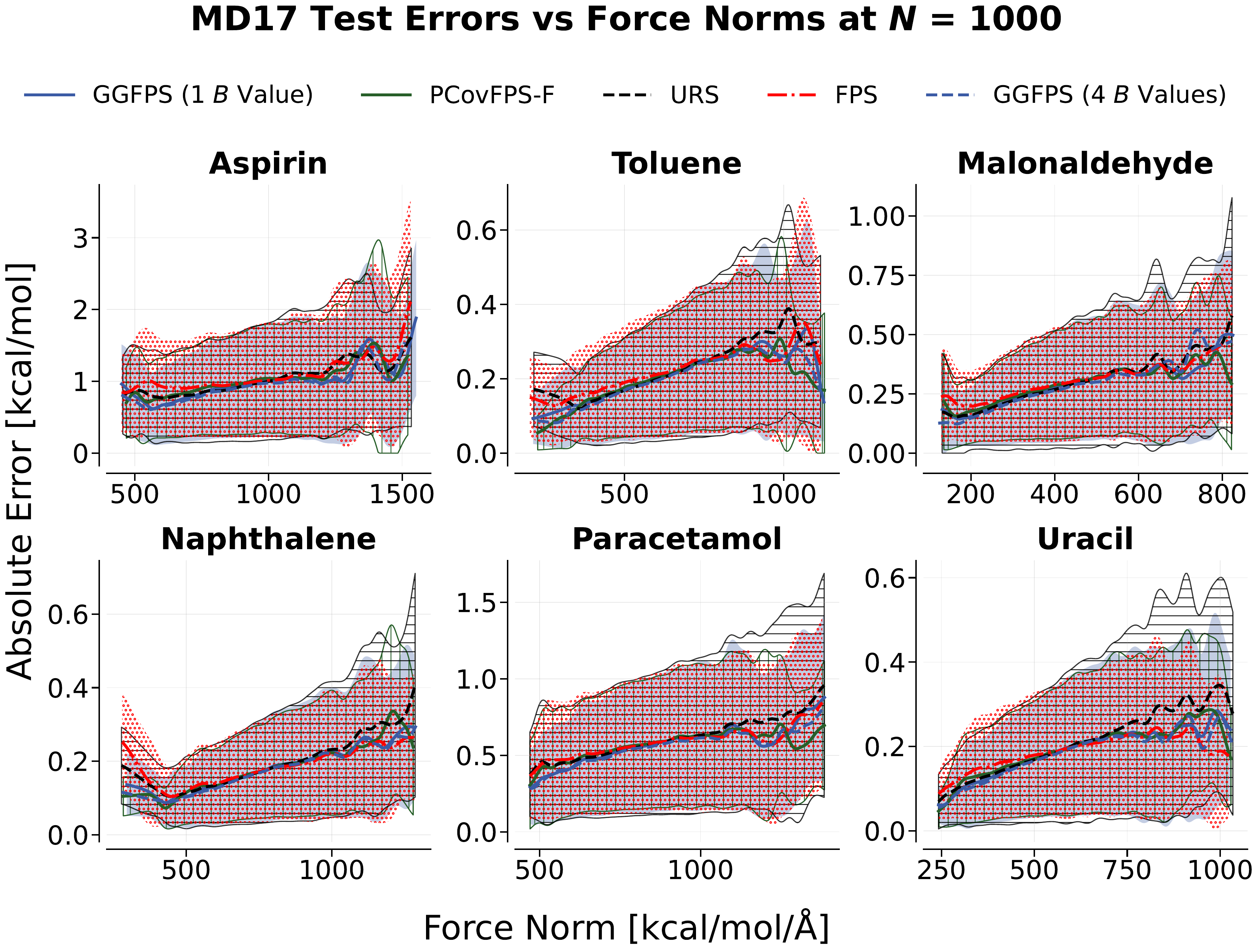}
    \caption{\added{MD17 absolute test errors versus force norms at $N=1,000$ when using PCov-FPS/PCov-CUR variants that incorporate \emph{force-label} information. This experiment is included only as a like-for-like supervised comparison; using force labels inside PCov is not the standard baseline definition.}}
    \label{fig:md17_test_errors_tss1000_force}
\end{figure}

\SISubsection{Timing benchmarks}{sec:SI_timing}

\begin{figure}
    \centering
    \includegraphics[width=0.4\linewidth]{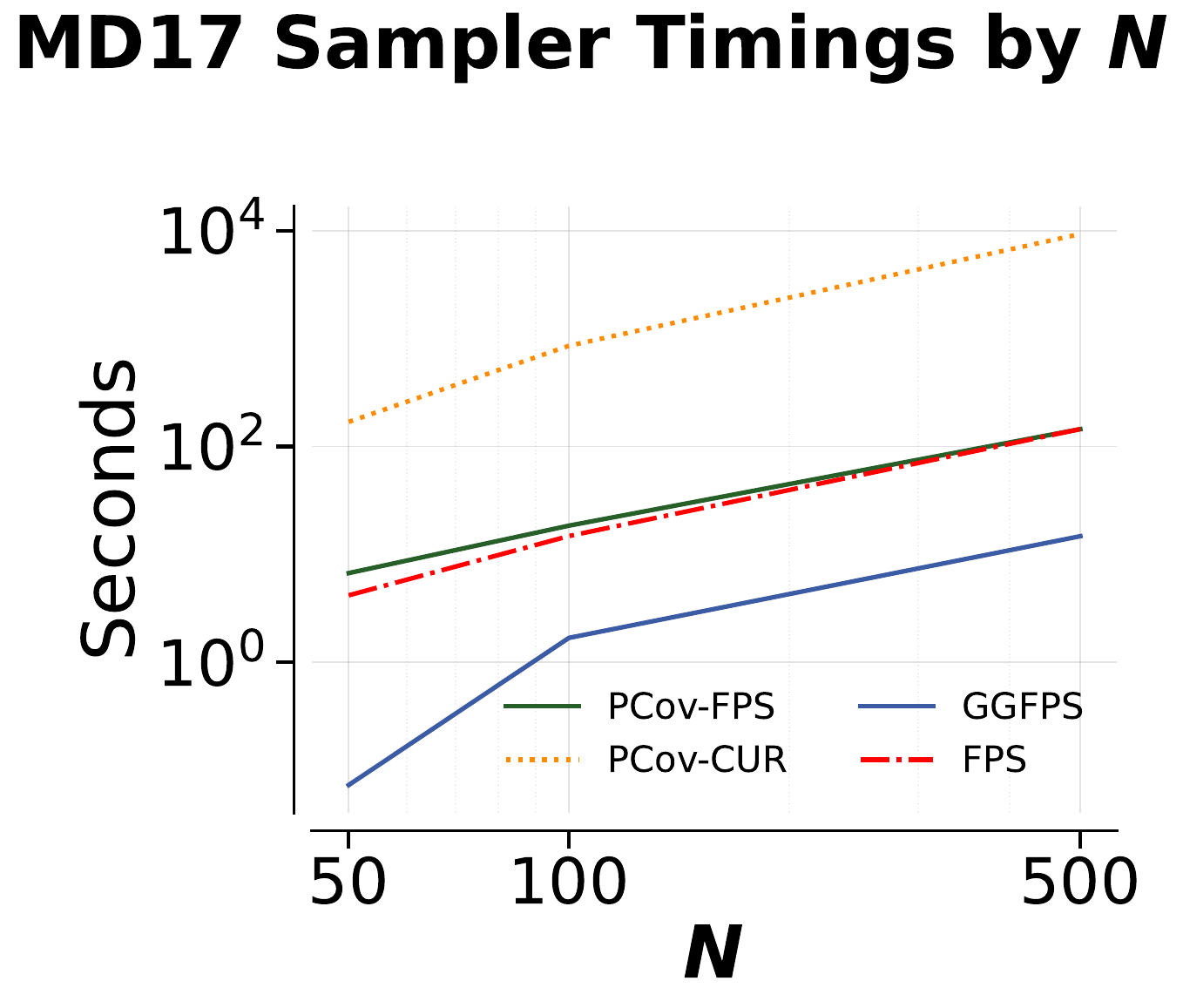}
    \caption{\added{Wall-clock runtime (seconds) versus training set size $N$ for different sampling methods on MD17 aspirin FCHL19 inputs. Curves compare on-the-fly GGFPS (blue), FPS (red), PCov-FPS (green), and PCov-CUR (orange). GGFPS is approximately an order of magnitude faster than the next-fastest methods (FPS and PCov-FPS, which exhibit similar runtimes), while PCov-CUR is substantially slower.}}
    \label{fig:SI_md17_sampler_timing_by_n}
\end{figure}

\begin{figure}
    \centering
    \includegraphics[width=0.4\linewidth]{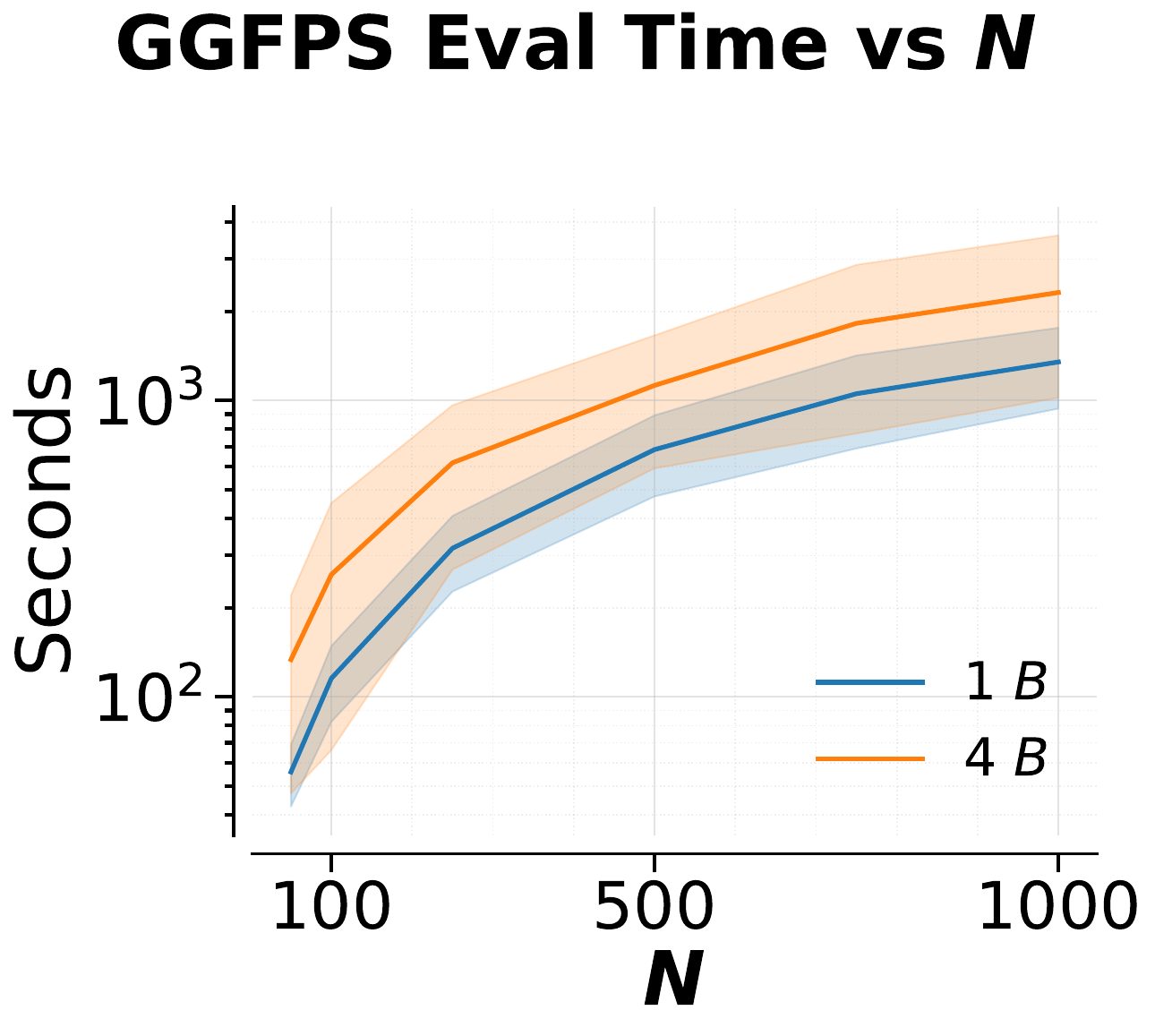}
    \caption{\added{Fully pre-computed distance matrix GGFPS evaluation (train/cross-validation/test cycle) timing benchmarks (seconds) versus training set size $N$, split gradient bias hyperparameter $B$ count. The $B=1$ setting (blue) is faster than the $B=4$ setting (orange) by a factor of 2.}}
    \label{fig:SI_ggfps_eval_time_vs_n}
\end{figure}

\begin{figure}
    \centering
    \begin{subfigure}{0.98\linewidth}
        \centering
        \includegraphics[width=\linewidth]{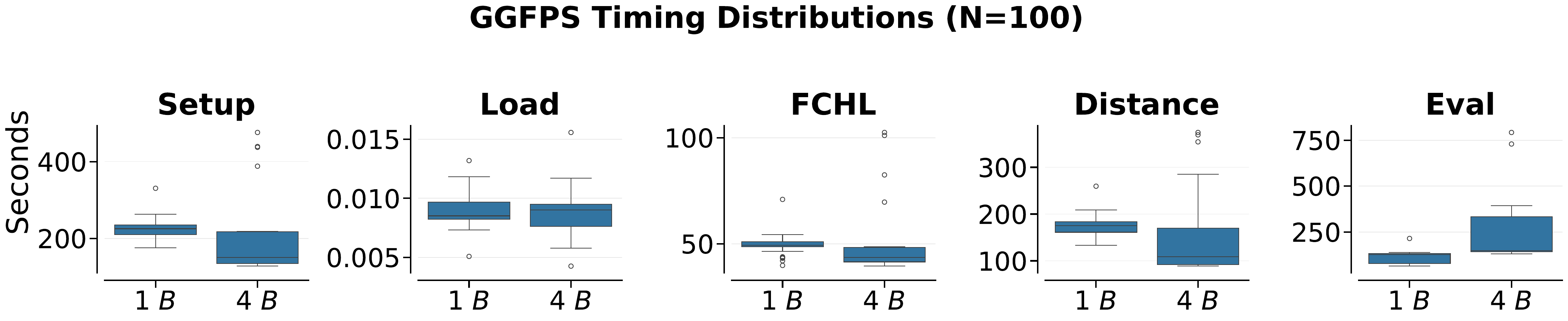}
        \label{fig:SI_ggfps_timing_distributions_n100}
    \end{subfigure}

    \vspace{0.5em}

    \begin{subfigure}{0.98\linewidth}
        \centering
        \includegraphics[width=\linewidth]{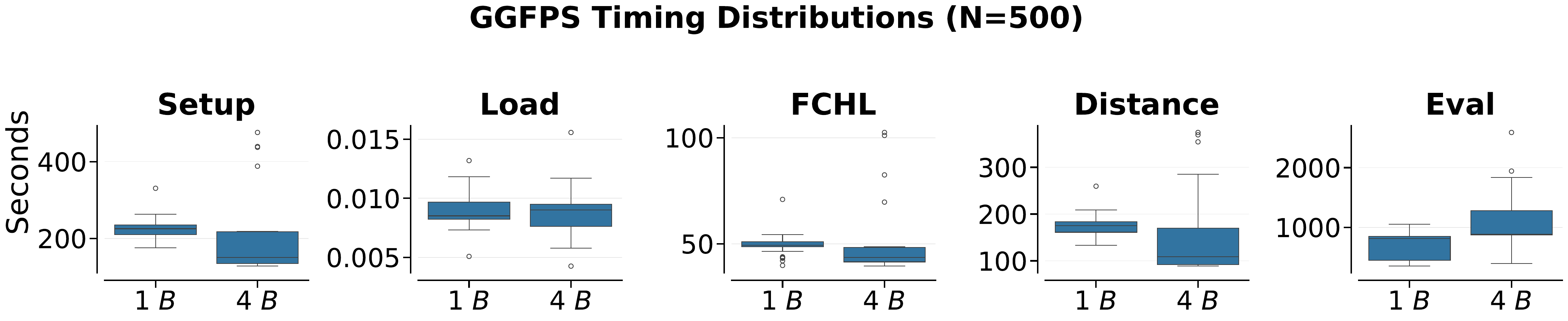}
        \label{fig:SI_ggfps_timing_distributions_n500}
    \end{subfigure}

    \caption{\added{Full GGFPS (with pre-computed distance matrix) pipeline timing breakdown shown as box plots (with whiskers and outliers) for $B=1$ and $B=4$ at $N=100$ and $N=500$. The five panels report the distributions of wall-clock time (seconds) for: setup, loading, FCHL descriptor generation, distance calculation, and evaluation.}}
    \label{fig:SI_ggfps_timing_distributions}
\end{figure}

\FloatBarrier